\documentclass[Afour,sageh,times]{sagej}

\usepackage{moreverb,url}

\usepackage[colorlinks,bookmarksopen,bookmarksnumbered,citecolor=blue,urlcolor=blue]{hyperref}
\usepackage{float}
\usepackage{amsmath}

\newcommand\BibTeX{{\rmfamily B\kern-.05em \textsc{i\kern-.025em b}\kern-.08em
		T\kern-.1667em\lower.7ex\hbox{E}\kern-.125emX}}

\def\volumeyear{2022}
\usepackage{makecell}
\usepackage{graphicx}
\usepackage{natbib}
\setcitestyle{square, comma, numbers, super}
\usepackage{epstopdf}
\usepackage{dirtree,float}
\usepackage{bbding}
\usepackage{graphicx} 
\usepackage{multirow}
\usepackage{color}
\newcommand{\revise}{\textcolor{black}}

\DeclareUnicodeCharacter{2033}{μ}

\newcommand{\orcid}[1]{\href{https://orcid.org/#1}{\includegraphics[width=10pt]{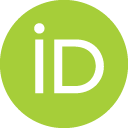}}}

\begin{document}
	
	\title{USTC FLICAR: A \revise{Sensors} Fusion Dataset of LiDAR-Inertial-Camera for Heavy-duty Autonomous Aerial Work Robots}
	
	\author{Ziming Wang \orcid{0000-0003-0499-6848}, Yujiang Liu \orcid{0009-0001-2425-3091}, Yifan Duan \orcid{0009-0004-0754-3953}, Xingchen Li \orcid{0009-0005-0161-9539}, Xinran Zhang \orcid{0009-0002-1589-4222}, \\Jianmin Ji \orcid{0000-0002-1515-0402}, Erbao Dong \orcid{0000-0002-4062-9730} and Yanyong Zhang \orcid{0000-0001-9046-798X}}
	\affiliation{University of Science and Technology of China, 96 Jinzhai Road, Hefei, 230026, Anhui, China.} 
	
	\corrauth{Erbao Dong, CAS Key Laboratory of Mechanical Behavior and Design of Materials, Department of Precision Machinery and Precision Instrumentation, University of Science and Technology of China, 96 Jinzhai Road, Hefei, Anhui Province,230026, China.}
	
	\email{\{zimingwang, lyj0910, dyf0202, starlet, zxrr\}@mail.ustc.edu.cn, \{jianmin, ebdong*, yanyongz\}@ustc.edu.cn}
	
	\begin{abstract} 
		In this paper, we present the \textit{USTC FLICAR Dataset}, which is dedicated to the development of simultaneous localization and mapping  and precise 3D reconstruction of the workspace for heavy-duty autonomous aerial work robots. In recent years, numerous public datasets have played significant roles in the advancement of autonomous cars and unmanned aerial vehicles (UAVs). However, these two platforms differ from aerial work robots: UAVs are limited in their payload capacity, while cars are restricted to two-dimensional movements. To fill this gap, we create the ``Giraffe" mapping robot based on a bucket truck, which is equipped with a variety of well-calibrated and synchronized sensors: four 3D LiDARs, two stereo cameras, two monocular cameras, Inertial Measurement Units (IMUs), and a GNSS/INS system. A laser tracker is used to record the millimeter-level ground truth positions. We also make its ground twin, the  ``Okapi" mapping robot, to gather data for comparison. \revise{The proposed dataset extends the typical autonomous driving sensing suite to aerial scenes, demonstrating the potential of combining autonomous driving perception systems with bucket trucks to create a versatile autonomous aerial working platform. Moreover, based on the Segment Anything Model (SAM), we produce the Semantic FLICAR dataset, which provides fine-grained semantic segmentation annotations for multimodal continuous data in both temporal and spatial dimensions.} The dataset is available for download at: \url{https://ustc-flicar.github.io/}.
	\end{abstract}
	
	\keywords{Dataset, aerial work, aerial robot, mobile robot, SLAM, \revise{semantic segmentation}, computer vision, LiDAR, cameras, bucket truck}
	
	\maketitle

        \begin{figure}[t]
		\centering
		\includegraphics[height=5.98cm,width=8.5cm]{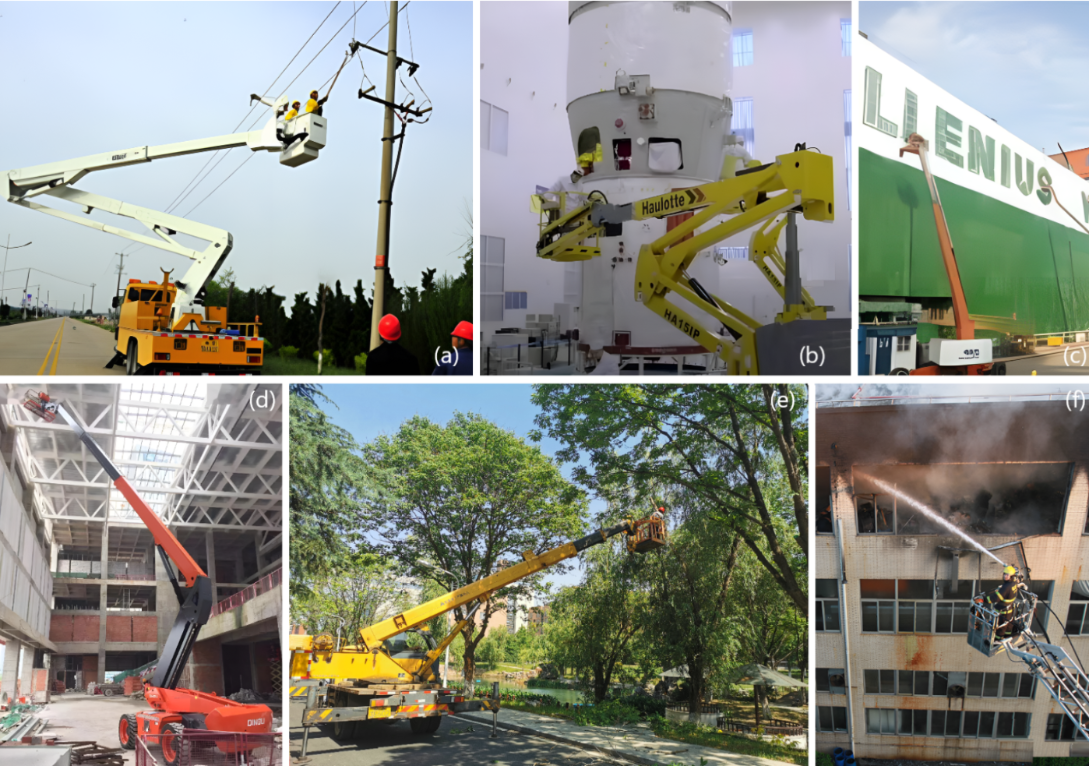}
		\caption{Typical aerial work scenes:\\ (a) repair and maintenance of electrical power facilities, (b) machinery manufacturing, (c) ship maintenance, (d) building construction,(e) tree trimming, and (f) aerial fire fighting and rescue}
		\label{F1}
	\end{figure}

	\section{Introduction}
    Aerial work is crucial in both daily life and industrial or agricultural production, as illustrated in Figure~\ref{F1}. Despite its significance, the low efficiency and high risks, including complicated workflow, falling from high places, electrocution from overhead power lines, and being trapped or squeezed, lead to numerous worker injuries or fatalities each year. Replacing workers with robots in dangerous aerial work environments can greatly enhance efficiency and safety, and potentially save lives.

    \begin{figure}[t]
		\centering
		\includegraphics[height=6.4cm,width=8.5cm]{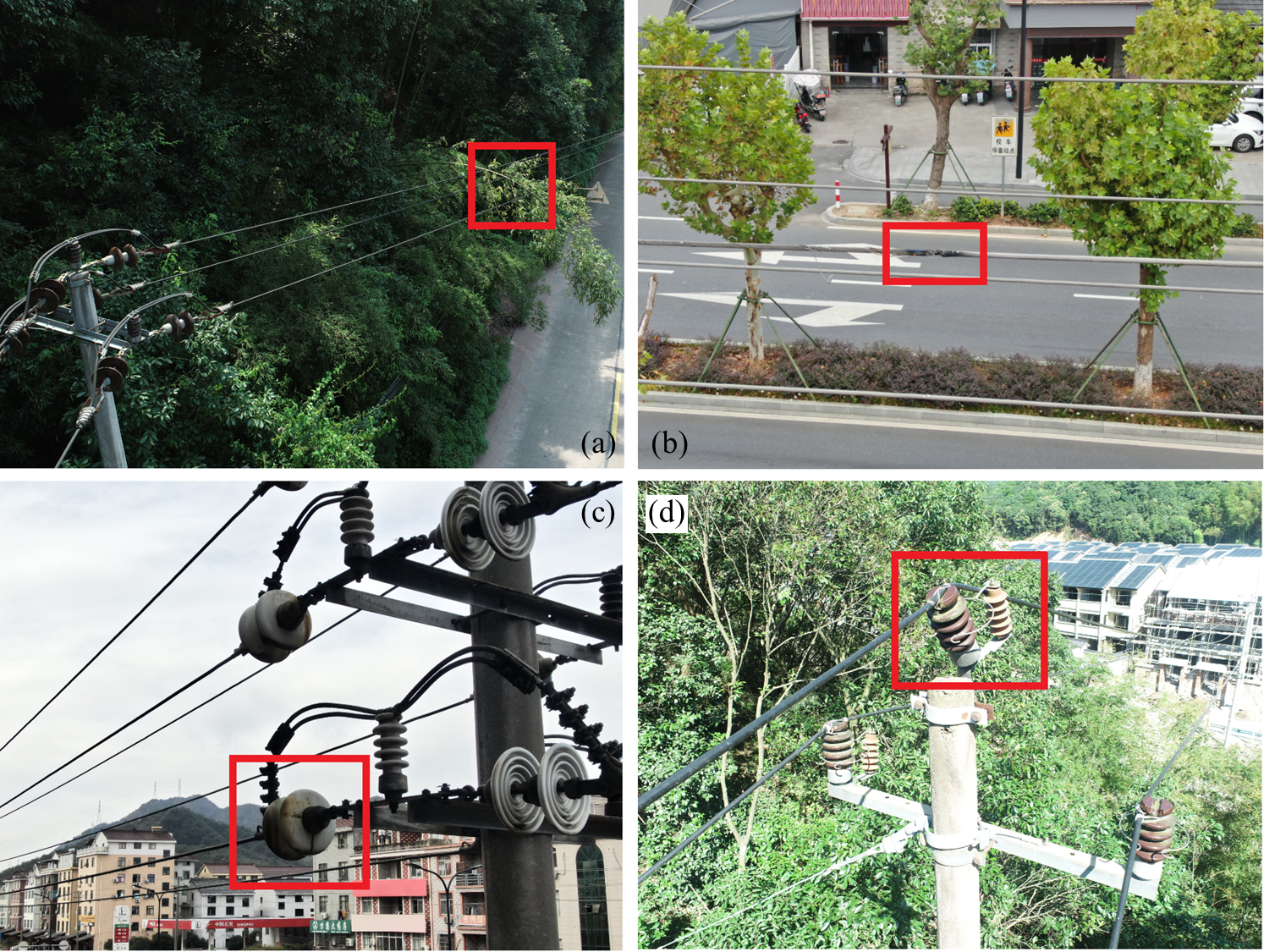}
		\caption{\revise{Typical power grid accidents:\\ (a) tree branch falling on the power line, (b) power line blown, (c) insulator contamination, (d) insulator tilt}}
		\label{Grid_problem}
    \end{figure}
    
    \revise{
    \subsection{Motivation and Challenges}    
    In August 2003, a massive blackout (Northeast Blackout of 2003) hit parts of the Northeastern and Midwestern United States and Ontario, Canada. The outage affected approximately 55 million people, caused \$25-30 billion in economic losses, and caused nearly 100 fatalities. And it all started with just a tree branch falling on a power line in Ohio, and the incorrect handling triggered a huge chain reaction.}

    \revise{As evident from the Northeast Blackout incident, it is crucial to address the challenges faced by the power grid to ensure a reliable and uninterrupted electricity supply. One solution that holds immense potential is the utilization of autonomous aerial work robots for performing maintenance and inspection tasks in the power grid.}

    \revise{What tasks would an aerial work robot system face when working in the power grid environment? And what challenges do these tasks reflect in terms of aerial work autonomy?}
    
    \revise{
    Some typical accidents in the power grid scenario are shown in Figure~\ref{Grid_problem}, such as tree branches falling onto power lines and power line blowing, which can pose a risk of power outages. Additionally, dirty insulators and tilted insulators can cause a decline in insulation performance, leading to power overload.
    }

    \revise{
    Therefore, starting with the tasks of aerial work robots in the power grid environment, which include inspection of power lines and equipment, identification, and removal of vegetation encroachments, reconnection of interrupted transmission lines, and replacement of dirty or tilted insulators, etc. The significant challenge that these robots must overcome is how to effectively perceive and interact with the complex aerial work environment.
    }
    
    \revise{
    In order to navigate and perform tasks in this environment, aerial robots require accurate and real-time localization, as well as comprehensive 3D perception and reconstruction capabilities through their visual and inertial systems. These capabilities are crucial for tasks such as object recognition, trajectory planning and control, and scene understanding. However, the aerial work environment and tasks present specific challenges for visual and sensors fusion localization and mapping. 
    }

    \revise{
    The presence of small-sized objects like tree branches, power lines, and trusses, combined with a lack of texture, makes their detection and reconstruction difficult. Unlike structured scenes such as warehouses or traffic, the aerial work environment is often cluttered and unstructured, making it challenging to rely on general structured features to improve algorithm performance. Furthermore, the aerial environment can be sparser than ground environments, making visual or LiDAR SLAM (Simultaneous Localization and Mapping) more challenging due to matching and loopback detection difficulties. With their increased degrees of freedom and exposure to sudden changes in motion,  aerial work robots face challenges related to complex motion and attitude estimation. 
    }

    \revise{
    Additionally, aerial work robots encounter similar difficulties as other outdoor robots, such as coping with a wide range of lighting conditions in different weather conditions. Direct sunlight or darkness at night can cause vision sensors to fail. To help overcome these challenges, the development of robust algorithms and the availability of appropriate datasets are crucial for advancing the capabilities of aerial work robots in the power grid environment. This is the motivation for us to make the USTC FLICAR dataset
    }
    
    \begin{figure}[t]
		\centering
		\includegraphics[height=3.19342cm,width=8.5cm]{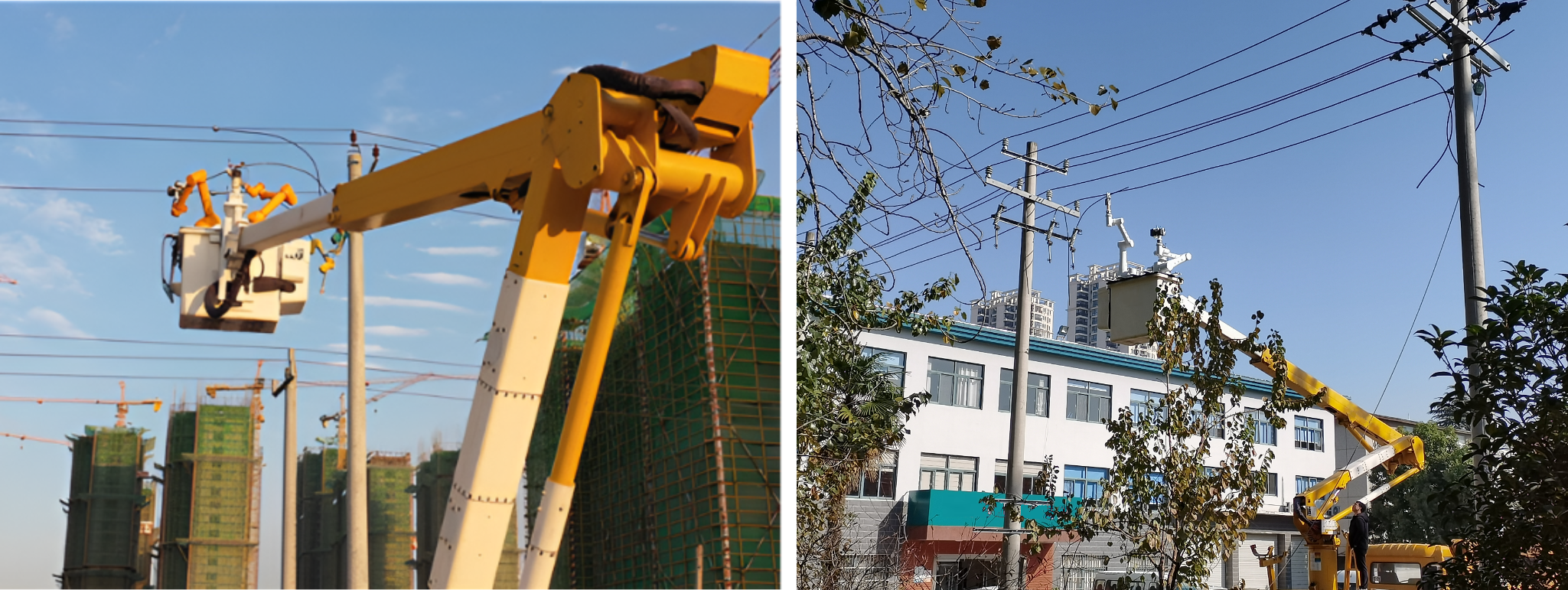}
		\caption{Aerial work robot for power grid tasks, equipped with two UR5 collaborative robot arms, lift into the air by bucket truck, developed by our lab. }
		\label{F2}
    \end{figure}

    \subsection{Public Dataset}
    Public datasets play a critical role in the implementation of autonomous systems in new scenarios. They provide a standard benchmark for evaluating algorithms, allowing for the quick investigation, verification, and development of algorithms without the need for expensive hardware, complicated calibration, and time-consuming data preparation. This section will review relevant datasets to the USTC FLICAR, which can be divided into two categories: ground and aerial. The details are summarized in Table~\ref{T1}.
    
    Ground scenarios, such as autonomous driving, have made significant progress in the past decade, partly due to the availability of diverse public datasets, such as the well-known KITTI dataset \cite{Geiger2013IJRR}. These datasets are remarkable for the abundance of sensors that can be incorporated because of the load-carrying capacity and space of ground vehicles. Ground scenarios rely on RTK-GPS or LiDAR SLAM for localization and mapping tasks, and generate ground truth values with centimeter-level accuracy.
    
    Aerial autonomous systems have also progressed in recent years. The EuRoC dataset \cite{burri2016euroc} was the first to equip UAVs with synchronously triggered high frame rate stereo cameras and IMUs, enabling a tightly coupled visual-inertial system for aerial robot localization and attitude estimation. The most commonly used platform for aerial data collection is micro or small drones, which are usually equipped with a few cameras and inertial sensors due to payload constraints. In indoor or small scenes, motion capture systems are used to generate millimeter-level ground truth, while in outdoor or larger scenes, laser trackers are used.

    \revise{However, it can be seen that there is a gap between ground and aerial datasets. The ground dataset comprises a large scene with rich sensors and provides centimeter-level accuracy for 2D motion ground truth. In contrast, the aerial dataset represents a smaller scene with limited sensors and offers millimeter-level accuracy for 3D motion ground truth. What if we combine the strengths of EuRoC and KITTI to create a new aerial dataset that incorporates both rich sensors and millimeter-level accuracy for 3D motion ground truth? This is where the USTC FLICAR dataset comes into play, bringing together the best aspects of both ground and aerial datasets.}

    Having rich sensors and highly accurate ground truth is essential for autonomous aerial working robots and systems. In the complex and unstructured aerial work environment, robots must be equipped with multiple complementary sensors to gather multi-modal data for perception. Moreover, as robots are expected to perform more precise tasks, such as precise interaction and manipulation of objects, millimeter-level accuracy ground truth becomes even more important.

    \begin{table*}[t]\scriptsize
		\sf\centering
		\caption{Summary and comparison of notable public datasets for ground and aerial autonomous systems.\label{T1}}
		\begin{tabular}{lllllll}
			\toprule
			& &Sensors& & & &  \\
			\cmidrule{2-4}
			Dataset&IMU&Camera&LiDAR&Gruond Truth&Environment&Platform\\
			\midrule
			\makecell[l]{MIT DARPA\\\cite{huang2010high}} & N/A & \makecell[l]{5 PointGrey Firefly MV:\\376$\times$240$\times$4/752$\times$480} & \makecell[l]{3D Velodyne HDL-64E\\2D SICK LMS 291-S05\\$\times$12} &RTK GPS/INS& \makecell[l]{Outdoor\\(Urban)}& Car\\
			\makecell[l]{Ford Campus\\\cite{pandey2011ford}} & \makecell[l]{6 axis Xsens \\MTi-G} & \makecell[l]{PointGrey LadyBug 3:\\1600$\times$600$\times$6} & \makecell[l]{3D Velodyne HDL-64E\\2D Riegl LMS$\times$2} &RTK GPS/INS&\makecell[l]{Outdoor\\(Urban)}& Car\\
			\makecell[l]{KITTI\\\cite{Geiger2013IJRR}} &\makecell[l]{6 axis OXTS\\ RT3003} & \makecell[l]{4 PointGrey FL2-14S3M/C-C :\\1392$\times$512$\times$4} & \makecell[l]{Velodyne HDL-64E} &RTK GPS/INS& \makecell[l]{Outdoor\\(Urban)} & Car\\
			\makecell[l]{NCLT\\\cite{carlevaris2016university}} & \makecell[l]{9 axis\\3DM-GX3-45} & \makecell[l]{PointGrey LadyBug 3:\\1600$\times$1200$\times$6} & \makecell[l]{3D Velodyne HDL-64E\\2D Hokuyo$\times$2} & \makecell[l]{RTK GPS/\\LiDAR SLAM} & \makecell[l]{Outdoor\\(Campus)}& \makecell[l]{Ground\\Robot}\\
			\makecell[l]{Oxford RobotCar\\\cite{maddern20171}} &\makecell[l]{6 axis NovAtel\\ SPAN-CPT ALIGN} & \makecell[l]{ PointGrey Bumblebee XB3:\\1280$\times$960$\times$3\\3 Point Grey Grasshopper2:\\1024$\times$1024} & \makecell[l]{2D SICK LMS-151$\times$2\\3D SICK LD-MRS} &RTK GPS/INS& \makecell[l]{Outdoor\\(Urban)}& Car\\
			\makecell[l]{Oxford Radar\\ RobotCar\\\cite{RadarRobotCarDatasetArXiv}} &\makecell[l]{6 axis NovAtel\\ SPAN-CPT ALIGN} & \makecell[l]{PointGrey Bumblebee XB3:\\1280$\times$960$\times$3\\3 PointGrey Grasshopper2:\\1024$\times$1024$\times$3} & \makecell[l]{3D Velodyne HDL-32E\\$\times$2} &RTK GPS/INS&\makecell[l]{Outdoor\\(Urban)}& Car\\
			\makecell[l]{Rosario\\\cite{pire2019rosario}} &\makecell[l]{6 axis LSM6DS0} & \makecell[l]{ZED stereo:\\672$\times$376$\times$2} & N/A &RTK GPS/INS& \makecell[l]{Outdoor\\(Agriculture)}& \makecell[l]{Ground\\ Robot}\\
			\makecell[l]{KAIST Urban\\\cite{jeong2019complex}} &\makecell[l]{9 axis Xsens\\ MTi-G-300} & \makecell[l]{ FLIR FL3-U3-20E4C-C:\\ 1280$\times$560$\times$2} & \makecell[l]{3D Velodyne VLP-16C\\$\times$2} & SLAM  &\makecell[l]{Outdoor\\(Urban)}& Car\\
			\makecell[l]{EU Long-term\\\cite{yan2020eu}} &\makecell[l]{9 axis Xsens\\ MTi-28A53G25} & \makecell[l]{PointGrey Bumblebee XB2/3\\2 Pixelink PL-B742F} & \makecell[l]{3D Velodyne HDL-32E$\times$2\\2D SICK LMS} & RTK-GPS &\makecell[l]{Outdoor\\(Urban)}& Car\\
			\makecell[l]{nuScenes\\\cite{caesar2020nuscenes}} &\makecell[l]{9 axis Advanced \\ Navigation Spatial} & \makecell[l]{6 Basler acA1600-60gc:\\1600$\times$1200$\times$6} & \makecell[l]{3D Velodyne HDL-32E} &RTK GPS/INS&\makecell[l]{Outdoor\\(Urban)}& Car\\
			\midrule
			\makecell[l]{EuRoC\\\cite{burri2016euroc}}&\makecell[l]{6 axis ADIS16448} & \makecell[l]{2 MT9V034: 752$\times$480$\times$2} & N/A &\makecell[l]{6DOF MoCap\\3D Laser Tracker}& Indoor& UAV\\
			\makecell[l]{Zurich Urban\\\cite{majdik2017zurich}}  &\makecell[l]{6 axis on PX4\\autopilot board} & \makecell[l]{GoPro Hero 4: 1920$\times$1080} & N/A  &\makecell[l]{Aerial-\\Photogrammetry\\Visual SLAM}& \makecell[l]{Outdoor\\(Urban)} & UAV\\
			\makecell[l]{UZH-FPV\\\cite{delmerico2019we}} &\makecell[l]{6 axis IMU integrated\\with the camera} & \makecell[l]{Snapdragon Fisheye Stereo:\\640$\times$480$\times$2\\mDAVIS Event: 346$\times$260} & N/A  &\makecell[l]{3D Laser Tracker}& 	\makecell[l]{Indoor\\Outdoor}& UAV\\
			\makecell[l]{NTU VIRAL\\\cite{nguyen2022ntu}} &\makecell[l]{9 axis VectorNav\\ VN100} & \makecell[l]{2 uEye 1221 LE:\\752$\times$480$\times$2} & \makecell[l]{3D Ouster OS1-16$\times$2}  &\makecell[l]{3D Laser Tracker}&\makecell[l]{Outdoor\\(Campus)}& UAV\\
			\midrule
			\makecell[l]{USTC FLICAR}&\makecell[l]{9 axis Xsens \\MTi-G-710} & \makecell[l]{ PointGrey Bumblebee XB3:\\1280$\times$960$\times$3\\PointGrey Bumblebee XB2:\\1024$\times$768$\times$2\\Hikvision MV-CB016-10GC\\1440$\times$1080\\Hikvision MV-CE060-10UC\\3072$\times$2048} & \makecell[l]{3D Velodyne HDL-32E\\3D Velodyne VLP-32C\\3D Ouster OS0-128\\3D LiVOX Avia} & \makecell[l]{3D Laser Tracker}& \makecell[l]{Outdoor\\(Urban/\\Aerial)}& \makecell[l]{Bucket\\Truck/\\ Ground\\Robot}\\
			\bottomrule
		\end{tabular}\\
		
	\end{table*}

    \subsection{Related Works: Bridge Ground-Aerial Gap}
	
	In the previous section, we compared two of the most prominent datasets in the ground and aerial domains: KITTI and EuRoC. We identified the gap between these datasets and proposed the creation of a new aerial dataset that features rich sensor data and 3D motion ground truth with millimeter-level accuracy. To achieve this goal, there are two options: adding sensors to a drone with a larger payload or enabling a car to ``fly" for 3D movement. 
 
    The NTU VIRAL dataset \cite{nguyen2022ntu} chose the first option by equipping a larger UAV with two Ouster 16-beam 3D LiDARs, two cameras, and an IMU. It is the pioneer in applying LiDARs to drone scenarios.
 
    The second option, which may seem like a fantasy, is the approach taken in the USTC FLICAR dataset. The USTC FLICAR dataset plans to extend the typical autonomous driving sensor suite to aerial scenes, with a large multi-sensor platform that includes four LiDARs (one Ouster 128-beam, two Velodyne 32-beam, one MEMS LiDAR), seven cameras (two stereo pairs), and three IMUs/INS. If we choose the first option like NTU VIRAL, only a few very expensive large drones, similar in size to helicopters, may have enough carrying capacity. However, it is nearly impossible and highly dangerous to fly such large aircraft equipped with a working robot at \revise{ultra-low} altitudes (below 50m) in cities for aerial operations and data collection, due to current regulations and safety concerns. Luckily, we found the bucket truck.
    
     \revise{A} Bucket truck is a type of construction machinery with \revise{a} high retention rate and \revise{is} widely used in aerial work. The large \revise{pieces of  machinery} in various aerial work scenarios in Figure~\ref{F1} are bucket trucks. They are heavy-duty construction vehicles equipped with an extendable hydraulic arm that carries a large bucket to elevate workers to elevated or inaccessible areas. These trucks have a strong payload capacity of around 200kg and can reach any target position in their 3D \revise{workspace} through arm extension and joint rotation. With the help of a bucket truck, it is feasible to allow a heavy multi-sensor perception platform for autonomous driving to perform flexible 3D motion within a certain range in the air. With current technology, we cannot make a car fly, but we can start by making its sensing part fly.
    
    When designing the sensor suite, we aimed to keep it as similar as possible to existing autonomous driving datasets. For example, our Bumblebee stereo camera is the same as those used in the Oxford RobotCar \cite{maddern20171} and EU Long-term \cite{yan2020eu} datasets. Our Velodyne HDL-32E horizontal 3D LiDAR is also the same as those used in the nuScenes \cite{caesar2020nuscenes}, Oxford Radar RobotCar \cite{RadarRobotCarDatasetArXiv}, EU Long-term, and NCLT \cite{carlevaris2016university} datasets. This makes it more convincing to compare algorithms using the same hardware between these datasets and the USTC FLICAR dataset. The USTC FLICAR dataset also shares similarities with existing aerial datasets but is geared \revise{toward} more delicate and heavy-duty tasks. Millimeter-level outdoor ground truth was obtained using a laser tracker.

    Looking back at Table~\ref{T1}, it is clear that USTC FLICAR has the most sensors among all aerial datasets. We believe that our dataset is a significant contribution that provides benchmarks for testing existing algorithms for autonomous systems and developing new ones that are more suited to the particularities of aerial work scenes.

     In addition, the USTC FLICAR dataset and experiments will demonstrate that the innovative combination of an autonomous driving sensing suite and a bucket truck results in a versatile autonomous aerial platform with significant potential. This platform has exceptional perception capabilities that are comparable to those of self-driving cars and can be further equipped with a variety of working tools, such as robotic arms, to effectively carry out diverse and heavy-duty aerial working tasks. A combined prototype of the platform is illustrated in Figure~\ref{F2}.

    \subsection{Paper Roadmap}
	The remaining parts of the paper are organized as follows: In Section~\ref{section2}, we provide a detailed description of the various components of the data acquisition systems, as well as the parameters, characteristics, and functions of each sensor. Section~\ref{section3} outlines the specific content of the dataset, the format of data storage, and the methods for accessing the data. Section~\ref{section4} discusses the time synchronization, intrinsic and extrinsic calibration of sensors, and the generation of ground truth. We also present an evaluation of some state-of-the-art SLAM algorithms on the dataset as baselines and analyze the results in Section~\ref{section5}. In Section~\ref{section6}, a comprehensive description is provided regarding the creation and contents of the Semantic FLICAR dataset. Section~\ref{section7} highlights the known issues that users need to be mindful of when utilizing the dataset. Finally, Section~\ref{section8} summarizes the paper and discusses future work.

 	\begin{figure}[t]
		\centering
		\includegraphics[height=4.9cm,width=8.5cm]{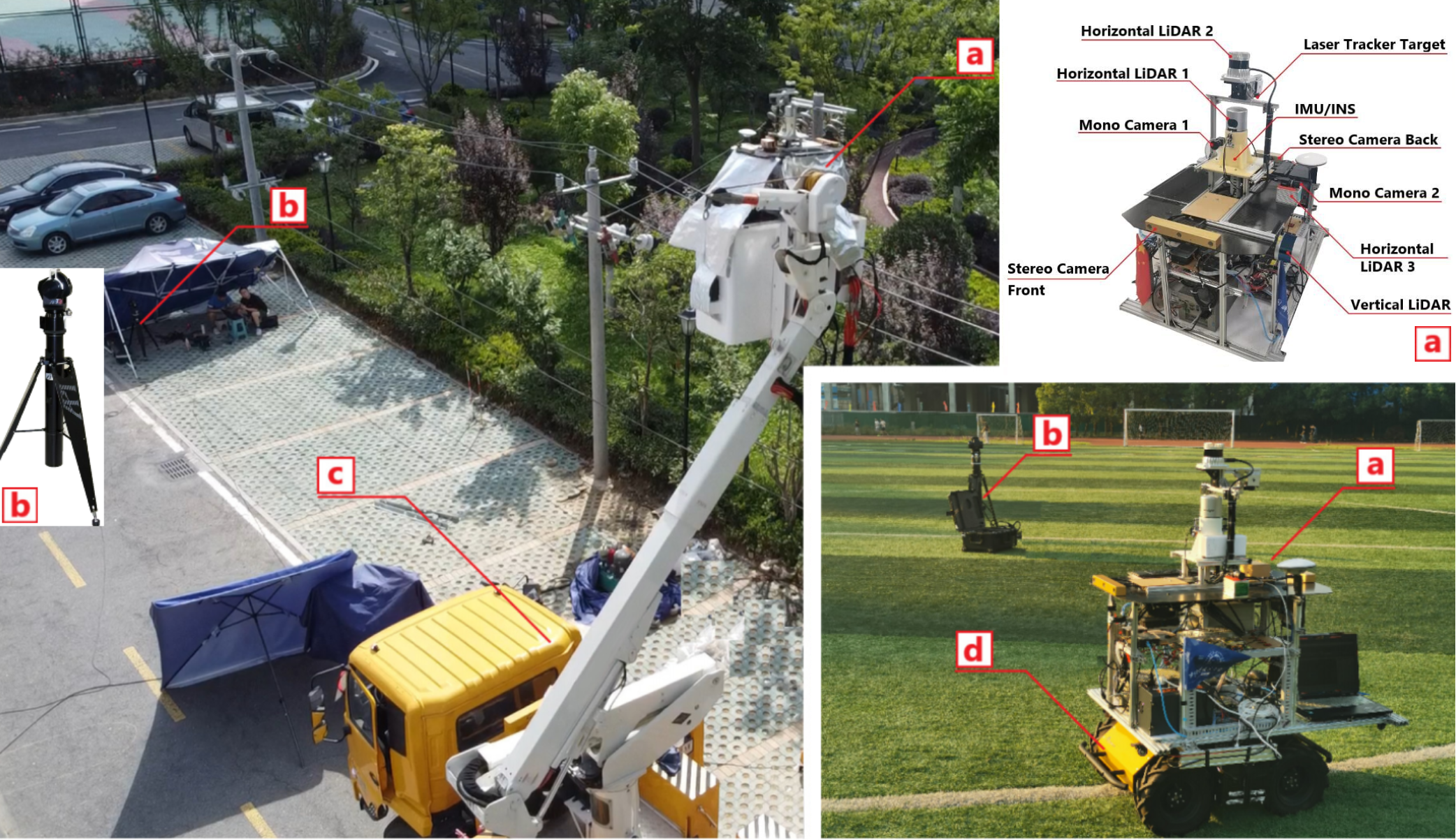}
		\caption{``Giraffe" and ``Okapi" acquisition systems:\\``Giraffe" aerial system: (a), (b) and (c).\\``Okapi" ground system: (a), (b) and (d)\\(a) multi-sensor data collection platform (Fig. ~\ref{System}), (b) laser tracker ground truth system, (c) bucket truck, (d) ground robot.}
		\label{F3}
	\end{figure}
 
	\section{Acquisition Systems and Sensor Setup} \label{section2}
	
	The data was acquired using two different systems, the ``\revise{Giraffe}" and ``Okapi" systems, as depicted in Figure~\ref{F3}. The ``Giraffe" system is an aerial platform consisting of a multi-sensor data collection platform (a), a laser tracker ground truth system (b), and a bucket truck (c). On the other hand, the ``Okapi" system is a ground-based system similar to an autonomous vehicle, equipped with the same sensors (a) and a ground truth recording system (b), and mounted on a ground robot (d) for the acquisition of ground-level data for comparison with the data collected by the aerial system. Section~\ref{section2} will provide more detailed information about the components of the data acquisition systems.
     
     \begin{figure*}[t]
		\centering
		\includegraphics[height=9.7cm,width=16cm]{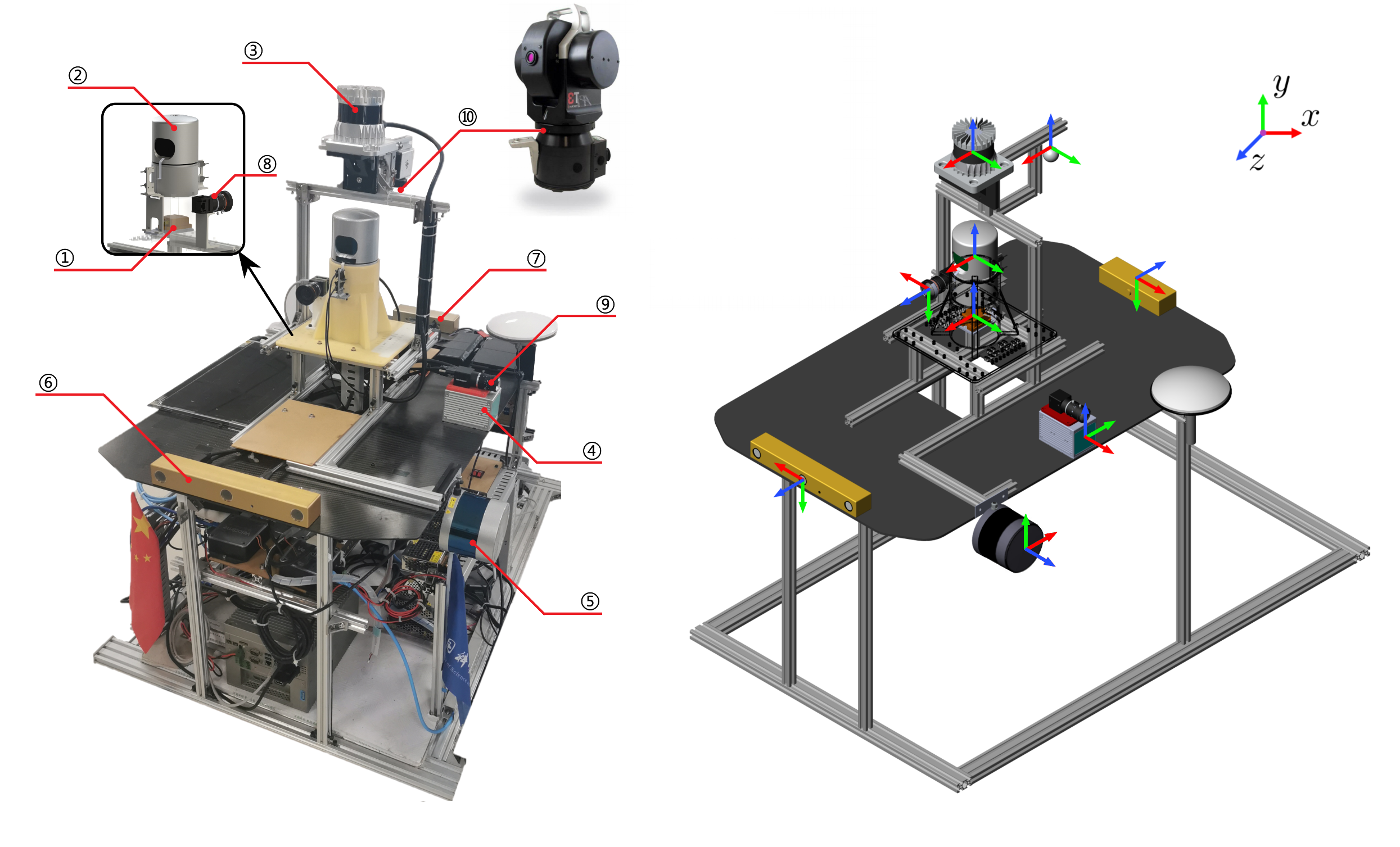}
		\caption{Sensor setup on a multi-sensor platform. The coordinate system shows the origin and orientation of each sensor mounted on the vehicle, with the following convention: X (red), Y (green), \revise{and} Z (blue). Note that Velodyne LiDAR has two different coordinate system conventions. The coordinate system in the product manual is: Y: forward X: right Z: up; and the default coordinate system of \revise{the} Velodyne ROS package is: X: forward Y: left Z: up. Here shows the second coordinate system of \revise{the} ROS package. The number of the sensor in the figure corresponds to Table~\ref{T2}.}
		\label{System}
	\end{figure*}
 
	\begin{figure*}[t]
		\centering
		\includegraphics[height=7.5459cm,width=17cm]{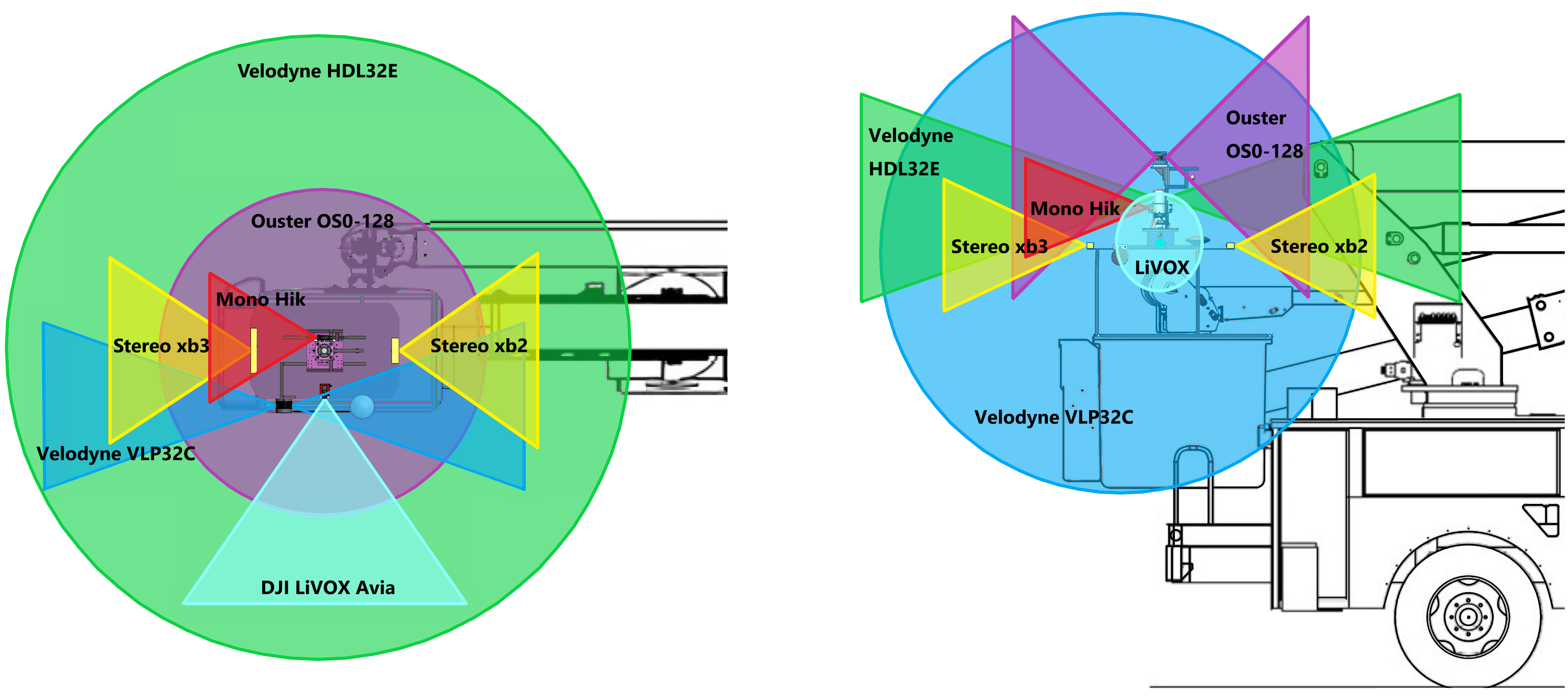}
		\caption{The visual scope of the acquisition system sensors, top view\revise{,} and side view.}
		\label{fov_all}
	\end{figure*}

   \begin{figure*}[t]
	\centering
	\includegraphics[height=9.8685cm,width=16cm]{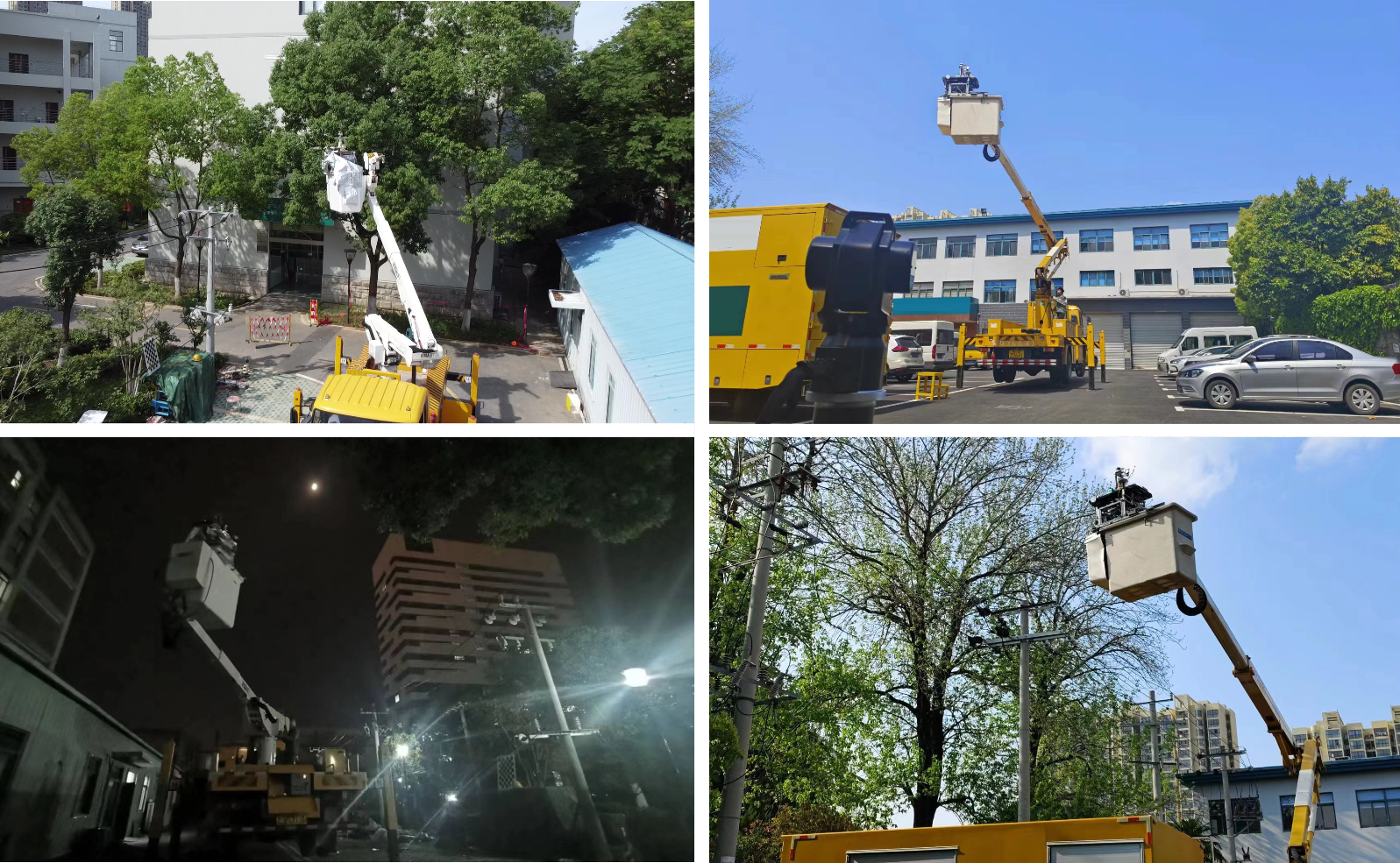}
	\caption{Images of several typical data acquisition sites. \revise{Left: HF0XX sites, Right: HN0XX sites.}}
	\label{from_day_to_night}
 \end{figure*}
	
	Both Systems are equipped with the following sensors:
	
	\subsection{Inertial Measurement Unit (IMU)}
The main IMU in the acquisition system is an Xsens MTi-G-710 INS/GNSS module, which is installed at the center of the system. The body frame is aligned with the Xsens sensor frame.

• 1 × Xsens MTi-G-710 INS/GNSS, 9-axis, 400 Hz, accuracy: $0.2^{\circ}$ in roll/pitch, $0.8^{\circ}$ in heading.

The Xsens module outputs the three-axis acceleration and three-axis angular velocity in its own coordinate system, as well as the quaternion attitude in the North-East-Down (NED) coordinate system. The Xsens module is hardware-synchronized to the same extrinsic GPS clock source with the cameras and LiDARs in the system, forming visual-inertial and LiDAR-inertial sensor units together. In addition, two extra 6-axis IMUs are installed in the OS0-128 LiDAR and LiVOX Avia LiDAR, forming part of the LiDAR-Inertial sensor unit.
	
	\subsection{3D LiDARs}
	A LiDAR (Light Detection and Ranging) is a remote sensing method that uses light in the form of a pulsed laser to measure distances to objects and create a 3D map of the environment. LiDARs are essential for autonomous systems to accurately perceive and understand 3D scenes. The system in this work includes four different LiDARs: one digital LiDAR, two mechanical LiDARs, and one MEMS LiDAR.

	• Digital LiDAR: 1 × Ouster OS0-128, 10 Hz, 128 beams, $0.7^{\circ}$ angular resolution, $\pm$ 1.5 to $\pm$ 5 cm distance accuracy, collecting 2.62 million points/second, field of view: $360^{\circ}$ HFoV, $90^{\circ}$ VFoV ($\pm 45^{\circ}$), range:50 m
	
	• Mechanical LiDAR: 1 × Velodyne HDL-32E, 5/10 Hz, 32 beams,  $1.33^{\circ}$ angular resolution, $\pm$ 2 cm distance accuracy, collecting 1.39 million points/second, \revise{the} field of view: $360^{\circ}$ HFoV, $41.3^{\circ}$ VFoV ($+10.67^{\circ}$ to $-30.67^{\circ}$), range: 100 m
	
	• Mechanical LiDAR: 1 × Velodyne VLP-32C, 10 Hz, 32 beams, $ 0.33^{\circ}$ angular resolution (non-linear distribution), $\pm$3 cm distance accuracy, collecting 1.20 million points/second, \revise{the} field of view:  $360^{\circ}$ HFoV, $40^{\circ}$ VFoV ($-25^{\circ}$ to $+15^{\circ}$), range: 200 m
	
	• MEMS LiDAR: 1 × DJI LiVOX Avia, 10 Hz, 2 cm distance accuracy, collecting 0.24 million points/second, \revise{the} field of view: $70.4^{\circ}$ HFoV, $77.2^{\circ}$ VFoV (Non-repetitive Scanning), range: 450 m

The system uses Ouster OS0-128 and Velodyne HDL-32E LiDARs as the main source of 3D environmental data, and Velodyne VLP-32C and DJI LiVOX Avia LiDARs to supplement the blind areas of vision. The Ouster OS0-128 is a digital LiDAR based on custom system-on-a-chip with single-photon avalanche diode detectors, which can output \revise{point clouds}, depth images, and signal-intensity images of LiDAR and visible light spectrum (Figure~\ref{datashow_num} (c),(d)). Velodyne HDL-32E has the highest point cloud accuracy, while Velodyne VLP-32C and DJI LiVOX Avia LiDARs are mounted vertically and horizontally, respectively, to provide a 360-degree view in both horizontal and vertical directions. The \revise{point clouds} from LiVOX LiDAR scans are uniformly accumulated on the map over time.

	\subsection{Monocular and Stereo Cameras}
	Cameras are an important part of autonomous system perception, as they capture high-resolution images of the surrounding environment, providing information about object shape, color, texture, and motion direction. And stereo cameras can effectively recover depth. We equipped the sensor platform with several cameras, including stereo cameras and monocular cameras:
	
	• 1 x Point Grey Bumblebee XB3 (BBX3-13S2C-38) trinocular stereo camera, 1280 $\times$ 960 $\times$ 3, 10Hz, Sony ICX445 CCD, 1/3", 3.75 µm, global shutter, 3.8mm lens, $66^{\circ}$ HFoV, 12/24cm baseline, IEEE 1394B, 54 dB Signal To Noise Ratio (SNR).
	
	• 1 x Point Grey Bumblebee XB2 (BBX2-08S2C-38) binocular stereo camera, 1024 x 768 × 2, 10-15Hz, Sony ICX204 CCD, 1/3", 4.65 µm, global shutter, 3.8mm lens, $70^{\circ}$ HFoV, 12cm baseline, IEEE 1394A, 60 dB SNR.
	
	• 1 x Hikvision MV-CB016-10GC-C industrial monocular camera, 1440 × 1080, 20Hz, Sony IMX296 CCD, 1/2.9", 3.45 µm, global shutter, 6mm lens (MVL-HF0628M-6MPE), $63.11^{\circ}$ HFoV, GigE, 41 dB SNR. 
	
	• 1 x Hikvision MV-CE060-10UC industrial monocular camera, 3072 × 2048, 20Hz, Sony IMX178 CCD, 1/1.8", 2.4 µm, global shutter, 6mm lens (MVL-HF0628M-6MP), $49.3^{\circ}$ HFoV, USB 3.0, 41.3 dB SNR. 

    Stereo cameras use two or more cameras to capture the same scene from different viewpoints, allowing the system to estimate depth information. On the other hand, monocular cameras use a single camera and rely on other techniques like motion parallax and perspective to estimate depth. Both types of cameras have their advantages and disadvantages, and their usage depends on the specific application and requirements.

	\subsection{Laser Tracker}
	A laser tracker is a high-precision instrument used to accurately measure the position and orientation of objects in 3D space. It works by emitting a laser beam and tracking a target that reflects the laser light back to the tracker. The laser tracker measures the time it takes for the laser beam to travel to the target and back and uses this information to calculate the position and orientation of the target. The API T3 laser tracker we use is fixed horizontally on the ground and is the only sensor that is independent of the multi-sensor platform, as shown in Figure~\ref{F3} (b). During the motion of the sensor platform, the laser tracker tracks the active target ball that \revise{is} rigidly fixed on the platform body and outputs the three-dimensional space coordinates of the target point with millimeter precision in its own coordinate system. 
	
	• API T3 Laser Tracker, 50Hz, azimuth: $\pm$ $320^{\circ}$ ($640^{\circ}$ end to end), angular resolution: $\pm$ 0.018 arc-seconds, angular accuracy: 3.5 µm/meter, system resolution: 0.1 µm, maximum lateral target speed:  4 meters/sec, maximum acceleration:  2 g, internal level accuracy: $\pm$ 2 arc-second, linear range: 80 m.

	\section{Dataset} \label{section3}

 \begin{table*}[t]
	\footnotesize\sf\centering
	\caption{The sensor model specifications and data information in this dataset.\label{T2}}
	\begin{tabular}{llllll}
		\toprule
		No&Sensor& Model&ROS Topic& Message type&Rate\\
		\midrule
		\texttt 1 & IMU/INS & Xsens MTi-G-710 & /imu/data & sensor\_msgs/Imu & 400Hz\\
		\midrule
		\texttt 2 & Horizontal LiDAR 1 & Velodyne HDL-32E & /velodyne\_points\_HDL32 & sensor\_msgs/PointCloud2 & \makecell[l]{5/10Hz\\(rotate at 10Hz)}\\
		\midrule
		\texttt 3 & Horizontal LiDAR 2 & Ouster OS0-128 & \makecell[l]{/os\_cloud\_node/imu \\ /os\_cloud\_node/points \\ /img\_node/reflect\_image \\ /img\_node/signal\_image} & \makecell[l]{sensor\_msgs/Imu \\ sensor\_msgs/PointCloud2 \\ sensor\_msgs/Image \\ sensor\_msgs/Image} & \makecell[l]{100Hz \\ 10Hz \\ 10Hz \\ 10Hz}\\
		\midrule
		\texttt 4 & Horizontal LiDAR 3 & LiVOX Avia & \makecell[l]{/livox/lidar \\ /livox/imu} & \makecell[l]{livox\_ros\_driver/CustomMsg \\ sensor\_msgs/Imu} & \makecell[l]{10Hz \\ 200Hz}\\
		\midrule
		\texttt 5 & Vertical LiDAR 1 & Velodyne VLP-32C & /velodyne\_points\_VLP32 & sensor\_msgs/PointCloud2 & 10Hz\\
		\midrule
		\texttt 6 & Stereo Camera front & PointGrey Bumblebee xb3 &\makecell[l]{/camera/left/image\_raw \\ /camera/center/image\_raw \\ /camera/right/image\_raw} & sensor\_msgs/Image & \makecell[l]{10-16Hz \\ 10-16Hz \\ 10-16Hz}\\
		\midrule
		\texttt 7 & Stereo Camera back & PointGrey Bumblebee xb2 &\makecell[l]{/cam\_xb2/left/image\_raw \\ /cam\_xb2/right/image\_raw} & sensor\_msgs/Image & \makecell[l]{10-20 Hz \\ 10-20 Hz}\\
		\midrule
		\texttt 8 & Mono Camera 1 &Hikvision MV-CB016-10GC-C & /hik\_camera/iamge\_raw & sensor\_msgs/Image & 20Hz\\
		\midrule
		\texttt 9 & Mono Camera 2 &Hikvision MV-CE060-10UC & /right\_camera/iamge & sensor\_msgs/Image & 20Hz\\
		\bottomrule
	\end{tabular}\\[10pt]
		
\end{table*}

\begin{table*}[t]
	\footnotesize\sf\centering
	\caption{\revise{Data sequences details and summary}
    \label{T_data_info}}
	\begin{tabular}{lllllll}
		\toprule
		Sequence & Time & Length (m) & Duration (s) & Size (GB) & Characteristic & Summary\\
		\cmidrule{1-7}
		HF001 & 2022-06-08 15:35 & 26.46 & 192 & 66.5 & sun, semantic & \multirow{3}{*}{\scriptsize \makecell[l]{Conducting data collection in a \\representative aerial work environment \\on the urban roadside, surrounding \\objects include utility poles, trees, \\buildings, etc. The collection\\ environment involves significant lighting\\ variations from daytime to dusk to nighttime.\\ Additionally, the keyframes in the data are\\ annotated with semantic segmentation. \\HF001-HF009 are the main sequences \\ in this dataset.}}\\
		HF002 & 2022-06-08 15:48 & 33.50 & 217 & 75.7 & sun, semantic & \\
        HF003 & 2022-06-08 16:14 & 34.26 & 217 & 83.2 &sun, rotation, semantic & \\
        HF004 & 2022-06-08 16:30 & 24.10 & 155 & 82.0 &sun, semantic & \\
        HF005 & 2022-06-08 17:20 & 22.82 & 260 & 90.3 &sun, semantic & \\
        HF006 & 2022-06-08 18:13 & 33.90 & 230 & 86.3 &clouds, semantic & \\
        HF007 & 2022-06-08 19:01 & 34.32 & 207 & 67.5 &dusk, rotation,semantic & \\
        HF008 & 2022-06-08 21:20 & 30.78 & 210 & 91.3 &night, rotation, semantic & \\
        HF009 & 2022-06-08 21:27 & 35.42 & 238 & 101.3 &night, rotation, semantic & \\
         &  &  &  &  &  \\
        \midrule
        HF010 & 2022-06-07 18:19 & 16.06 & 210 & 91.7 &clouds & \multirow{3}{*}{\scriptsize \makecell[l]{Additional Sequences Group I:\\Data collection environment same to the \\ main sequence, except for the existing sensors,\\ an extra down-looking camera mounted\\ with a vertical lidar Velodyne VLP32 access-\\ible via  ROS topic ``\texttt{/down\_camera/image}"}}\\
		HF011 & 2022-06-07 19:13 & 17.81 & 207 & 25.5 &dusk & \\
        HF012 & 2022-06-07 19:26 & 26.15 & 231 & 121.1 &dusk & \\
        HF013 & 2022-06-07 20:40 & 26.25 & 187 & 100.9 &night & \\
        HF014 & 2022-06-07 20:54 & 25.57 & 201 & 119.2 &night & \\
        \midrule
        HN001 & 2023-04-07 15:53 & 38.44 & 390 & 79.2 & sun, rotation, semantic & \multirow{3}{*}{\scriptsize \makecell[l]{Additional Sequences Group II:\\Collecting data in a new parking lot environment,\\ which includes not only buildings, poles\\ and trees but also numerous vehicles. \\An extra FLIR ADK infrared camera is installed, \\and it can be accessed via the ROS topic \\``\texttt{/flir\_boson/image\_raw}". \\Bumblebee XB3 and LiVOX are offline.}}\\
		HN002 & 2023-04-07 12:12 & 44.97 & 395 & 56.1 & sun, rotation, semantic & \\
        HN003 & 2023-04-07 12:23 & 38.64 & 442 & 62.2 & sun, rotation & \\
        HN004 & 2023-04-07 13:02 & 42.50 & 417 & 59.1 & sun, rotation & \\
         &  &  &  &  &  \\
         &  &  &  &  &  \\
         &  &  &  &  &  \\
        \midrule
        GR001 & 2022-06-15 15:50 & 46.51 & 333 & 35.9 & sun, turn & \multirow{3}{*}{\scriptsize \makecell[l]{Additional Sequences Group III:\\Collecting ground data using multi-sensor platform \\mounted on a Husky chassis. The data collection \\involves capturing different motion patterns \\compared to aerial scenarios, such as 2D movements,\\ turns, etc. Bumblebee XB3 and Ouster are offline.}}\\
		GR002 & 2022-06-15 19:19 & 66.89 & 116 & 12.1 & dusk, turn & \\
        GR003 & 2022-06-15 19:24 & 40.59 & 190 & 19.5 & dusk, turn & \\
        GR004 & 2022-06-15 19:34 & 69.16 & 211 & 25.7 & dusk, turn & \\  
         &  &  &  &  &  \\
        \bottomrule
	\end{tabular}\\[10pt]
\end{table*}

	\begin{figure}[t]
		\centering
		\includegraphics[height=6.967cm,width=8.5cm]{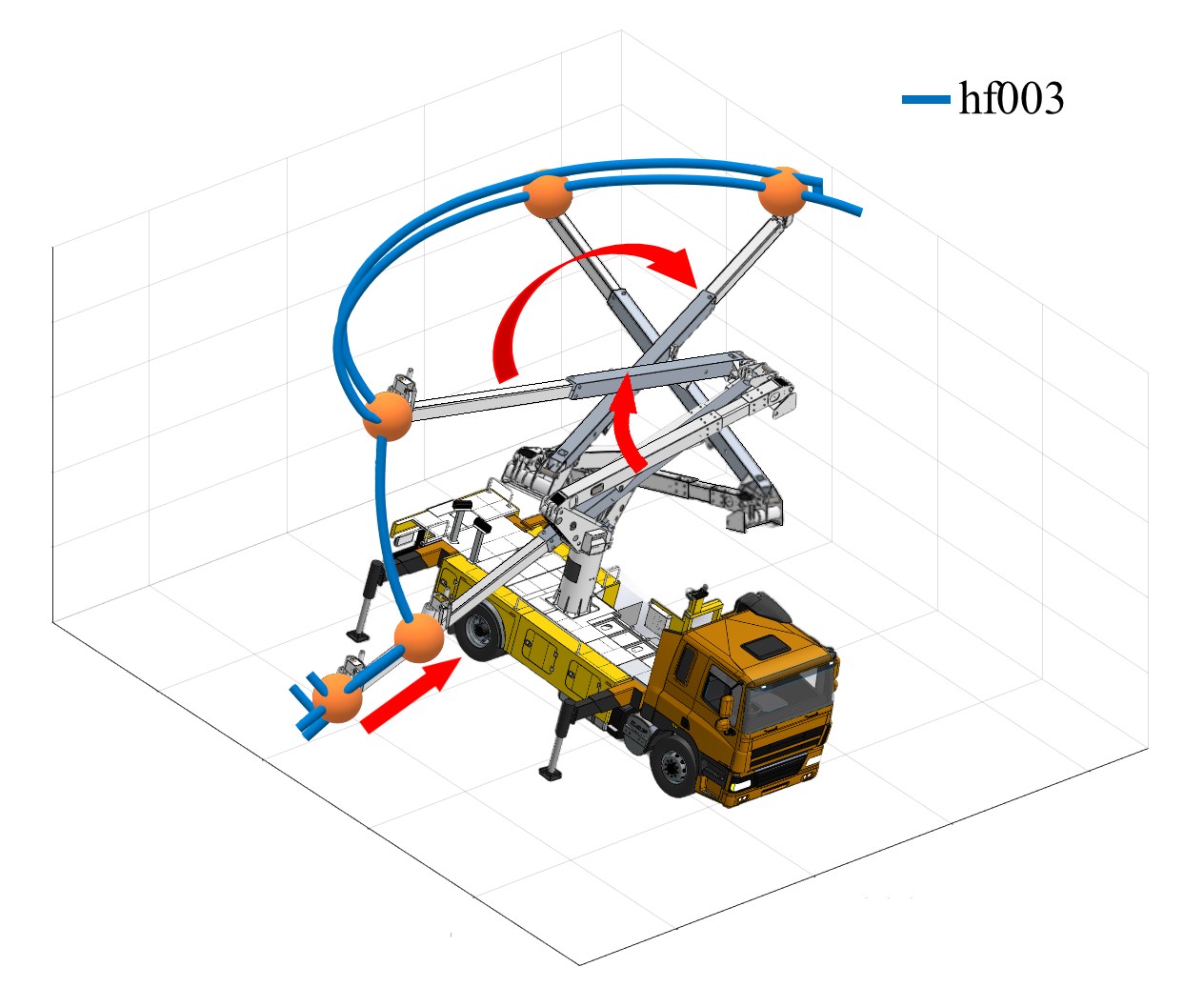}
		\caption{The joint action of the hydraulic arm of the bucket truck corresponds to the movement trajectory of the end sensor platform, taking the hf003 sequence as an example.}
		\label{doubi_shuoming}
	\end{figure}

     \begin{figure}[t]
		\centering
		\includegraphics[height=8.73856cm,width=8.5cm]{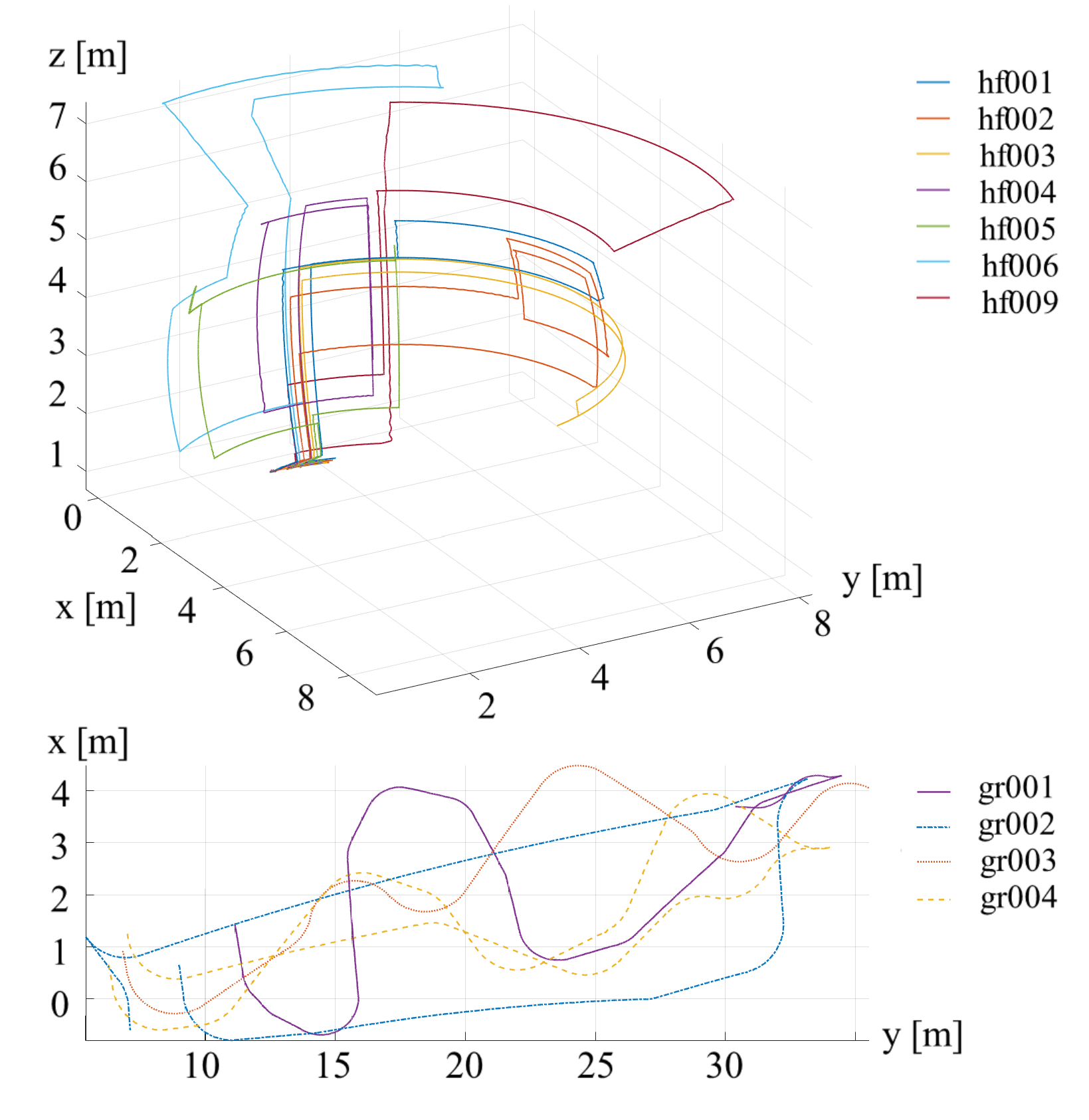}
		\caption{Representative aerial and ground trajectories in the dataset.}
		\label{all_p_new}
	\end{figure}

     \begin{figure}[t]
		\centering
		\includegraphics[height=9.04134cm,width=8.5cm]{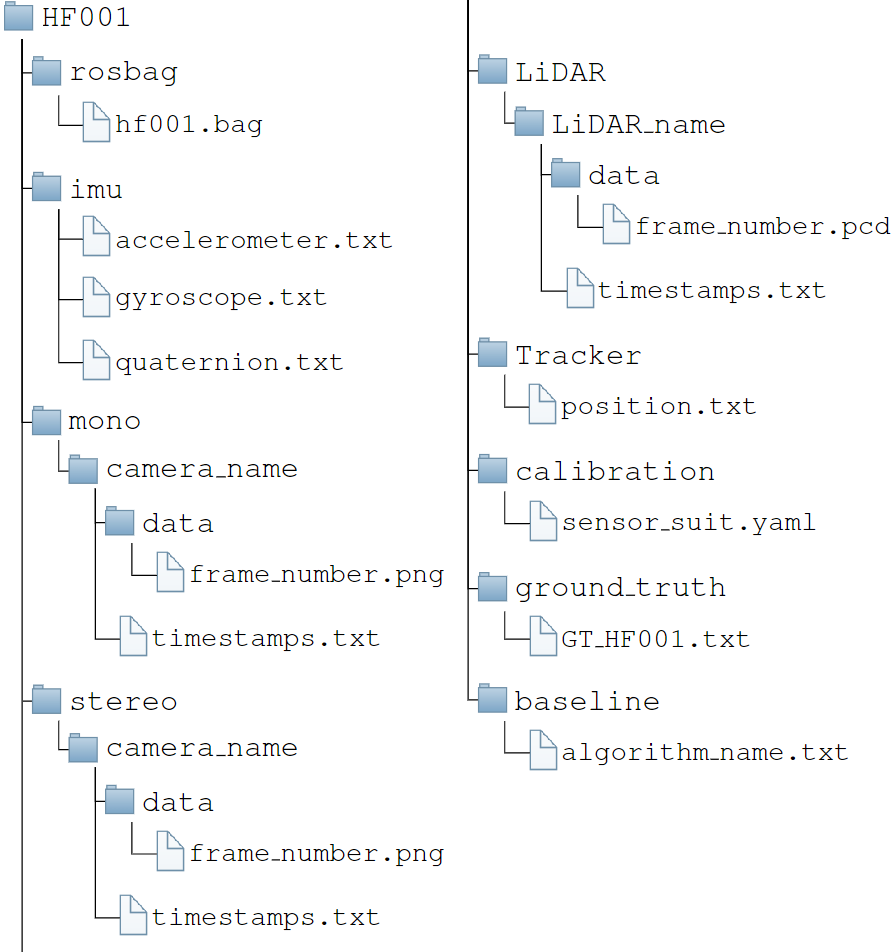}
		\caption{Dataset file structure.}
		\label{filetree}
    \end{figure}
    
        \begin{figure*}[t]
		\centering
		\includegraphics[height=9.699cm,width=17cm]{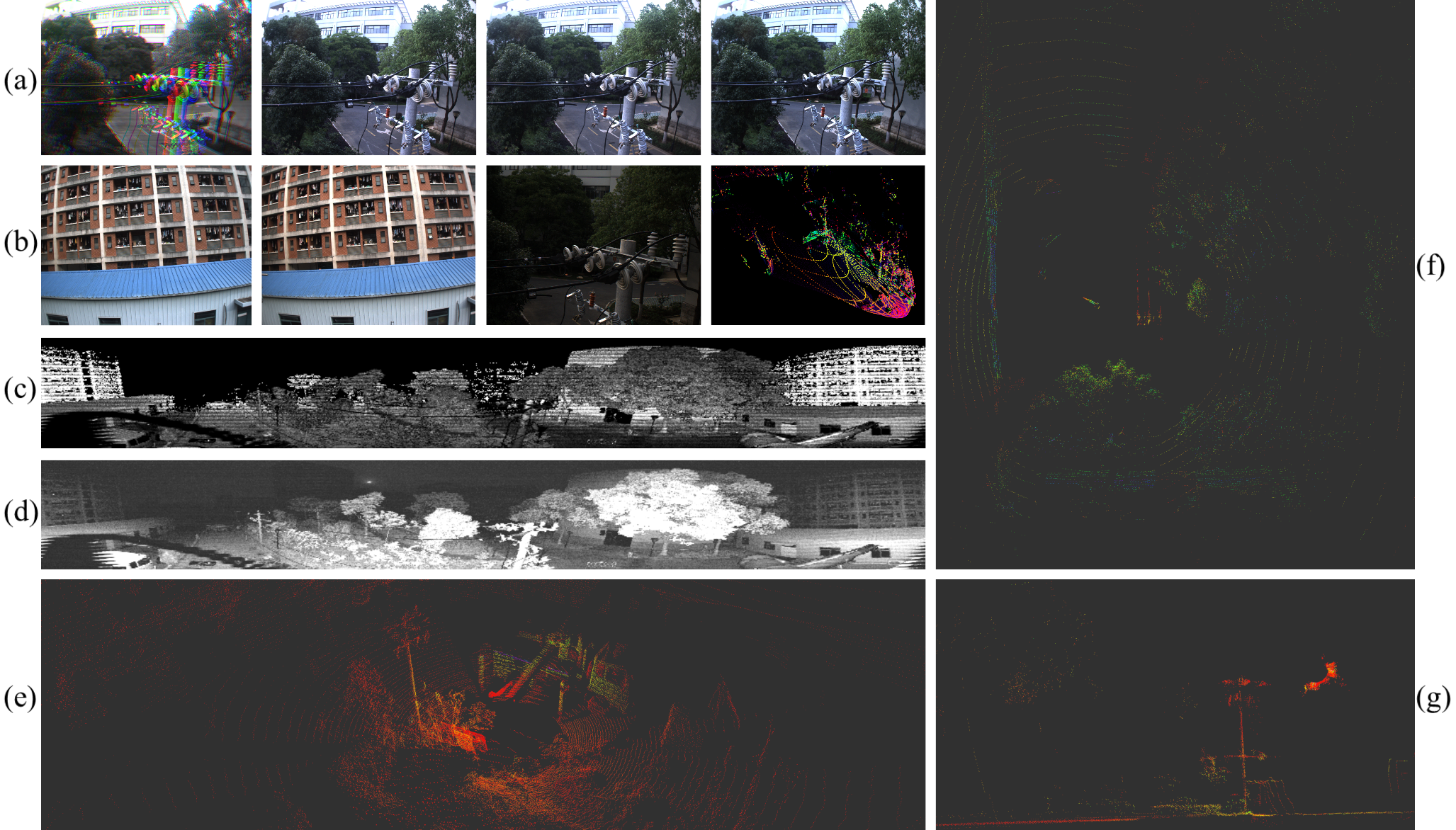}
		\caption{Visualization of multi-sensor data, at the same time. (a) Bumblebee-XB3 raw Bayer image, left, center, right views. (b) Bumblebee-XB2 left, right views; Hikvision camera; LiVOX Avia \revise{point cloud}. (c) reflect image of Ouster OS0-128. (d) signal image of Ouster OS0-128. (e) Ouster OS0-128 \revise{point cloud}. (f) Horizontal Velodyne HDL-32E \revise{point cloud}. (g) Vertical Velodyne VLP-32C \revise{point cloud}.}
		\label{datashow_num}
	\end{figure*}
 
	As shown in Figure~\ref{from_day_to_night}, representative scenes of aerial work were selected to collect data. The surrounding objects include buildings, trees, power lines, roads, etc. At the same time, we collected data from morning to night and under different weather conditions, in order to ensure that the aerial work robot can work around the clock. All data is available here: \url{https://ustc-flicar.github.io/datasets/}. 
	
	The sensor platform \revise{completes} various aerial \revise{motions} through the bucket truck, including movement in XYZ direction and large range rotations. Figure~\ref{doubi_shuoming} shows one of these trajectories and the corresponding bucket truck movement, providing an intuitive explanation of how the motion trajectories in the dataset were generated.
	
	To get an impression of the trajectories, some of the flight paths are shown in Figure~\ref{all_p_new}. These paths were provided by an API T3 laser tracker and recorded on the base station as ground truth position measurements. The laser tracker measurements can be found under the \begin{ttfamily}Tracker\end{ttfamily} folder in the file \begin{ttfamily}position.txt\end{ttfamily}. A short summary of the paths is given in Table~\ref{Table_ATE}. The measurement data from the 9-axis Xsens IMU/INS is stored in the \begin{ttfamily}imu\end{ttfamily} folder. The file \begin{ttfamily}accelerometer.txt\end{ttfamily} contains acceleration data and \begin{ttfamily}gyroscope.txt\end{ttfamily} contains angular velocity data, both in the IMU frame $\boldsymbol{I}$. The file \begin{ttfamily}quaternion.txt\end{ttfamily} contains attitude data in the world frame $\boldsymbol{W}$. To facilitate the use of the dataset, we have provided the optimal estimate of the full pose $\in\mathbb{R}^{3} \times \mathbb{H}$ (both position and attitude) in the body frame $\boldsymbol{B}$ as ground truth based on the original measurement. More details about the generation of ground truth can be found in section~\ref{section_gt}. The ground truth data is provided under the \begin{ttfamily}ground\_truth\end{ttfamily} folder in the file \begin{ttfamily}GT\_HF0XX.txt\end{ttfamily} in TUM format \cite{sturm2012benchmark} \begin{ttfamily}<timestamp,x,y,z,qx,qy,qz,qw>\end{ttfamily}.
	
	The storage format of camera and LiDAR data in our dataset refers to the KITTI dataset. Images are stored with lossless compression using 8-bit PNG files. When collecting data, we only record the original images in Bayer format and do not perform parsing, compression, or filtering (such as Bayer demosaicing) in order to preserve the original information of the data to the greatest extent and improve data utility. For example, as shown in Figure~\ref{datashow_num} (a), the three views from the left, center, and right camera of Bumblebee-XB3 are encoded in the red, green, and blue channels, respectively, and logged to the disk as one Bayer format image. We perform operations such as view separation and Bayer demosaicing offline for the raw image, which strictly guarantees the timestamp and brightness consistency of stereo camera images because they were logged to the disk at exactly the same time in one image. Images of each view are provided separately in the \texttt{left/center/right} folder under the \texttt{xb3} folder.
	
	The LiDAR scans are stored as PCD \revise{files} which save the \textit{(x,y,z)} coordinate of the point cloud. If you want to use additional information of LiDAR scans such as reflection intensity \textit{intensity}, scanning lines \textit{ring} corresponding to the point cloud. The raw binary file we provide in rosbag contains these data.
	
	For the convenience of users who use Robot Operation System (ROS), all sensor data is packaged and provided together in rosbags, and information about the topic name, topic type, frame rate, etc. of the data corresponding to the sensor is organized in Table~\ref{T2}.
	
	The calibration files of the sensors are provided under \begin{ttfamily}calibration\end{ttfamily} folder in \begin{ttfamily}sensor\_suit.yaml\end{ttfamily}. The specific content of each \begin{ttfamily}.yaml\end{ttfamily} file is related to the calibration method of each sensor or sensor suite. We will elaborate necessary information and definition in the next section Section~\ref{section4}. We ran some SOTA SLAM algorithms on the dataset baselines. These results are available in \begin{ttfamily}algorithm\_name.txt\end{ttfamily} under folder \begin{ttfamily}baseline\end{ttfamily}. For more information, please refer to Section ~\ref{section5}.
 
	\section{Sensor Synchronization and Calibration} \label{section4}
	
	  \begin{figure}[t]
		\centering
		\includegraphics[height=7.48536cm,width=8.5cm]{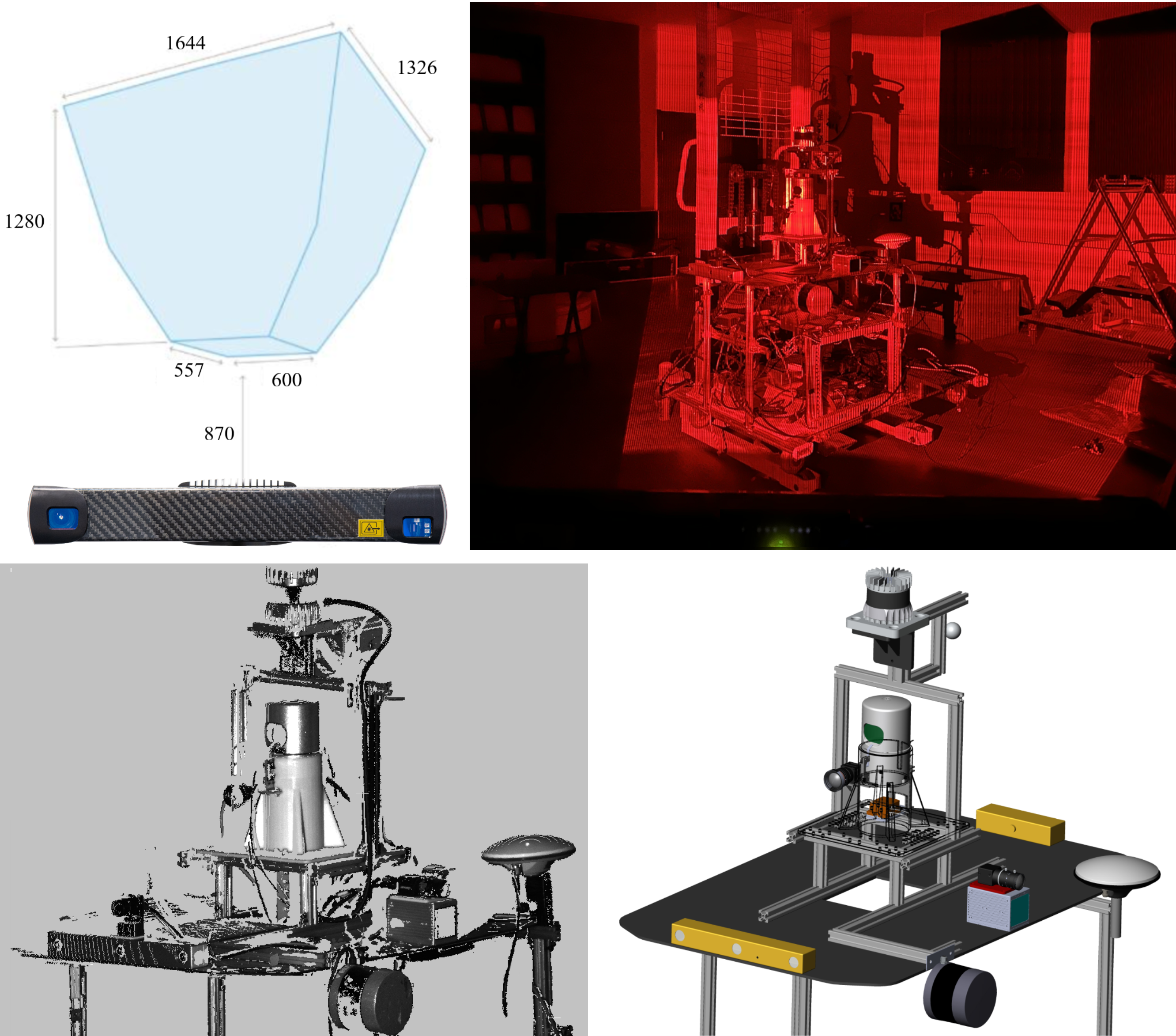}
		\caption{\revise{Precise modeling of the data acquisition multi-sensor platform using Photoneo PhoXi 3D Scanner. The scan result point cloud and the SolidWorks design model}}
		\label{accurate_model}
    \end{figure}

	\begin{figure}[t]
		\centering
		\includegraphics[height=9.71cm,width=8.5cm]{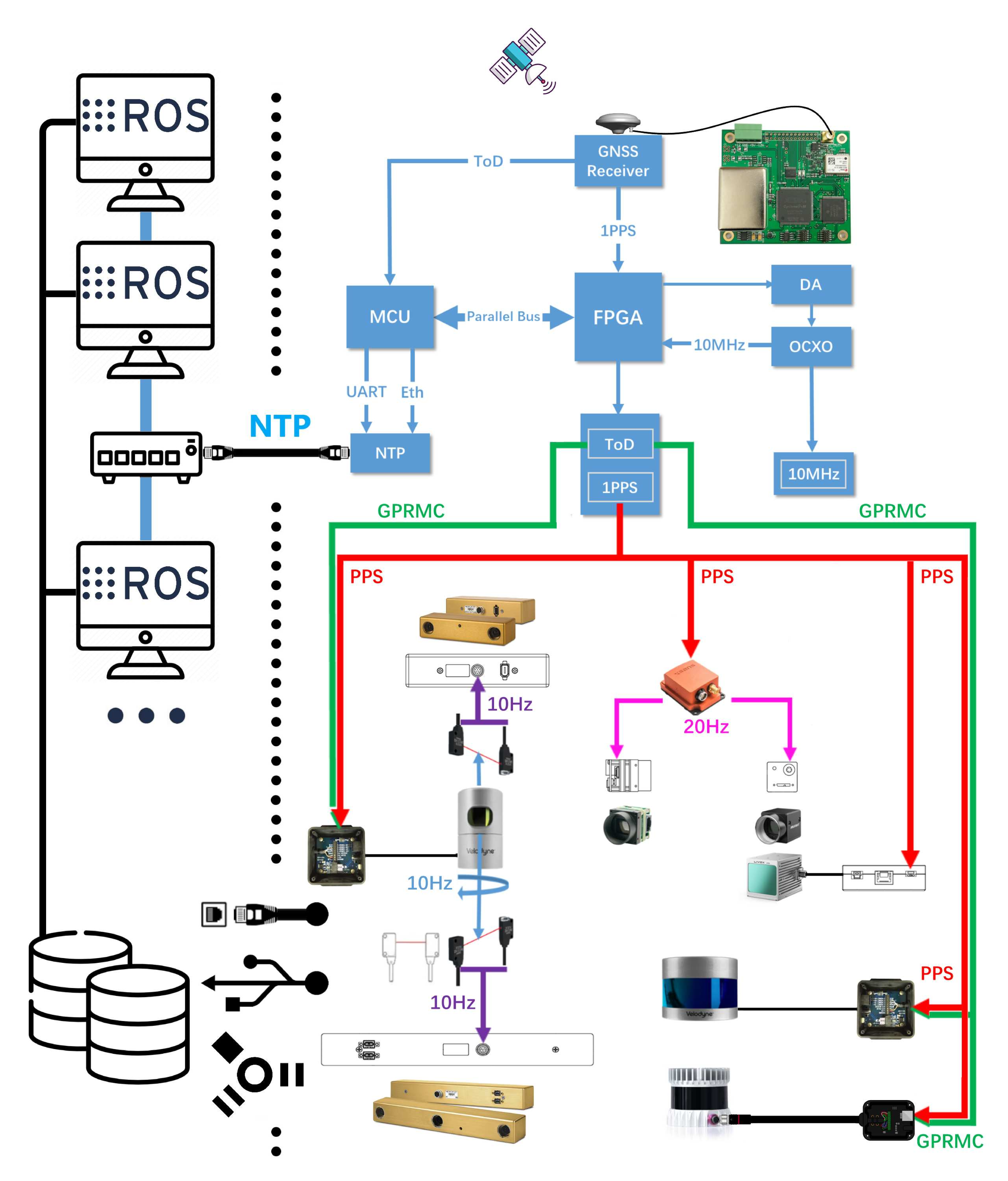}
		\caption{The multi-sensor time synchronization system structure, which is explained in detail in section~\ref{section4-1}.}
		\label{1}
	\end{figure}
	
	Accurate time synchronization and spatial calibration of multiple sensors are necessary for sensor \revise{spatiotemporal} fusion. In the system, sensors are securely mounted using \revise{aluminum} profile brackets, 3D prints, and carbon \revise{fiber} sheets. Time synchronization of multiple sensors and data acquisition computers is achieved using an FPGA-based hard trigger circuit and NTP synchronization network. Calibration data and methods can be obtained from this web page: \url{https://ustc-flicar.github.io/calibration/}.
    \revise{
    \subsection{Accurate Modeling of Multi-sensor Platform} \label{Acc_model}
    }
    \revise{
    As shown in Figure \ref{accurate_model}, we utilize the Photoneo PhoXi 3D Scanner L for precise modeling of the data acquisition multi-sensor platform. Its technical specifications are as follows:
    }
    \revise{
    \begin{itemize}
    \item Scanner Type: structured light scanner
    \item Calibration Accuracy (1 $\sigma$): 0.200 mm
    \item Temporal Noise (1 $\sigma$): 0.190 mm
    \item Resolution: Up to 3.2 million 3D points
    \item Scanning Range: 870 - 2150 mm
    \item Optimal Scanning distance: 1239 mm
    \item Scanning Area (sweet spot): 1082 x 772 mm
    \item Point-to-Point Distance: 0.524 mm
    \item Scanning Time: 250 - 2750 ms
    \end{itemize}
    }
    \revise{
    These specifications exemplify the Photoneo PhoXi 3D Scanner L's capabilities, allowing for precise and detailed modeling for our multi-sensor platform.
    }
    
    \revise{
    Multiple-angle modeling point cloud of data acquisition platform with the PhoXi 3D scanner, SolidWorks assembly models, and drawings are provided along with the dataset. Before users run the calibration program using the USTC FLICAR dataset, we recommend referring to the structured light scanning modeling data and SolidWorks design models to obtain reliable initial calibration values.
    }

	\subsection{Time Synchronization} \label{section4-1}
	The time synchronization module performs the time synchronization of the camera, LiDAR, IMU\revise{,} and the main control computing module. The first level of the time synchronization module is the GNSS receiving module, which obtains the UTC actual time data with nanosecond precision through the satellite. The logic circuit processes the timing information, converts the GNSS signal into PPS and NMEA signals, and connects the LiDAR to the two signals for time synchronization. At the same time, the PPS signal is connected to the IMU module and the frequency divider module. The PPS signal synchronizes the timestamp of the IMU's inertial data with the actual UTC time, the frequency divider module is used to trigger the camera at the desired frame rate using the PPS signal. The camera trigger signal is aligned with the PPS signal at the edge of the whole second, and the delay between the two signals is within a few tens of ns. Therefore, the camera exposure image time is synchronized with the IMU data acquisition time. The time of each camera in a stereo camera is time-synchronized during device manufacture, so the trigger signal triggers all cameras on the serial trigger line simultaneously. The computing master accepts the NTP network data packets converted and sent by the FPGA and performs time synchronization through the NTP protocol.
	
	For the horizontal LiDAR and the stereo camera on the axis, a photoelectric trigger sensor is designed. When the LiDAR rotates to coincide with the camera's field of view, the excitation switch is turned on. The camera is exposed to collect images to ensure the spatiotemporal synchronization of point clouds and images.
	
	\subsection{Mono and Stereo Cameras Calibration}\label{visual-calib}
In order to make full use of the metric information of 2D images for 3D tasks, we calibrate the intrinsic parameters of each camera and the extrinsic parameters between stereo cameras. The calibration approach we use is proposed by \cite{zhang2000flexible}. A known size checkerboard is placed at different distances and attitudes relative to the cameras, and the cameras in the same stereo pair are triggered synchronously. They collect images of the checkerboard at a fixed frame rate as calibration data. The camera parameters are provided in the OpenCV format, which are stored in the \texttt{camera\_name.yaml} calibration file.

The camera parameters are notated as:

• \emph{image size} $\in\mathbb{N}^2$

• \emph{camera\_matrix} $\in\mathbb{R}^{3 \times 3}$

• \emph{distortion\_coefficients} $\in\mathbb{R}^5$

• \emph{rectification\_matrix} $\in\mathbb{R}^{3 \times 3}$

• \emph{projection\_matrix} $\in\mathbb{R}^{3 \times 4}$

Here, the \emph{distortion\_coefficients} vector is used to rectify the tangential and radial distortion of images, using \revise{the} pinhole camera distortion model. The \emph{rectification\_matrix} is only applicable to stereo cameras, which is used to align the epipolar lines between two stereo images for 3D stereo vision geometry calculation. It is \revise{an} identity matrix for monocular cameras.

The camera projection matrix is used to project objects in the 3D world to the camera\revise{'s} 2D image pixels:
\begin{gather}
	P_{proj}
	=
	\begin{bmatrix}
		f_{x} & 0 & c_{x} & T_{x} \\
		0 & f_{y} & c_{y} & T_{y}\\
		0 & 0 & 1 & 0
	\end{bmatrix}
\end{gather}
The left $3 \times 3$ portion is the intrinsic \emph{camera\_matrix} for the rectified image. The fourth column $\begin{bmatrix}T_{x} & T_{y} & 0\end{bmatrix}^{T}$ is to translate the optical center of the second camera to the position in the frame of the first camera. For monocular cameras, $T_{x} = T_{y} = 0$. The average calibration error of monocular cameras is around 0.08 pixels, while the average calibration error of stereo cameras is around 0.1 pixels.

\subsection{Visual Inertial Calibration}\label{visual-inertial}
\begin{figure}[t]
	\centering
	\includegraphics[height=5.41cm,width=8.5cm]{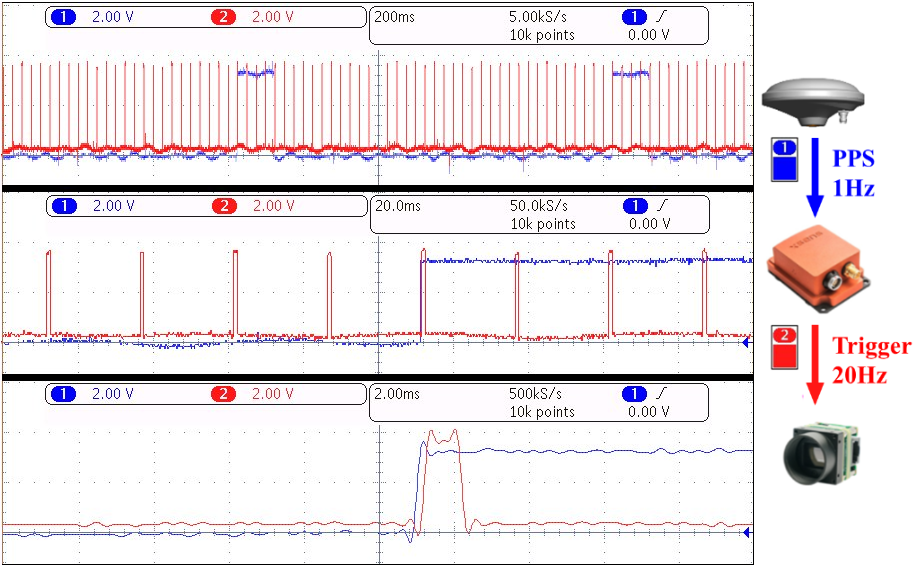}
	\caption{Time synchronization signals in \revise{the} visual-inertial system. The time resolution of the abscissa, from top to bottom, is 200ms, 20ms, and 2ms, from a Tektronix MDO3024  oscilloscope. Blue: 1Hz PPS signal; Red: 20Hz camera trigger signal.}
	\label{imu_cam_time}
\end{figure}
\begin{figure}[t]
	\centering
	\includegraphics[height=4.91cm,width=8.5cm]{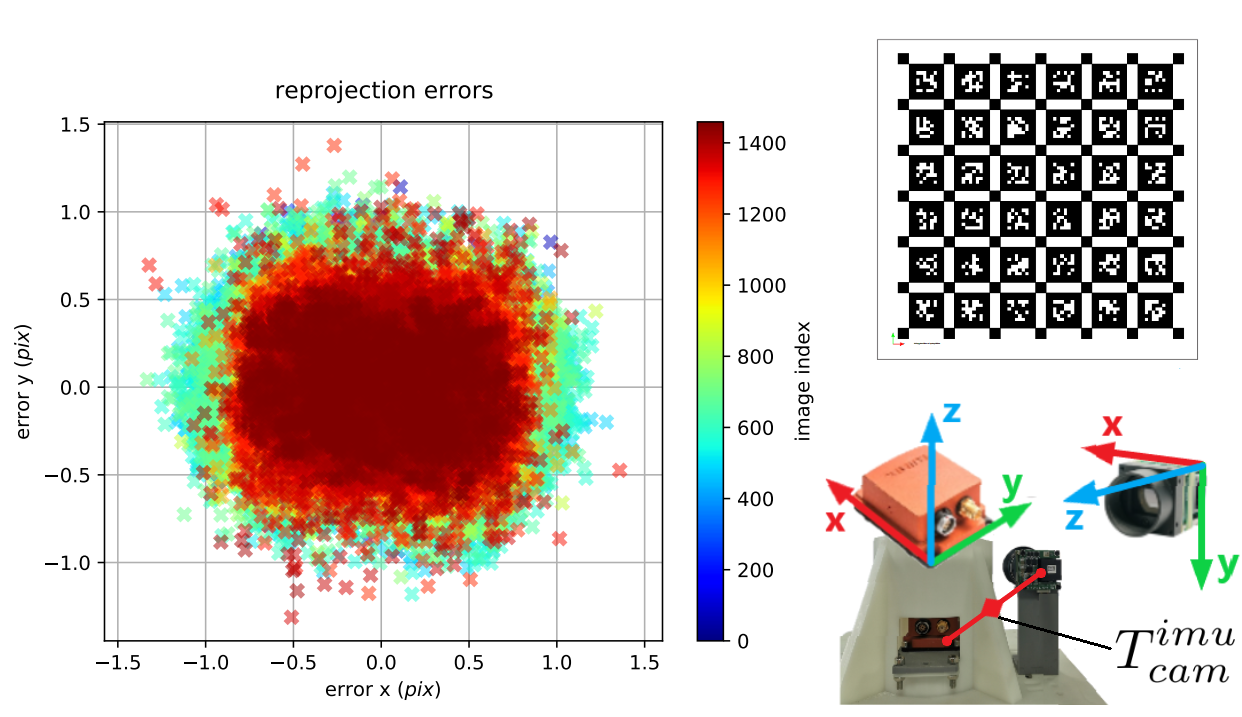}
	\caption{Reprojection error of camera-IMU extrinsics calibration using Kalibr.}
	\label{kalibr_error}
\end{figure}
The fusion of visual and inertial sensors will greatly improve the robustness of the \revise{visual-based} SLAM system. The camera provides \revise{high-resolution} measurements of the environment, while the IMU measures the internal ego-motion of the sensor platform.

The first task is to calibrate the intrinsic parameters of the IMU. The IMU sensor will drift over time, therefore it is necessary to add an error term into the motion model to correct the IMU raw data based on \revise{the} IMU noise model. We fixed the IMU still on the anti-shake optical table for 4 hours and recorded the data. The toolbox \href{https://github.com/gaowenliang/imu_utils}{imu\_utils} is used for calibration.

IMU intrinsic parameters in the corresponding yaml file are as followed:

• $\sigma_g$ --- gyroscope white noise

• $\sigma_a$ --- accelerometer white noise

• $\sigma_{bg}$ --- gyroscope bias instability

• $\sigma_{ba}$ --- accelerometer bias instability

Note that calibration is done in a nearly ideal static setup. In a dynamic setting, the noise will be higher with other factors such as temperature changes. Therefore, it is beneficial to appropriately increase these parameters when using IMU data for camera-IMU extrinsic calibration or visual-inertial odometry.

The second step is to calibrate the extrinsic parameters between the IMU and the camera. The intrinsic parameters of the camera have already been calibrated in section ~\ref{visual-calib}. The time synchronization accuracy between the IMU and the monocular camera is shown in Figure~\ref{imu_cam_time}. The time drift between the IMU clock reference PPS signal and the camera trigger signal is within 0.2 ms.
The \href{https://github.com/ethz-asl/kalibr}{Kalibr} \cite{rehder2016extending} visual-inertial calibration toolbox is used to calibrate the relative spatial relationship between the IMU and the camera. The camera and IMU are rigidly fixed with the base bracket. The overall visual-inertial system performs translation along the XYZ three-axis and full rotation around each axis in front of \revise{an} AprilTag \cite{olson2011apriltag} grid sequences with known size, and records the data for calibration.

camera-IMU extrinsics in the corresponding yaml file are as followed:

• \emph{rotation matrix}: R$_{cam}^{imu}$ $\in SO(3)$ $\subset \mathbb{R}^{3 \times 3}$ 

• \emph{translation vector}: t$_{cam}^{imu}$ $\in\mathbb{R}^{1 \times 3}$

\begin{gather}
	T_{cam}^{imu} 
	=
	\begin{bmatrix}
		R_{cam}^{imu} & t_{cam}^{imu}\\
		0_{1 \times 3} & 1 
	\end{bmatrix}
\end{gather}

The reprojection error of the Camera-IMU extrinsic parameter calibration is shown in Figure ~\ref{kalibr_error},  for most images the reprojection error is within 1.0 \revise{pixels}. The mean, median, and standard deviation of the reprojection error are 0.352 pixels, 0.321 pixels, and 0.197 pixels, respectively.

\begin{figure}[t]
	\centering
	\includegraphics[height=11.5398cm,width=8.5cm]{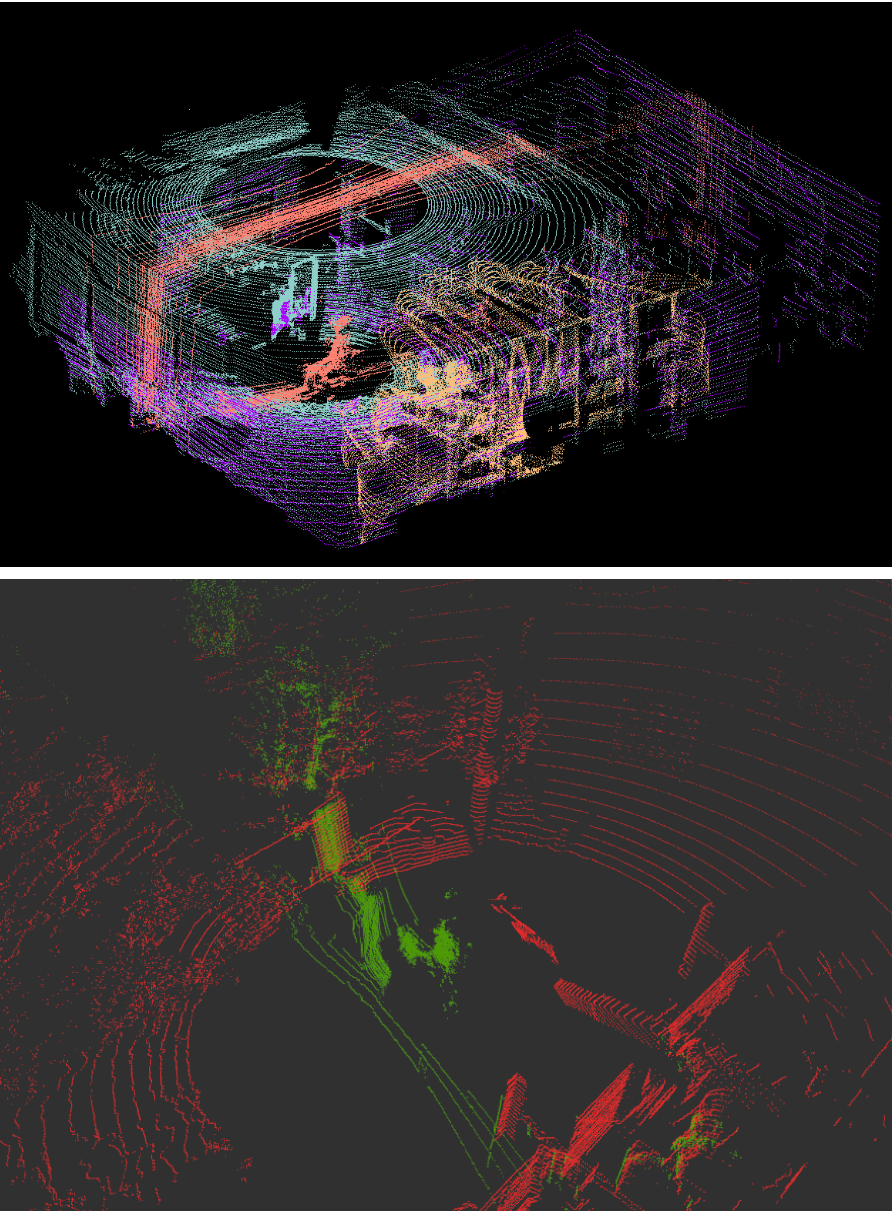}
	\caption{\revise{
 The point cloud was obtained from the calibration process. The image above shows point clouds in the indoor calibration room. The green point cloud is the Ouster OS0-128 LiDAR, the purple point cloud is the Velodyne HDL-32E LiDAR, the red point cloud is the Velodyne VLP-32C LiDAR, and the orange point cloud is the LiVOX Avia LiDAR. \\ The image below shows outdoor measurements. The red point cloud is the Velodyne HDL-32E LiDAR, and the green point cloud is the Velodyne VLP-32C LiDAR.
    }}
	\label{multi_lidar_calib}
\end{figure}

\subsection{Multiple LiDAR Calibration} \label{multi-lidar}

A single LiDAR has problems such as low information density and vertical blind spots. Therefore, we equip the aerial platform with LiDARs from different angles for environmental perception. Extrinsic parameter calibration between multiple LiDARs is a prerequisite for the fusion of LiDAR data.

\revise{Specifically, we refer to the Ouster OS0-128 as the base LiDAR because it has the largest FOV. The extrinsic parameters between the remaining three LiDARs and the OS0-128 are derived through the Normal Distributions Transform (NDT), as proposed by~\cite{ndt}. To minimize the adverse impact of the environment on the calibration accuracy, we choose a non-degenerate scene~\cite{zhang2016degeneracy}  characterized by a complex structure for the calibration task. Meanwhile, to improve the accuracy, a series of frames for each LiDAR were acquired from diverse positions and angles, which optimize the extrinsic parameters jointly.}

\revise{As shown in Figure~\ref{multi_lidar_calib}, according to the calibration results, the point clouds from the four LiDAR sensors (OS0-128, HDL-32, VLP-32\revise{,} and LiVOX) are transformed and displayed in the same coordinate system.}

The extrinsic parameters for multiple LiDARs in the corresponding yaml file are as follows:

• \emph{rotation matrix}: R$_{Lidar1}^{Lidar2}$ $\in SO(3)$ $\subset \mathbb{R}^{3 \times 3}$ 

• \emph{translation vector}: t$_{Lidar1}^{Lidar2}$ $\in\mathbb{R}^{1 \times 3}$
\begin{gather}
	T_{Lidar1}^{Lidar2}
	=
	\begin{bmatrix}
		R_{Lidar1}^{Lidar2} & t_{Lidar1}^{Lidar2}\\
		0_{1 \times 3} & 1 
	\end{bmatrix}
\end{gather}

\subsection{LiDAR Inertial Calibration} \label{lidar-inertial-calib}
We use the online method to calibrate the extrinsic parameters of the rotation between the \revise{Velodyne HDL-32E} LiDAR and the \revise{Xsens MTi-G-710} IMU, which can be seen as \revise{a} kind of hand-eye calibration. The attitude preintegration result of the IMU measurement value from time $t_k$ to time $t_{k+1}$ is denoted as $\boldsymbol{q_{i_k}^{i_{k+1}}}$. $\boldsymbol{q_{L_k}^{L_{k+1}}}$ is the attitude change of the LiDAR scan at time \textit{$t_{k}$} relative to the LiDAR scan at time \textit{$t_{k+1}$}, obtained by \revise{Generalized-ICP \cite{segal2009generalized}}. $\boldsymbol{q_{L}^{i}}$ is the rotation transformation from LiDAR to IMU.

According to the properties of the rotation matrix, we can get:
\begin{align}
	&	\boldsymbol{q_{i_k}^{i_{k+1}}} = \boldsymbol{q_{L}^{i}} \otimes \boldsymbol{q_{L_k}^{L_{k+1}}} \otimes \boldsymbol{q_{i}^{L}}\\
	&   \boldsymbol{q_{i_k}^{i_{k+1}}} \otimes \boldsymbol{q_{L}^{i}} = \boldsymbol{q_{L}^{i}} \otimes \boldsymbol{q_{L_k}^{L_{k+1}}} 
\end{align}

According to the quaternion properties described in \cite{sola2012quaternion}, we transform the above formula $\boldsymbol{q}$ into matrix representation $\boldsymbol{Q}$: 
\begin{align}
	(\boldsymbol{Q_{i_k}^{i_{k+1}}}^{+} - \boldsymbol{Q_{L_k}^{L_{k+1}}}^{-}) \boldsymbol{q_{L}^{i}} = 0 
\end{align}

We fully moved the LiDAR-Inertial system to collect N sets of measurement data. The final task is to solve the following overdetermined system:

\begin{align}
	\begin{bmatrix}
		\boldsymbol{Q_{i_0}^{i_{1}}}^{+} - \boldsymbol{Q_{L_0}^{L_{1}}}^{-}\\
		...\\
		\boldsymbol{Q_{i_{N-1}}^{i_{N}}}^{+} - \boldsymbol{Q_{L_{N-1}}^{L_{N}}}^{-}
	\end{bmatrix} \boldsymbol{q_{L}^{i}}
	= \boldsymbol{A_{4N \times 4}} \boldsymbol{q_{L}^{i}} = 0
\end{align}
We use the SVD method to solve this overdetermined system, perform SVD decomposition on $\boldsymbol{A_{4N \times 4}}$, and then take the eigenvector corresponding to the smallest singular value as the final result of $\boldsymbol{q_{L}^{i}}$. 

The translation vector $t_{Lidar}^{imu}$ between LiDAR and IMU is calibrated and estimated based on the initial value of the size of the 3D model. Since the connecting part is 3D printed, the accuracy of the calibration can be trusted.

LiDAR-IMU extrinsics in the corresponding yaml file are as followed:

• \emph{rotation matrix}: R$_{Lidar}^{imu}$ $\in SO(3)$ $\subset \mathbb{R}^{3 \times 3}$ 

• \emph{translation vector}: t$_{Lidar}^{imu}$ $\in\mathbb{R}^{1 \times 3}$

\begin{gather}
	T_{Lidar}^{imu}
	=
	\begin{bmatrix}
		R_{Lidar}^{imu} & t_{Lidar}^{imu}\\
		0_{1 \times 3} & 1 
	\end{bmatrix}
\end{gather}

\begin{figure}[t]
	\centering
	\includegraphics[height=7.44737cm,width=8.5cm]{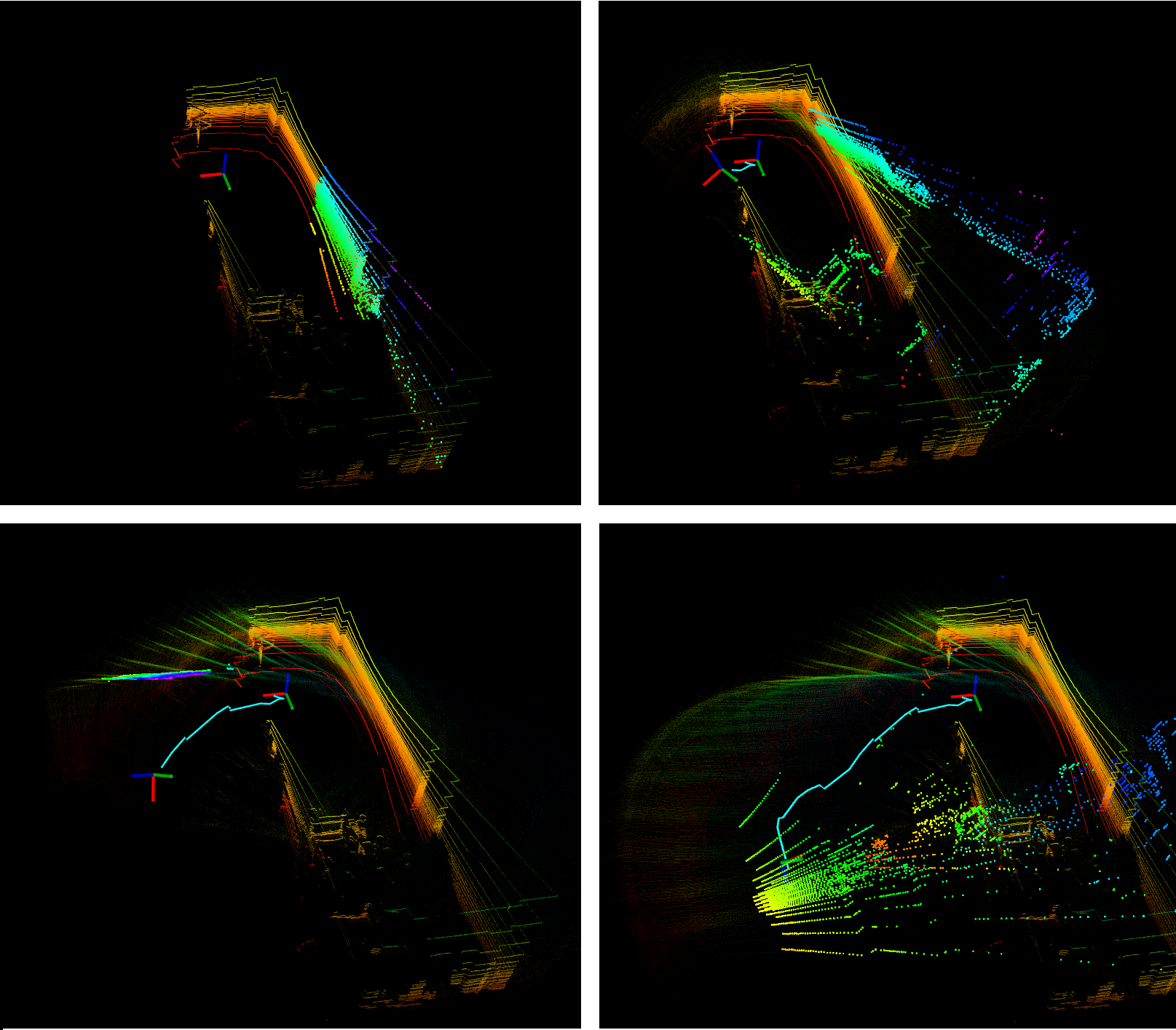}
	\caption{\revise{The failure scenario of direct calibration between Velodyne VLP-32C and Xsens MTi-G-710, when the sensor platform is in a slow movement during initialization, occurring significant drift in the calibration algorithm. LiDAR\_IMU\_Init \citeauthor{zhu2022robust} is used. Similar problems also exist in some other open-source and our own calibration methods.}}
	\label{lidar_imu_fail}
\end{figure}

\revise{
Apart from the Velodyne HDL-32E LiDAR, there are three other LiDAR sensors. Among them, the LiVOX Avia LiDAR and Ouster OS0-128 LiDAR come with built-in IMUs. We recommend utilizing their built-in IMUs for algorithm development involving LiDAR-IMU coupling, such as SLAM. These IMUs are meticulously time-synchronized and spatially calibrated during the manufacturing process, ensuring precise alignment with the LiDAR sensor. For accurate extrinsic parameters, please refer to the manufacturer's manual.
}

\revise{
However, when it comes to the Velodyne VLP-32C LiDAR, we do not recommend directly calibrating it with Xsens. This is primarily due to the greater distance between the two devices and the VLP-32C's vertical installation, which leads to a sparse feature set in the captured point cloud. Consequently, the accuracy of calculating LiDAR odometry through frame-to-frame matching is compromised and the initial value estimation will be more difficult, making it challenging to obtain precise extrinsic parameters in the Xsens coordinate system through direct calibration using existing technology. Although we have explored various good open-source calibration methods for directly calibrating the VLP-32C and Xsens, such as LiDAR\_IMU\_Init \cite{zhu2022robust}, the results have proven unsatisfactory, as depicted in Figure~\ref{lidar_imu_fail}.
}

\revise{
To address this issue, we propose a feasible solution. We recommend leveraging the multi-LiDAR calibration extrinsic parameters obtained in Section~\ref{multi-lidar}, indirectly calculated using the Velodyne HDL-32E LiDAR as an intermediary. Additionally, it is advisable to consult the detailed structured light modeling and Solidworks model described in Section~\ref{Acc_model}. This approach will yield significantly higher accuracy in determining extrinsic parameters compared to direct calibration. Similarly, the LiVOX Avia and Ouster OS0-128 LiDARs can also obtain precise extrinsic parameter transformations to the Xsens Body frame by utilizing the Velodyne HDL-32E LiDAR as an intermediary.
}

\revise{
On the other hand, the USTC FLICAR dataset also provides reliable baselines for the extrinsic parameters calibration of the LiDAR-IMU. We look forward to a LiDAR-IMU extrinsic parameters calibration algorithm with higher robustness, accuracy\revise{,} and versatility.
}

\subsection{LiDAR Camera Calibration} \label{lidar-camera-calib}

\begin{figure}[t]
	\centering
	\includegraphics[height=6.3926cm,width=8.5cm]{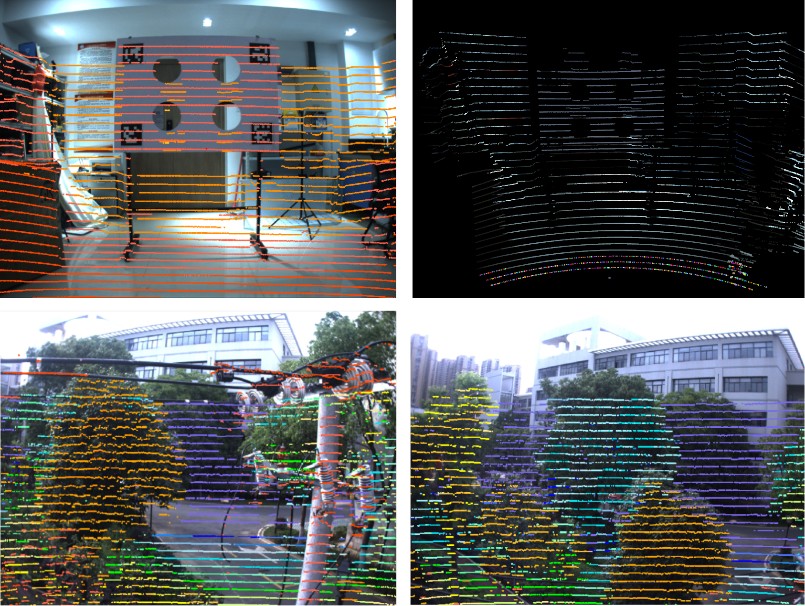}
	\caption{Velo2Cam \citeauthor{beltran2022} camera-LiDAR calibration. \textit{Up-Left}: special calibration board, project LiDAR point cloud to \revise{the} image. \textit{Up-Right}: colorize LiDAR point cloud with image.\\\textit{Down}: Aerial scenes LiDAR points fusion with images. (Velodyne-HDL-32E and Bumblebee-xb3-center)}
	\label{velo2cam}
\end{figure}

Image data has rich and dense object information, but lacks the depth information of the picture. The LiDAR data can just make up for this defect, giving accurate depth information and object structure information. In the process of 3D target detection, the fusion of image and LiDAR \revise{point cloud} information can achieve higher accuracy.

Accurate camera-LiDAR calibration is a necessary condition for the fusion. We use the method proposed by Velo2cam \cite{beltran2022} to get the extrinsic parameters of LiDAR and cameras. Figure~\ref{velo2cam} illustrates the calibration scene and effect of Velo2cam. A special calibration board with four ArUco tags and four circular reference holes is placed in different positions as a calibration target. The 3D pose of each ArUco marker relative to the cameras is obtained by solving a classic perspective-n-point (PnP) problem to obtain the 3D position and orientation of the reference holes in space. Besides, \revise{automatic} targetless calibration approach can also be used to get the extrinsic calibration between LiDARs and cameras. These methods are convenient and can be performed online. Our team conducted a careful survey \cite{li2022automatic} of \revise{automatic} targetless camera-LiDAR calibration.

camera-LiDAR extrinsics in the corresponding yaml file are as followed:

• \emph{rotation matrix}: R$_{velo}^{cam}$ $\in SO(3)$ $\subset \mathbb{R}^{3 \times 3}$ 

• \emph{translation vector}: t$_{velo}^{cam}$ $\in\mathbb{R}^{1 \times 3}$
\begin{gather}
	T_{velo}^{cam}
	=
	\begin{bmatrix}
		R_{velo}^{cam} & t_{velo}^{cam}\\
		0_{1 \times 3} & 1 
	\end{bmatrix}
\end{gather}

The commonly used camera-LiDAR multimodal data fusion schemes directly use LiDAR points as multimodal data aggregation points. LiDAR points are projected to the image plane as follows:
\begin{gather}
        z_{cam}
	\begin{bmatrix}
		x \\
		y \\
		1
	\end{bmatrix}
	=
        h
        P_{proj}
        T_{velo}^{cam}
	\begin{bmatrix}
		X_{velo} \\ 
		Y_{velo} \\
		Z_{velo} \\
		1
	\end{bmatrix}
\end{gather}
where $X_{velo}$, $Y_{velo}$, $Z_{velo}$ denote 3D location of LiDAR point, $x$, $y$, $z_{cam}$ denote its 2D position and projected depth on the image plane. $P_{proj}$ got in section~\ref{visual-calib} denotes the camera intrinsic parameter.And $h$ represents the scaling factor due to down-sampling. Examples of data fusion are shown in Figure~\ref{velo2cam}.

\begin{figure}[t]
    \centering
    \includegraphics[height=6.7cm,width=8.5cm]{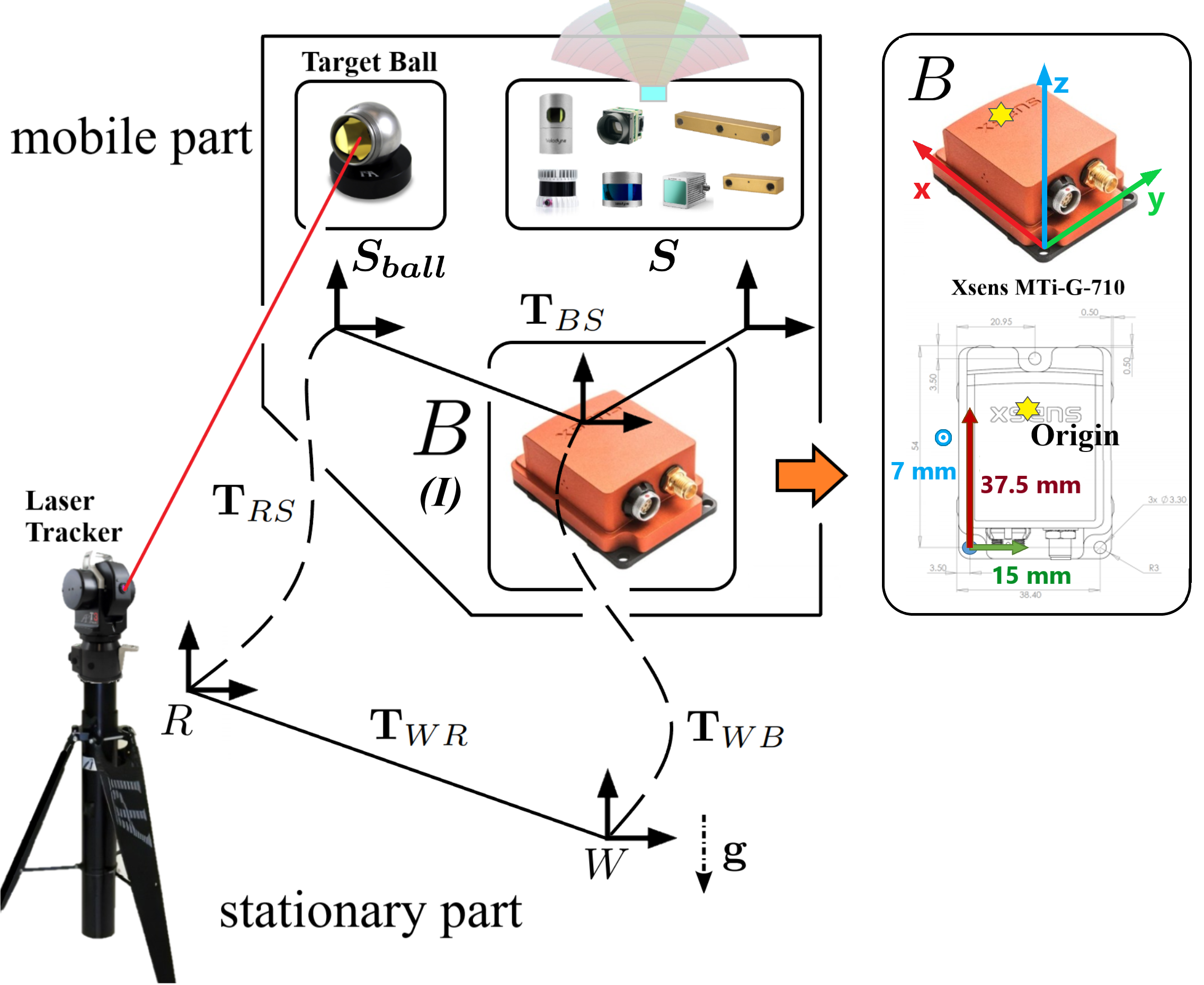}
    \caption{The sensor system used to capture the datasets consists of multiple sensors, each reporting measurements in its own reference frame $\boldsymbol{S}$. The datasets also include raw data from ground truth instruments, reported in the target ball frame $\boldsymbol{S_{ball}}$ and Laser tracker frame $\boldsymbol{R}$.  The body frame $\boldsymbol{B}$ is aligned with the IMU sensor frame $\boldsymbol{I}$.  Calibration information for all extrinsic parameters linking the sensors to the body frame $\boldsymbol{B}$ and intrinsic parameters is included in the dataset. The definition and transformation of the coordinate system of the data acquisition system will be further discussed in Section~\ref{section_gt}}
    \label{system_trans}
\end{figure}

\subsection{Ground Truth Alignment} \label{section_gt}

\begin{figure*}[t]
	\centering
	\includegraphics[height=6.989cm,width=17cm]{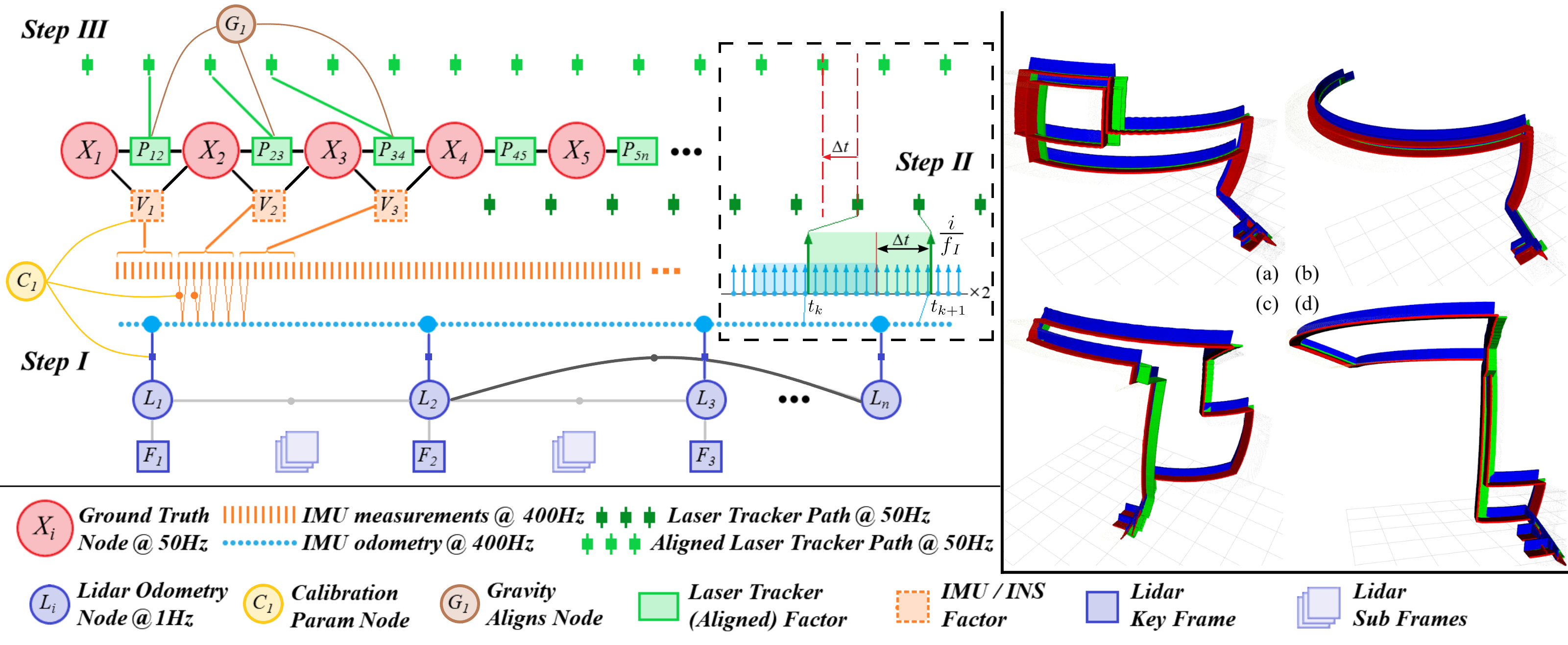}
	\caption{Example of a factor graph created by our system. The states to be estimated are represented by circles, and the measured values are represented by squares. For visualization, we group all calibration parameters into one node $C_1$, $G_1$ is the rotation from gravity alignment to laser tracking frame. On the right are the example ground truth trajectories generated on our dataset. On the right is the visualization of the ground truth of several sequences: (a) HF002, (b) HF003, (c) HF006, (d) HF009.}
	\label{gt_graph}
\end{figure*}

To provide useful and accurate ground truth, the measurements from the laser tracking system are spatiotemporally aligned with the sensor system (body frame $\textbf{\emph{B}}$ is defined at the IMU sensor frame $\textbf{\emph{I}}$). The basic information of the raw measurement data used to generate ground truth is as follows. The definition of the coordinates below can be reviewed in Figure ~\ref{system_trans} :

• The API laser tracking system is mounted horizontally on the ground using a tripod. The target ball frame $\boldsymbol{S_{ball}}$ it tracks is rigidly fixed to the body of the sensor platform. The 3D position of the target point trajectory in the laser tracking frame $\boldsymbol{R}$, represented as $\boldsymbol{p_{S_{ball}}^{R}}$ $\in\mathbb{R}^{3}$, is output at 50Hz (second highest). The time \textit{$t_{R}$} is referenced to the intrinsic clock of the tracking system.

• The Xsens IMU/INS is rigidly fixed to the center of the sensor platform body with the body frame $\textbf{\emph{B}}$ defined at its own frame $\textbf{\emph{I}}$. Acceleration $\boldsymbol{a_{I(B)}}$ $\in\mathbb{R}^{3}$ and angular velocity $\boldsymbol{\omega_{I(B)}}$ $\in\mathbb{R}^{3}$ are output under the IMU frame $\textbf{\emph{I}}$ at 400Hz (highest). The origin of the body frame $\textbf{\emph{B}}$ coincides with the origin of the accelerometer. At the same time, the AHRS (Attitude and Heading Reference System) system of Xsens outputs the attitude $\boldsymbol{q_B^W}$ $\in\mathbb{H}$ of the sensor platform in world frame $\boldsymbol{W}$ (NED coordinates). $\boldsymbol{q_W}$ is a statistical optimal 3D orientation estimate computed by \revise{the} Xsens Kalman Filter algorithm (XKF3) using signals of the rate gyroscopes, accelerometers\revise{,} and magnetometers. Xsens is synchronized with the GPS clock, the time \textit{$t_{I(B)}$} = \textit{$t_{gps}$}.

• The Velodyne HDL-32E forms \revise{a} LiDAR-Inertial system together with Xsens IMU. Here its data is not used directly but used as LiDAR odometry to constrain and optimize the trajectory obtained by IMU preintegration. It is synchronized with the same GPS clock as Xsens, therefore \textit{$t_{lidar}$} = \textit{$t_{gps}$} = \textit{$t_{I(B)}$}.

In order to make full use of these data to generate reliable ground truth, three steps are performed as shown in Figure ~\ref{gt_graph}. Here comes the details of each step :

• \textbf{Step I. IMU Preintegration}

The purpose of Step I is to preintegrate the high-frequency data of the IMU to obtain trajectory data at the same frequency as the IMU (400Hz), which will serve as a benchmark for Step II. Based on \revise{the} IMU noise model, the measurements of angular velocity and acceleration
from IMU are defined as:
\begin{align}
	&	\boldsymbol{^{t}{\hat{\omega}}_{B}} = \boldsymbol{^{t}{\omega_{B}}} \boldsymbol{^{t}{b_{\omega}}} + \boldsymbol{^{t}{n_{\omega}}}
	\\
	&	\boldsymbol{^{t}{\hat{a}}_{B}} = \boldsymbol{R_{W}^{B}} \left(\boldsymbol{^{t}{a_{B}}} - \boldsymbol{g_{W}}\right) + \boldsymbol{^{t}{b_{a}}} + \boldsymbol{{n_{a}}}
\end{align}
where $\boldsymbol{g_{W}}$ is the constant gravity vector in the world frame $\boldsymbol{W}$, and $\boldsymbol{R_{W}^{B}}$ is the rotation matrix from $\boldsymbol{W}$ to $\boldsymbol{B}$. Adding a slowly varying bias $\boldsymbol{^{t}{b}}$ and white noise $\boldsymbol{{n}}$ to the model corrects for $\boldsymbol{^{t}{{\omega}}_{B}}$ and $\boldsymbol{^{t}{{a}}_{B}}$. During time \textit{$t + \delta t$}, velocity $\boldsymbol{^{t + \delta t}{v_{B}}}$ $\in\mathbb{R}^{3}$, position $\boldsymbol{^{t + \delta t}{p_{B}}}$ $\in\mathbb{R}^{3}$ and rotation $\boldsymbol{^{t + \delta t}{q_{B}}}$ $\in\mathbb{H}$ are obtained by preintegrating (discrete calculus) the corrected acceleration and angular velocity measurements according to Newton's laws of motion.

The 1Hz LiDAR odometry is modified based on LOAM \cite{zhang2014loam} \revise{using Velodyne HDL-32E LiDAR data}, which has good accuracy and robustness. Due to the long-term movement of the IMU will produce drift errors, the LiDAR odometry provides an initialization reference for the IMU preintegration trajectory \revise{per second}, making the results more accurate. The transformation $T_{lidar}^{imu}$ $\in\textit{SE(3)}$ from $\boldsymbol{S_{lidar}}$ to $\boldsymbol{B}$ has been obtained by the method mentioned in Section ~\ref{lidar-inertial-calib}. 

The final result of the first step is a 400Hz IMU odometry, its final form is: 
\begin{align}
&	\boldsymbol{x_{B_0}} = \left[ \boldsymbol{t_{B}}\ \boldsymbol{p_{B_0}}\ \boldsymbol{q_{B_0}} \right]^{T}
\end{align}

\begin{figure}[t]
	\centering
	\includegraphics[height=6.5467cm,width=8.5cm]{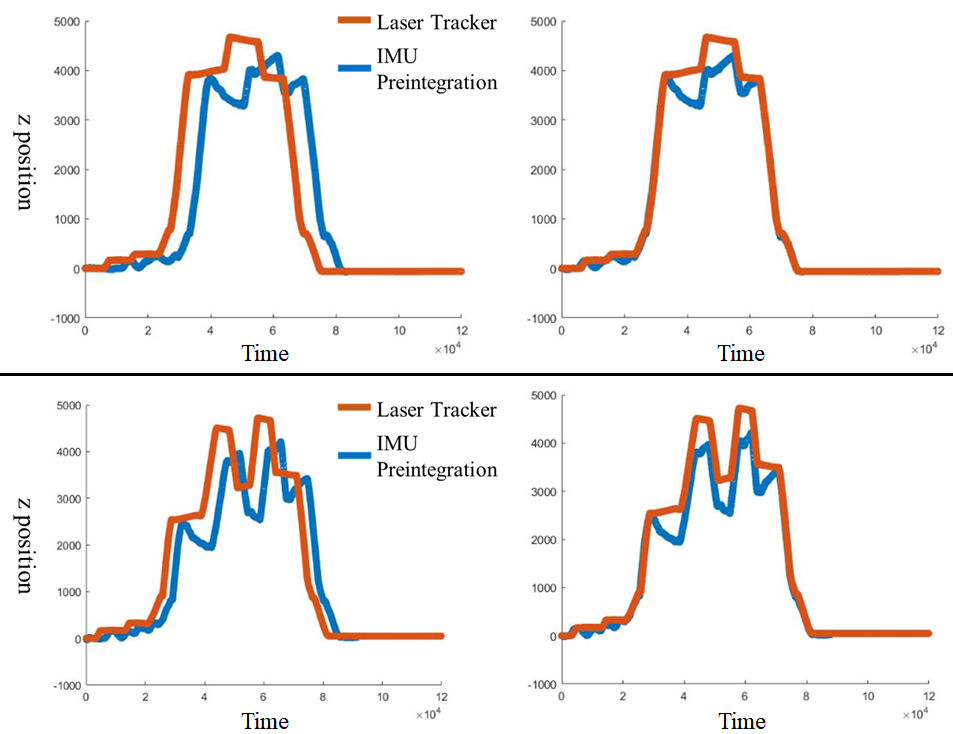}
	\caption{\revise{
    Temporal alignment of IMU preintegration trajectories with laser tracker trajectories via motion correlation analysis in Step II.
    }}
	\label{signal_align}
\end{figure}

• \textbf{Step II. Motion Correlation Analysis}
The purpose of step II is to \revise{spatiotemporally} align the laser tracker measurements with the body frame based on motion correlation analysis, \revise{as shown in figure~\ref{signal_align}.} Our method is similar to \cite{qiu2020real}. In this step\revise{,} we have to \revise{analyze} two signals, the first is IMU odometry $\boldsymbol{x_{B_0}}$ gotten in step I, \revise{and} the second is the measurement of laser tracking system $\boldsymbol{x_{R}} = \left[ \boldsymbol{t_{R}}\ \boldsymbol{p_{S_{ball}}^{R}}\ \boldsymbol{{p_0}_{S_{ball}}^{B}} \right]^{T}$, $\boldsymbol{{p_0}_{S_{ball}}^{B}}$ indicates the initial transformation from $\boldsymbol{S_{ball}}$ to $\boldsymbol{B}$ estimated from the size of the mechanical installation. For the two signals, since the laser tracker and the bucket truck are kept horizontal to the ground using \revise{a} tripod and hydraulic legs respectively during the measurement process, the motion in the z direction already has a high correlation without rotation estimate. Therefore, the time delay analysis is performed by calculating the cross-correlation between $z_{B_0} (t_B)$ and $z_{R} (t_R)$. We enumerate the time offset, \revise{and} the maximum of the cross-correlation function indicates the point in time where the signals are best aligned:
\begin{align}
{\displaystyle \tau _{\mathrm {delay} }={\underset {t_d\in \mathbb {R} }{\operatorname {arg\,max} }}((z_{B_0}\star z_R)(t_d))}
\end{align}
After \revise{finishing} this step state $\boldsymbol{x_{R}}$ is aligned to $\boldsymbol{t_{B}}$. Here we get: 
\begin{align}
&	\boldsymbol{x_{B_1}} = \left[ \boldsymbol{t_{B}}\ \boldsymbol{p_{S_{ball}}^{R}}\ \boldsymbol{{p_0}_{S_{ball}}^{B}} \right]^{T}
\end{align}
Finally\revise{,} the raw data and states \revise{prepared} for step III is as followed ($\boldsymbol{S_{b}} = \boldsymbol{S_{ball}}$) :
\begin{align}
\boldsymbol{x_{raw}} = \left[ \boldsymbol{t_{B}}\ \boldsymbol{p_{S_{b}}^{R}}\ \boldsymbol{{p_0}_{S_{b}}^{B}}\ \boldsymbol{q_{B}^{W}}\ \boldsymbol{g_{W}}\ \boldsymbol{a_{B}}\ \boldsymbol{\omega_{B}}\ \boldsymbol{{b_{a,\omega}}}\ \boldsymbol{{n_{a,\omega}}} \right]^{T}
\end{align}

• \textbf{Step III. Factor Graph Optimization}

The final step is to use the known measurements and states to generate an optimal estimate of all states of the ground truth in the body frame $\boldsymbol{B}$. Our method is modified based on the vicon2gt \cite{Geneva2020TRVICON2GT} toolbox. Specifically\revise{,} the following states are estimated:
\begin{align}
&  \boldsymbol{x} = \left[ \boldsymbol{x_{B_1}}\ ... \boldsymbol{x_{B_N}}\  \boldsymbol{x_C}\ \bar{q}\boldsymbol{_{W}^{R}}\ \right]^{T}
\\
&  \boldsymbol{x_{B_i}} = \left[ \bar{q}\boldsymbol{_{R}^{B_i}}\ \boldsymbol{p_{B_i}^R}\ \boldsymbol{v_{B_i}^R}\ 
\boldsymbol{b_{a,i}}\ \boldsymbol{b_{\omega,i}}\ \right]^{T}
\\
&  \boldsymbol{x_{C}} = \left[ \bar{q}\boldsymbol{_{S_b}^{B}}\ \boldsymbol{p_{S_b}^{B}}\  \Delta{t_B^R} \right]^{T}
\end{align}

Here we are estimating N inertial states at laser tracker frequency 50Hz, along with a calibration state $\boldsymbol{x_C}$ containing the \revise{spatial-temporal} parameters between the \revise{target} ball frame $\boldsymbol{S_{b}}$ and IMU frame $\boldsymbol{B}$, and $\bar{q}\boldsymbol{_{W}^{R}}$ the rotation between the global laser tracking frame $\boldsymbol{R}$ and global inertial frame $\boldsymbol{W}$. $\bar{q}\boldsymbol{_{R}^{B_i}}$ is the unit quaternion parameterizing the rotation from the global laser tracking frame of reference $\boldsymbol{R}$ to the IMU local frame $\boldsymbol{B_i}$ at time $t_i$. $\boldsymbol{{p_{B_i}^{R}}}$ and $\boldsymbol{{v_{B_i}^{R}}}$ are the position and velocity of the IMU body frame $\boldsymbol{B}$ expressed in the global laser tracking frame $\boldsymbol{R}$, respectively. $\boldsymbol{b_{a,i}}$ and $\boldsymbol{b_{\omega,i}}$ are the biases of accelerometer and gyroscope. $\Delta{t_B^R}$ is the time offset between the laser tracking system and the IMU body frame that we further estimate on the basis of step II. The inertial state $\boldsymbol{x_{B_i}}$ lies on the manifold defined by the product of the unit quaternions $\mathbb{H}$ with the vector space $\mathbb{R}^{12}$ (i.e. $\mathcal{M}$ = $\mathbb{H}$ $\times$ $\mathbb{R}^{12}$) and has 15 DoF.

An overview of the nonlinear factor graph we solved is shown in Figure~\ref{gt_graph}.

The final ground truth is available in TUM format \cite{sturm2012benchmark}:\\
\begin{ttfamily}\footnotesize
timestamp tx ty tz qx qy qz qw\\
1654673708.251979 -0.000063 0.000072 -0.000068 -0.050800 -0.025895 0.001970 0.998371 \\
...
\end{ttfamily}

\begin{table*}[t]
\footnotesize\sf\centering
\caption{ATE \revise{(m)} of state-of-the-art SLAM methods over USTC FLICAR datasets. * points out that runs have not been successful on less than 1/2 of a sequence. $\times$ points out that runs have not been successful on more than 1/2 of a sequence. — points out that data is not available in a sequence. \revise{The ATE in this table is calculated using evo \cite{grupp2017evo}.}
}
\label{Table_ATE}
\begin{tabular}{lllllllllll}
	\toprule
	Sequence & Sensor Suite & hf001 & hf002 & hf003 & hf004 & hf005 & hf006 & hf007 & hf008 & hf009\\
	\midrule
	Time & & 15:35 & 15:48 & 16:14 & 16:30 & 17:20 & 18:13 & 19:01 & 21:20 & 21:27\\
	\midrule
	\makecell[c]{ORB-SLAM3} & Hikcam & 0.097 & 0.028 & 0.093 & 0.081 & 0.163 & 0.144* & 0.084 & \makecell[c]{$\times$} & \makecell[c]{$\times$} \\
	\makecell[c]{(Visual)}& Xb3-C & 0.144 & 0.182 & 0.106 & 0.113 & 0.417* & 0.178 & 0.126 & \makecell[c]{$\times$} & \makecell[c]{$\times$} \\
	\makecell[c]{ORB-SLAM3} & Hikcam+Xsens & 0.017 & 0.181 & 0.090 & 0.082 & 0.086* & 0.118 & 0.078 & \makecell[c]{$\times$} & \makecell[c]{$\times$} \\
	\makecell[c]{(V-Mono-Inertial)} & Xb3-C+Xsens & 0.150 & 0.276 & 0.156 & 0.116 & 0.159* & 0.184* & 0.407 & \makecell[c]{$\times$} & \makecell[c]{$\times$} \\
	\makecell[c]{VINS-Mono} & Hikcam+Xsens & 0.146 & 0.291 & 0.332 & 0.092 & 0.204 & 0.232 & 0.131 & 1.381 & 1.794 \\
	\makecell[c]{(V-Mono-Inertial)} & Xb3-C+Xsens & 0.108 & 0.166 & 0.163 & 0.089 & 0.156 & 0.150 & 0.125 & 0.074 & 0.116 \\
	\makecell[c]{ORB-SLAM3\\(Stereo)} & Xb3-L/R & 0.150 & 0.184 & 0.135 & 0.117 & 0.215 & 0.200 & 0.137 & \makecell[c]{$\times$} & \makecell[c]{$\times$} \\
	\makecell[c]{VINS-Fusion\\(Stereo)} & Xb3-L/R & 0.155 & 0.211 & 0.147 & 0.097 & 0.213 & 0.250 & 0.131 & 0.154 & 1.277 \\
	\makecell[c]{ORB-SLAM3\\(V-Stereo-Inertial)} & \makecell[l]{Xb3-L/R + Xsens} & 0.406 & 0.184 & 0.205 & 0.136 & 0.226 & 0.282 & 0.218 & 0.568* & 0.551* \\
	\makecell[c]{VINS-Fusion\\(V-Stereo-Inertial)} & \makecell[l]{Xb3-L/R + Xsens} & 0.116 & 0.192 & 0.096 & 0.089 & 0.164 & 0.161 & 0.166 & 0.059 & 0.136\\
	\midrule
	\makecell[c]{A-LOAM\\(Horizontal-LiDAR)} & Velo-HDL32 & 0.089 & 0.118 & 0.091 & 0.078 & 0.171 & 0.115 & 0.078 & 0.053 & 0.062\\
	\makecell[c]{A-LOAM\\(Vertical-LiDAR)} & Velo-VLP32 & 0.265 & 0.278 & 0.526 & \makecell[c]{—} & 0.423 & 0.481 & 0.489 & 0.507 & 0.644\\
	\makecell[c]{LeGO-LOAM\\(Horizontal-LiDAR)} & Velo-HDL32 & 0.094 & 0.121 & 0.075 & 0.081 & 0.165 & 0.116 & 0.080 & 0.046 & 0.059\\
	\makecell[c]{LeGO-LOAM\\(Vertical-LiDAR)} & Velo-VLP32 & 0.698 & 1.528 & 1.047 & \makecell[c]{—} & 0.771 & 0.484 & 1.214 & 1.962 & 1.325\\
	\makecell[c]{LIO-SAM\\(H-LiDAR-Inertial)} & \makecell[l]{Velo-HDL32\\ + Xsens} & 0.086 & 0.114 & 0.075 & 0.079 & 0.161 & 0.111 & 0.073 & 0.042 & 0.054\\
	\makecell[c]{FAST-LIO\\(H-LiDAR-Inertial)} & \makecell[l]{Velo-HDL32\\ + Xsens} & 0.088 & 0.115 & 0.081 & 0.078 & 0.168 & 0.117 & 0.075 & 0.052 & 0.059\\
	\makecell[c]{FAST-LIO\\(H-MEMS-Inertial)} & \makecell[l]{LiVOX-Avia\\ + Internal IMU} & 0.063 & 0.069 & 0.111 & 0.079 & 0.147 & 0.119 & 0.082 & 0.050 & 0.060\\
	\midrule
	Duration (s) & & 192.5 & 217.8 & 217.1 & 155.9 & 260.4 & 230.6 & 207.6 & 210.6 & 238.7 \\
	Length (m) & & 26.46 & 33.50 & 34.26 & 24.10 & 22.82 & 33.90 & 34.32 & 30.78 & 35.42 \\
	Avg. Vel./ (m/s) & & 0.137 & 0.154 & 0.158 & 0.155 & 0.088 & 0.147 & 0.165 & 0.146 & 0.148 \\
	Ang. Vel. ($^{\circ}$/s) & & 0.537 & 0.529 & 0.503 & 0.660 & 0.556 & 0.614 & 0.612 & 0.574 & 0.148 \\
	\bottomrule
\end{tabular}\\[10pt]

\end{table*}

\section{Evaluation and Baselines} \label{section5}

We run some state-of-the-art baselines on several sensor suites and data sequences to illustrate the characteristics and challenges of our dataset. The absolute trajectory error (ATE, as defined in \cite{sturm2012benchmark}) is used as the indicator to measure the effect of the SLAM algorithms. To ensure fairness, we carefully tuned the parameters for the algorithms evaluated on each data sequence to make the results of each algorithm close to their best. And parameters of the algorithms using the same sensor suite are set to be exactly the same, the online parameter estimation of some algorithms will not be enabled. Table~\ref{Table_ATE} summarizes necessary information and the corresponding results. The algorithms evaluated are run on a PC with Ubuntu 18.04 operating system, ROS melodic, Intel® Core™ i7-8750H CPU @ 2.20GHz, and 16GB RAM.

For the evaluation of visual SLAM, we have tested several state-of-the-art algorithms on different sensor suites and data sequences. These include ORB-SLAM3 \cite{campos2021orb} on two monocular cameras, ORB-SLAM3 and VINS-Mono \cite{qin2018vins} on two monocular-inertial systems, and ORB-SLAM3 and VINS-Fusion on a stereo and stereo-inertial system. Our stereo camera consists of three cameras, resulting in three stereo pairs in total. In this case, we have chosen the left and right cameras with the longest baseline of 24cm.

\begin{figure}[t]
\centering
\includegraphics[height=4.281cm,width=8.5cm]{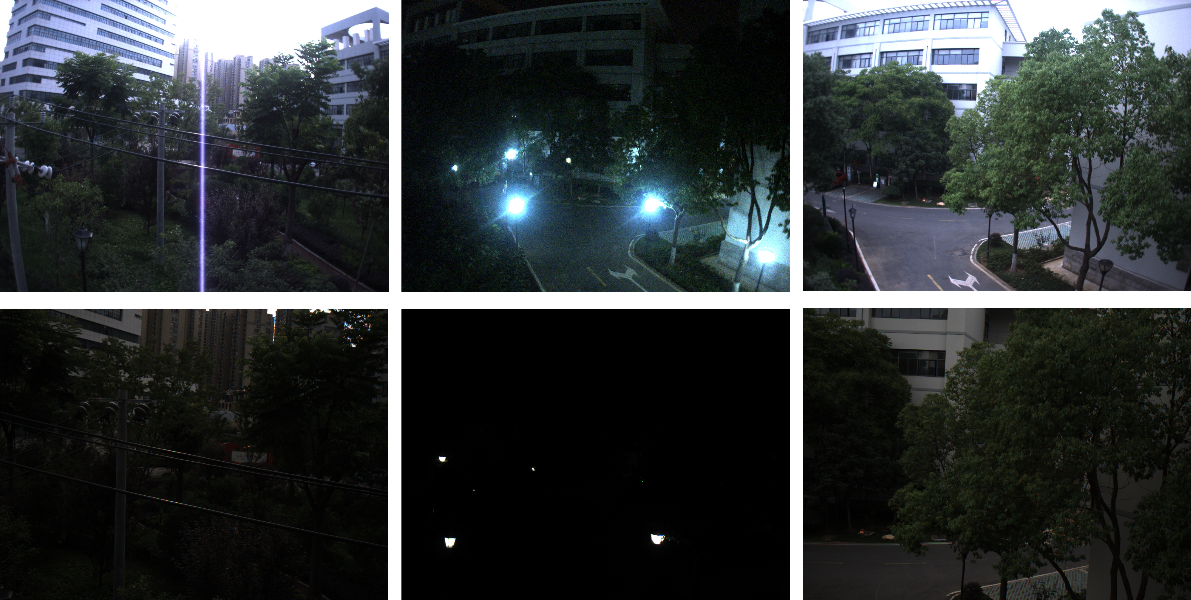}
\caption{The different performances of the two cameras under complex lighting conditions. The upper part is the Bumblebee XB3 center camera, and the lower part is the hikvison camera. The upper and lower pairs are at the same moment.}
\label{camera_light}
\end{figure}

As shown in the table, we have tested the same algorithm on two different monocular cameras because they are complementary to each other. As shown in Figure~\ref{camera_light}, the Bumblebee-XB3 camera has a larger field of view and higher sensitivity to light compared to the Hikvision camera and can capture more environmental information in the darkness, but it may produce glare in strong light. The Hikvision camera has the opposite strengths and weaknesses. We hope that at least one camera can provide a good environment perception for visual SLAM in each working environment. At the same time, we are analyzing why the accuracy of visual SLAM decreases in some sequences by comparing the results on the two cameras. For example, in the hf005 sequence, the accuracy of ORB-SLAM3 on the Bumblebee-XB3 camera is significantly lower than on the Hikvision camera and fails halfway, which is probably due to glare from the xb3 camera. Similarly, in the hf009 sequence, the accuracy of VINS-Mono on the Bumblebee-XB3 camera is significantly higher than on the Hikvision camera, which is because the Bumblebee-XB3 camera provides more environmental information in the darkness.

Overall, the accuracy of the ORB-SLAM3 method is slightly higher than the VINS-based SLAM method in sequences with good lighting conditions (no glare, no darkness), consistent with the experimental results reported in their paper, which is based on the EuRoC dataset. However, the robustness of the VINS-based SLAM method is significantly higher in extreme light and dark environments, and both can maintain the same accuracy as in good lighting conditions. For example, in the hf008 and hf009 night sequences, we used the Bumblebee-XB3 camera which can image effectively in the darkness. However, the ORB-SLAM3 algorithm still failed due to its inability to extract features. On the other hand, both VINS-Mono and VINS-Fusion were able to maintain the same accuracy as \revise{the situation} in good lighting conditions.

Besides, the accuracy and robustness of visual SLAM under semi-failures can be improved by tightly coupling a well-calibrated IMU. For example, in the hf009 night sequence with \revise{a} wide range of rotation, the ATE of VINS-Fusion that only uses a stereo camera is 1.277 m. If IMU data is used, the ATE of VINS-Fusion will decrease to 0.136 m.

For the evaluation of LiDAR SLAM, we tested A-LOAM (an implementation of LOAM \cite{zhang2014loam} modified by \citeauthor{qin2018vins}) and LeGO-LOAM \cite{shan2018lego} on two 32-beam Velodyne LiDARs, one horizontal and one vertical. We also used a LiDAR-Inertial system consisting of a horizontal Velodyne HDL-32E LiDAR and an Xsens IMU to test LIO-SAM \cite{shan2020lio} and FAST-LIO \cite{xu2021fast}. FAST-LIO was also tested on a LiVOX MEMS LiDAR.

\begin{figure}[t]
	\centering
	\includegraphics[height=11.649449cm,width=8.5cm]{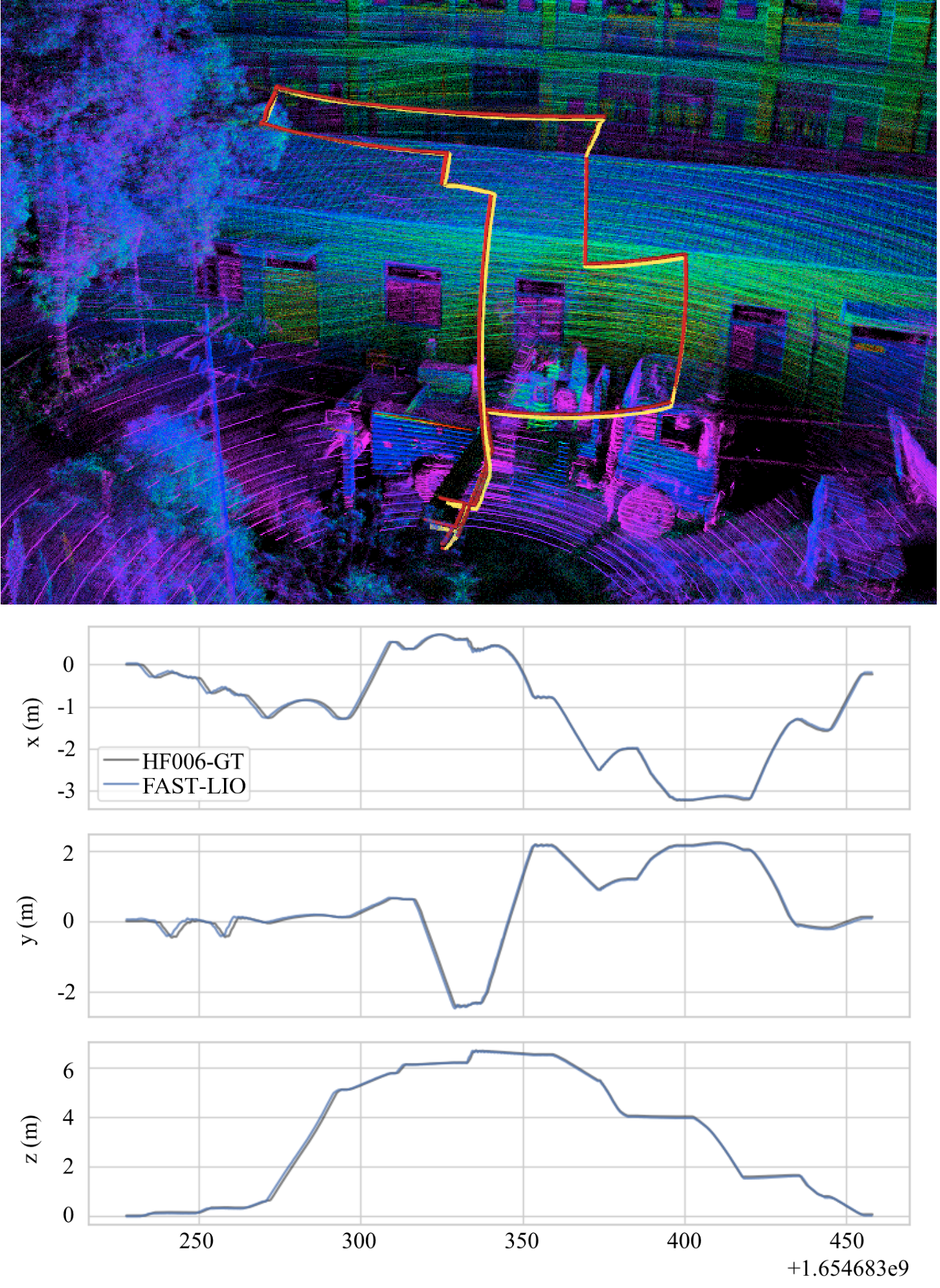}
	\caption{The ground truth is marked in red and the SLAM trajectory is marked in yellow. The figure shows the running results of Fast-LIO on the \revise{hf006} sequence.}
	\label{fast_lio_gt}
\end{figure}

The LiDAR SLAM algorithms that were tested on data sequences from a horizontal LiDAR achieved good results, with an ATE of approximately 0.1 meters. The results are reported in Figure~\ref{fast_lio_gt} Additionally, the accuracy of LiDAR SLAM can be further improved by tightly coupling a well-calibrated IMU. This suggests that among current sensing technologies, LiDAR is relatively reliable for the localization and mapping of aerial work robots.

However, LiDAR is not without its flaws. Issues have been observed when using the vertical LiDAR. It is necessary for aerial work robots to have 360-degree perception in the vertical direction, as obstacles may come from any direction while in motion. Merely perceiving in the horizontal direction is not sufficient. In this comparison, we have chosen algorithms that use a single LiDAR to minimize the influence of calibration errors. As shown in the table, the ATE of A-LOAM on the vertical LiDAR is 3 to 10 times higher than on the horizontal LiDAR in the same sequence, despite slight differences in the LiDAR models. This experimental result is convincing enough. When the field of view rotates 180 degrees, the world seen by the LiDAR is very different. For the lower half of the LiDAR, the ground and walls occupy most of the point cloud; for the upper half of the LiDAR, only scattered small objects such as wires and branches in the air can reflect LiDAR echoes, and most of the lasers disappear in the sky. This is a typical scenario where LiDAR odometry fails — low texture. Besides, due to the large range of rotations of the aerial platform, the overlap between adjacent LiDAR frames is small during rotations, which is not conducive to the continuous tracking of features. This is evident in sequences hf003, hf007, hf008, and hf0009, which all have large rotations and therefore larger ATE values compared to other sequences. The aerial platform itself can also act as a dynamic object and cause interference. Among the algorithms tested, LeGO-LOAM tested on the vertical LiDAR performed the worst because it is difficult to extract the ground plane from the point cloud during aerial motion. In this case, optimization for ground motion deteriorates the accuracy of the algorithm.

Overall, the results of the evaluations show that LiDAR-based SLAM algorithms tend to have good accuracy, with an ATE of approximately 0.1 meters when using a horizontal LiDAR. The accuracy of visual SLAM algorithms varied depending on the sensor and the lighting conditions, with ORB-SLAM3 generally performing better in good lighting conditions and VINS-based algorithms being more robust in extreme lighting conditions. The accuracy of visual SLAM can also be improved by tightly coupling a well-calibrated IMU. LiDAR-based SLAM algorithms tended to have difficulty in scenarios with low texture or large rotations, with the ATE increasing significantly in these cases.

We did not aim to intentionally make it difficult for these algorithms. To verify whether the accuracy of the algorithms decreases under faster or more aggressive movements, you can try downsampling the data and running the algorithms to simulate faster speeds. We have only evaluated a select number of classic SLAM methods in this study. More recent SLAM methods, such as LVI-SAM \cite{shan2021lvi}, which combines VINS-Mono and LIO-SAM, may improve upon these classic methods.

\begin{figure*}[t]
	\centering	\includegraphics[height=8.25cm,width=16cm]{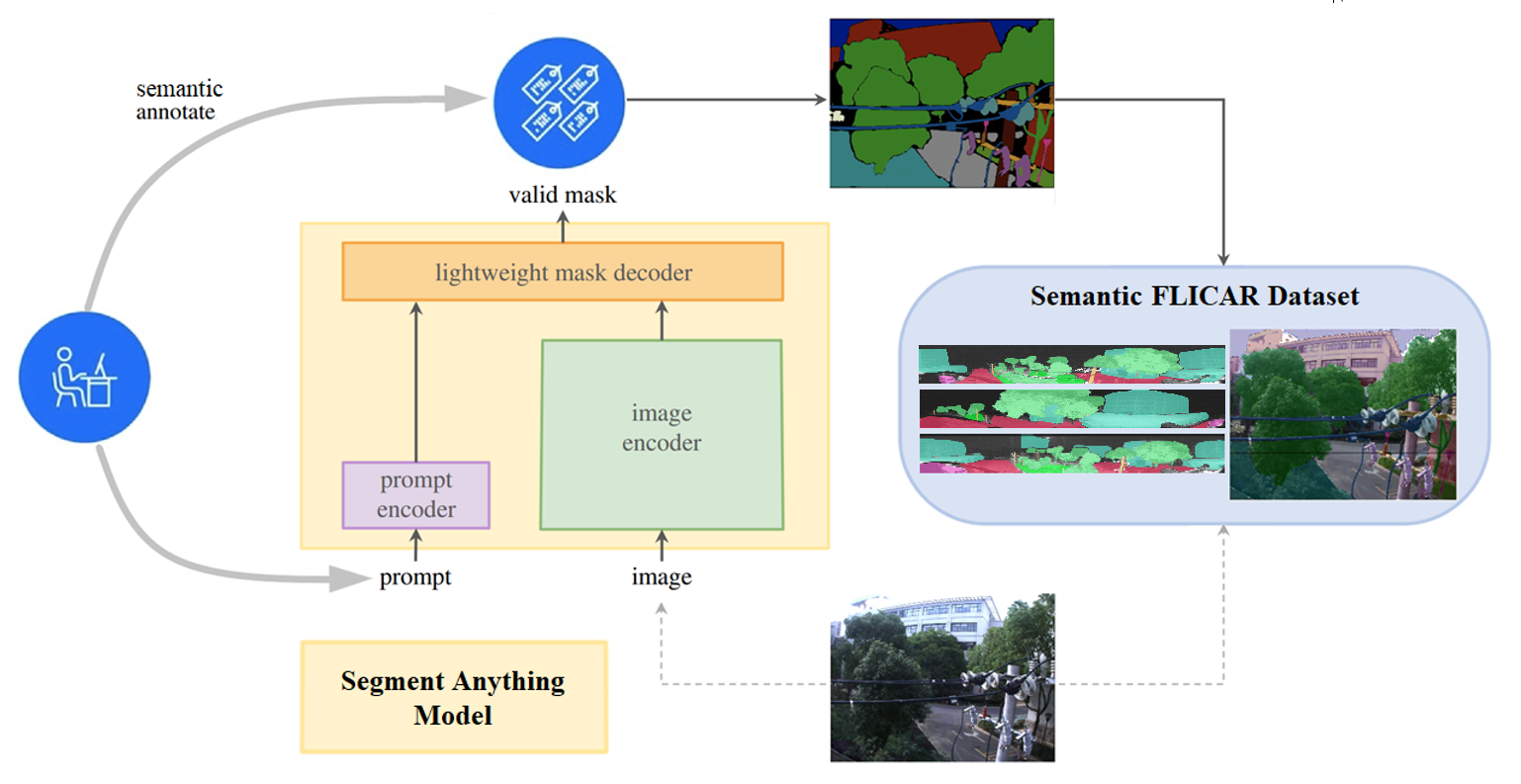}
	\caption{\revise{Semantic annotation data engine based on SAM: Semantic FLICAR dataset involves the initial generation of global mask segmentations by the Segment Anything Model (SAM), followed by collaborative refinement by human annotators through interactive prompts for regions of interest, resulting in precise and detailed annotations.}}
	\label{SAM}
\end{figure*}

\begin{figure*}[t]
	\centering	\includegraphics[height=12.9828cm,width=17cm]{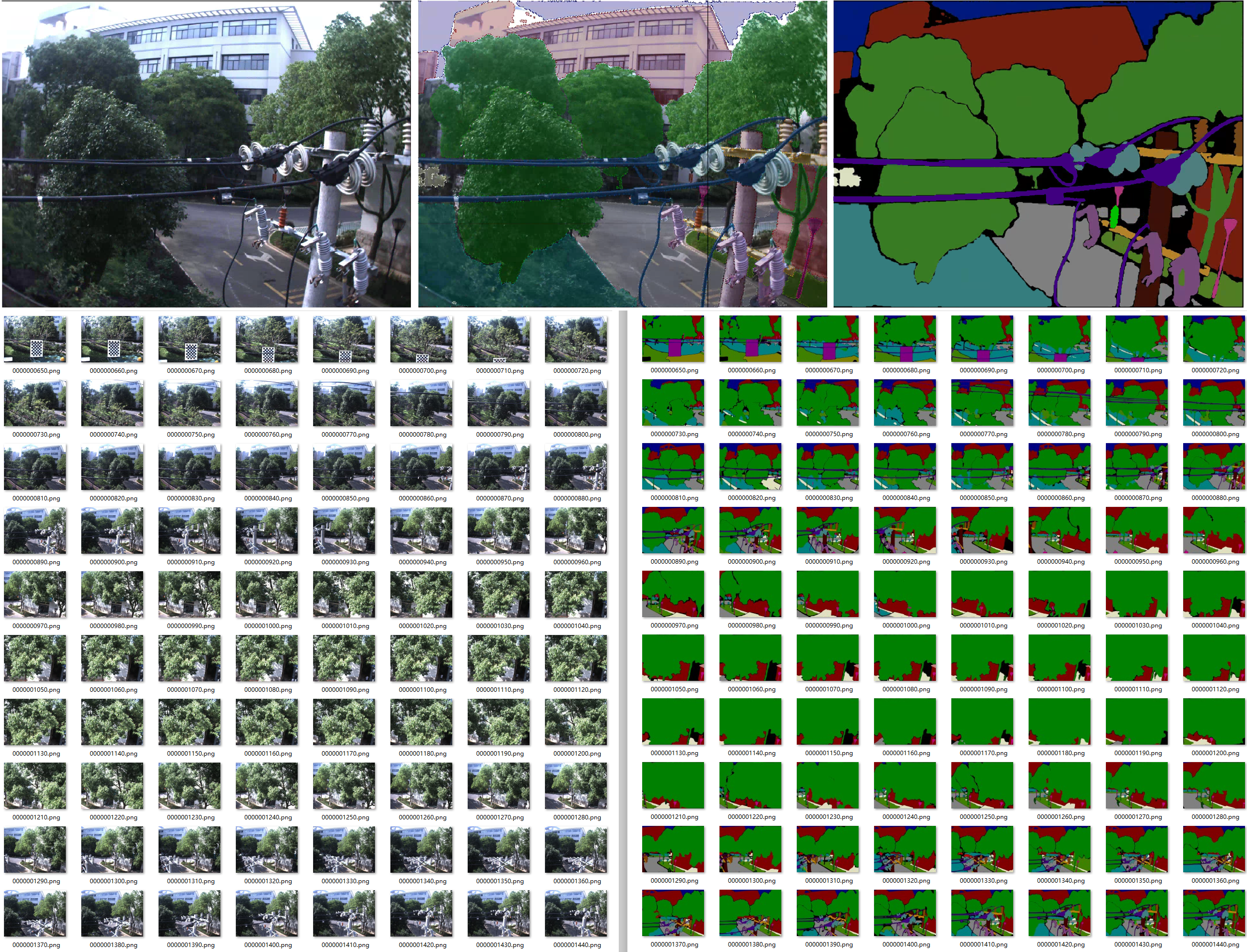}
	\caption{\revise{Bumblebee XB3 center image semantic segmentation preview in HF001 sequence}}
	\label{xb3_annotation_preview}
\end{figure*}

\begin{figure}[t]
	\centering	\includegraphics[height=11.33333cm,width=8.5cm]{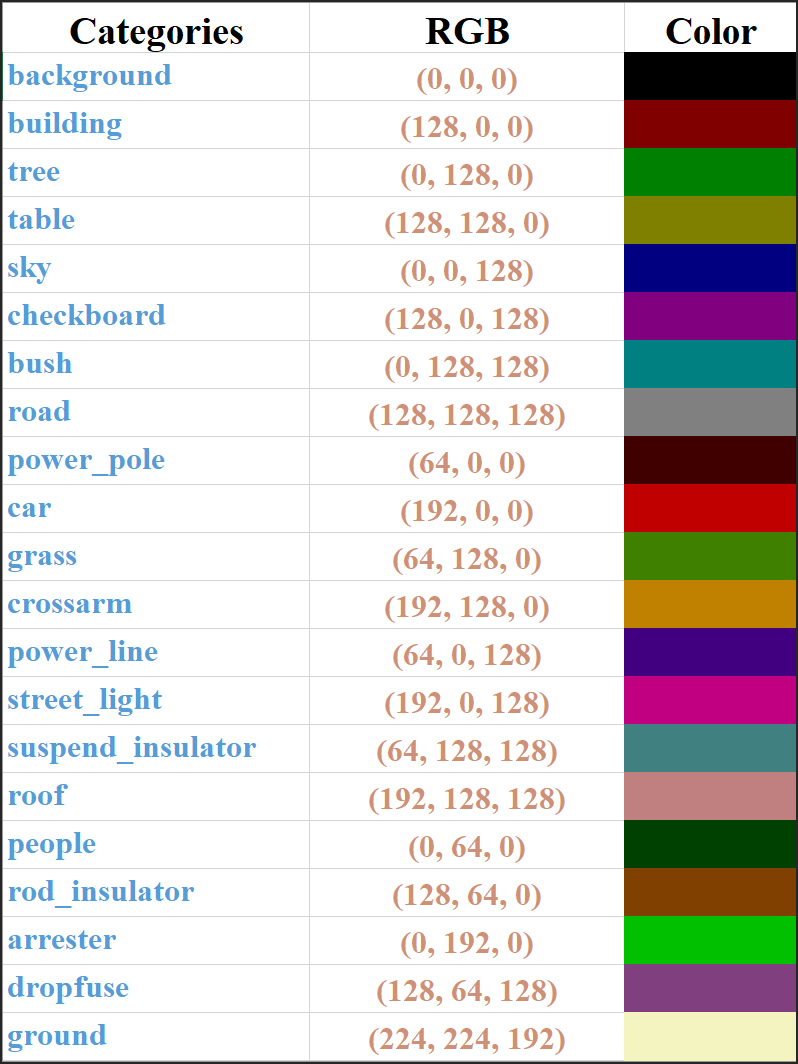}
	\caption{\revise{Semantic label categories for Bumblebee XB3 center image data}}
	\label{xb3_sam_color}
\end{figure}

\begin{figure}[t]
	\centering	\includegraphics[height=5.1cm,width=8.5cm]{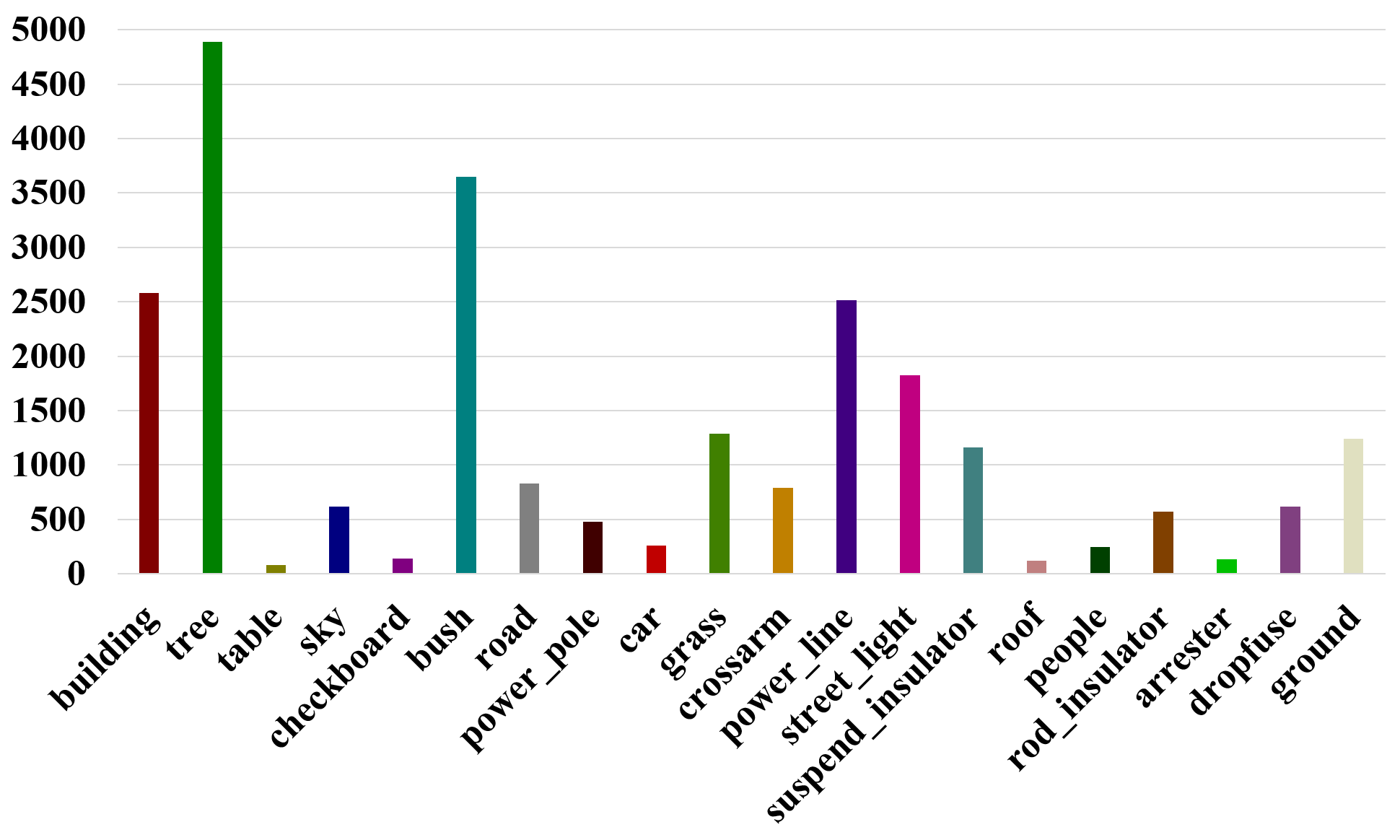}
	\caption{\revise{Semantic maskes annotation statistics for Bumblebee xb3 center image data}}
	\label{xb3_sam_stat}
\end{figure}

\begin{figure}[t]
	\centering	\includegraphics[height=6.15454545cm,width=8.5cm]{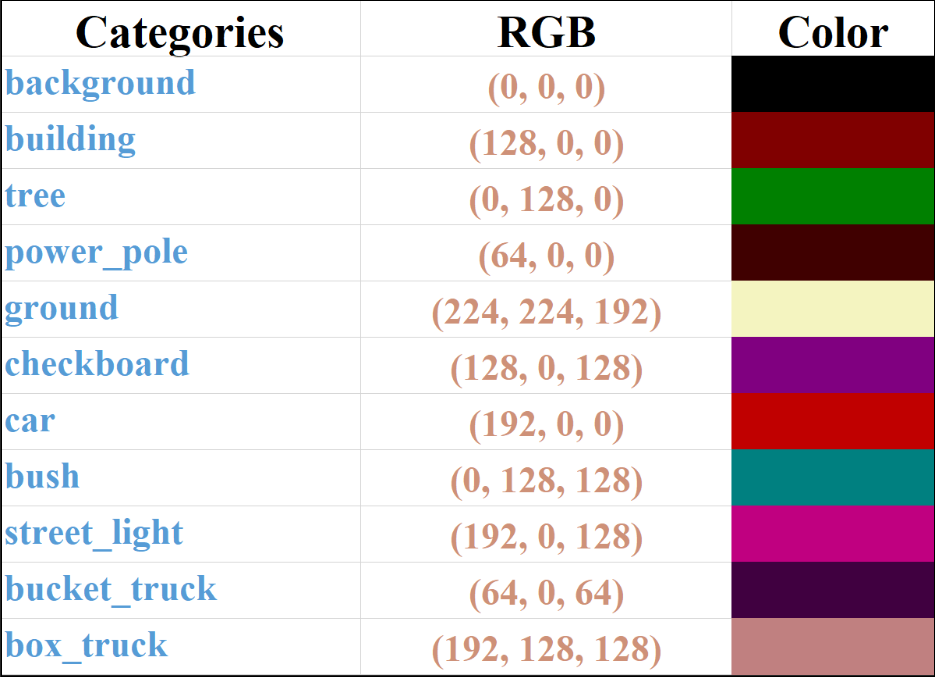}
	\caption{\revise{Semantic label categories for Ouster OS0-128 signal image data}}
	\label{ouster_sam_color}
\end{figure}

\begin{figure}[t]
	\centering	\includegraphics[height=5.1cm,width=8.5cm]{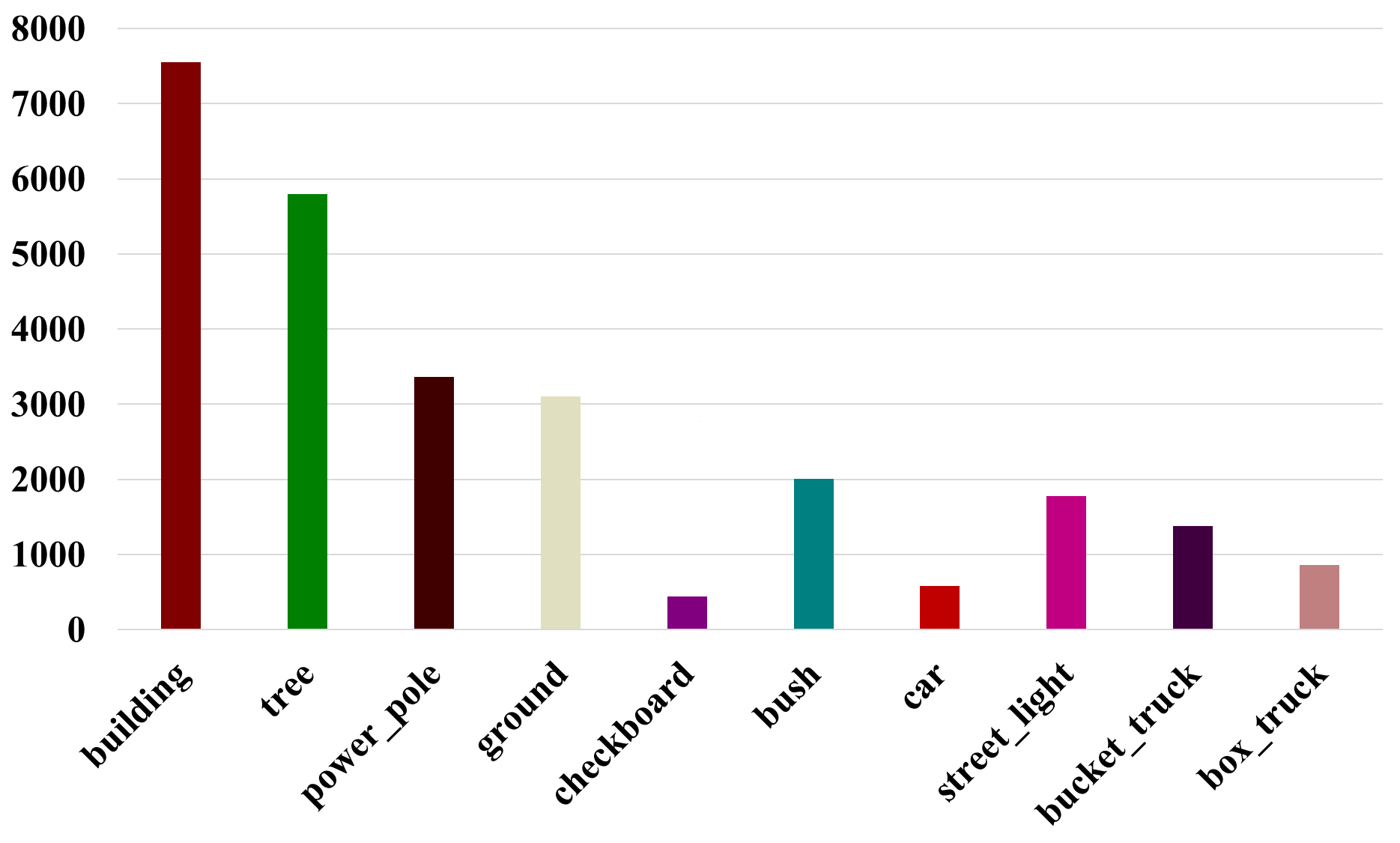}
	\caption{\revise{Semantic maskes annotation statistics for Ouster OS0-128 signal image data}}
	\label{ouster_sam_stat}
\end{figure}

\begin{figure*}[t]
	\centering	\includegraphics[height=12.3396cm,width=17cm]{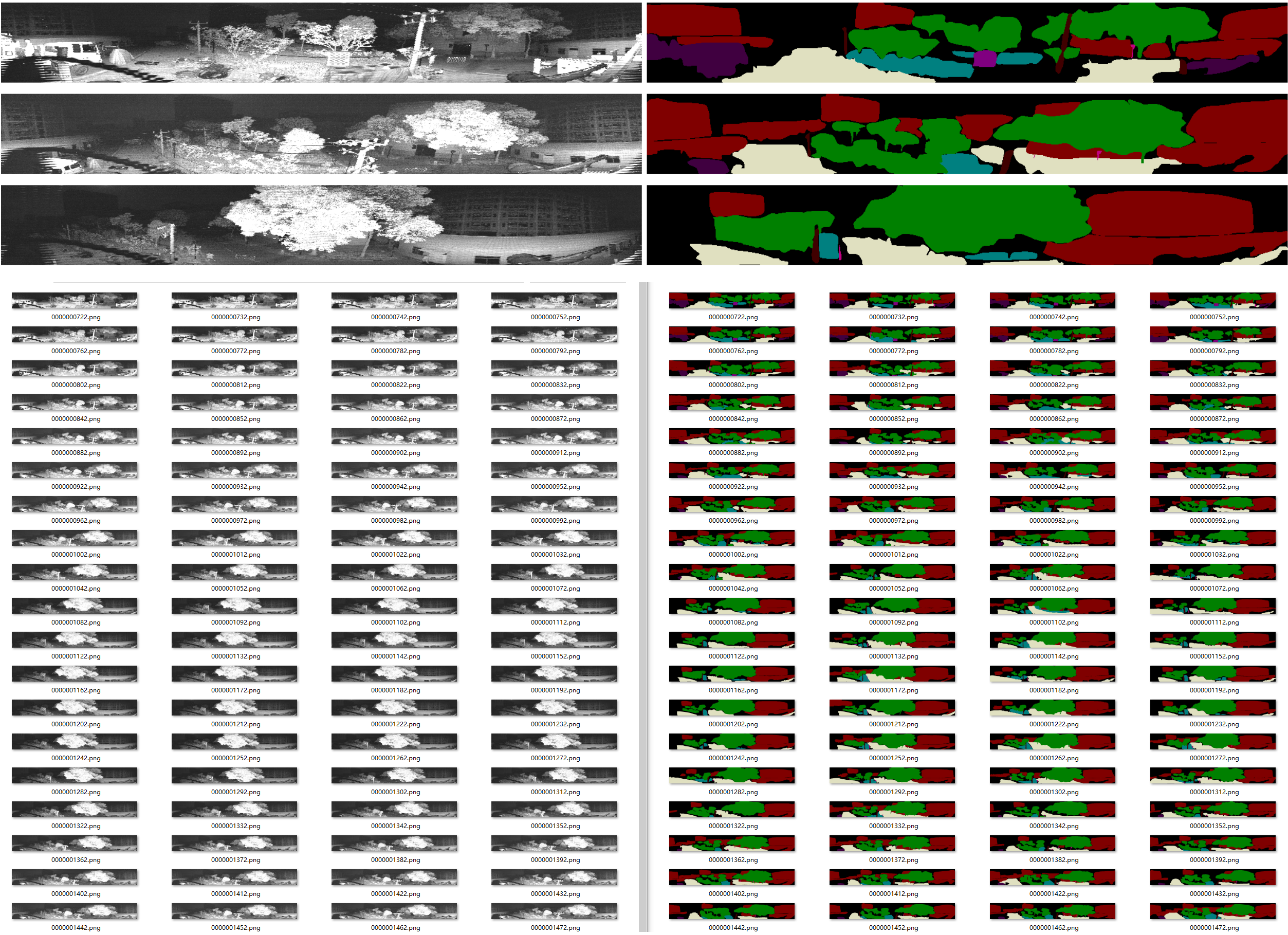}
	\caption{\revise{Ouster OS0-128 signal image semantic segmentation preview in HF001 sequence}}
	\label{ouster_annotation_preview}
\end{figure*}

\revise{
\section{Semantic Annotations} \label{section6}
In this section, we will present the Semantic FLICAR dataset, including its generation process, content, and distinctive features. This dataset can be downloaded from the following website: \url{https://ustc-flicar.github.io/semantic/}.
}

\revise{
\subsection{Annotation Process}
The rapid progress in artificial intelligence (AI) and the foundation AI Large Vision Models (LVMs) has opened up new possibilities for collaborative data annotation between human annotators and AI systems, resulting in significant improvements in data annotation and production efficiency.
}

\revise{
Building upon the capabilities of the Segment Anything Model (SAM) \cite{kirillov2023segment}, we present the Semantic FLICAR dataset, which provides fine-grained image semantic segmentation and 128-beam LiDAR ring signal scene segmentation for the USTC FLICAR dataset. The Semantic FLICAR dataset offers detailed annotations that greatly enhance our understanding of object boundaries and semantic information in aerial work scenes, making it an invaluable resource for various aerial work computer vision applications.
}

\revise{
The data annotation process for the Semantic FLICAR dataset is illustrated in Figure~\ref{SAM}. Initially, the USTC FLICAR dataset is input into the Segment Anything Model (SAM), automatically generating global mask segmentations. This step leverages SAM's efficiency and promptability to produce initial annotations quickly. Human annotators actively participate in the process to further improve the quality and precision of the annotations. Human annotators interact with the initial masks generated by SAM and provide prompts on regions of interest. This collaborative approach allows the annotators to refine the boundaries and assign accurate semantic information, resulting in more precise annotations. The combination of AI-generated masks and human expertise ensures that the Semantic FLICAR dataset achieves a high level of accuracy and detail.
}

\revise{
\subsection{Semantic FLICAR Dataset}
The semantic segmentation data of Semantic FLICAR is provided in the format of the VOC segmentation dataset \cite{everingham2010pascal}. In total, 1,774 camera keyframes and 1,426 LiDAR keyframes were annotated at 1 Hz, resulting in 56,889 unique semantic masks. 
}

\revise{
 When annotating camera images, the higher resolution allows for the definition of more diverse and fine-grained object categories. As depicted in Figure~\ref{xb3_sam_color}, the annotation of the Bumblebee xb3 center image in the HF0XX sequence includes 20 distinct object categories, covering various objects commonly found in aerial work scenarios. The statistical charts for each category label can be found in Figure~\ref{xb3_sam_stat}. For a preview of the annotated camera dataset, please refer to Figure~\ref{xb3_annotation_preview}.
}

\revise{
Keyframe annotation for Ouster LiDAR is typically paired with camera data. This means that the closest timestamped pair of camera and LiDAR data will be annotated. Unlike camera images, LiDAR has a lower resolution but a wide field of view. In our system, we choose to annotate using the highest-resolution Ouster OS0-128 LiDAR. The annotation is performed on the signal image data, which is a panoramic image with a resolution of 1024$\times$128. This signal image is derived from the near-infrared signals and environmental data captured by the Ouster LiDAR. It is spatially aligned with other data layers of the Ouster LiDAR, such as point clouds, which is perfectly spatially correlated, with zero temporal mismatch or shutter effects, and has 16 bits per pixel and linear photo response. As depicted in Figure~\ref{ouster_sam_color}, the annotation of the Ouster OS0-128 signal image includes 10 distinct object categories. The statistical charts for each category label can be found in Figure~\ref{ouster_sam_stat}. For a preview of the annotated LiDAR dataset, please refer to Figure~\ref{ouster_annotation_preview}.
}

\begin{figure}[t]
\centering
\includegraphics[height=8.94986cm,width=8.5cm]{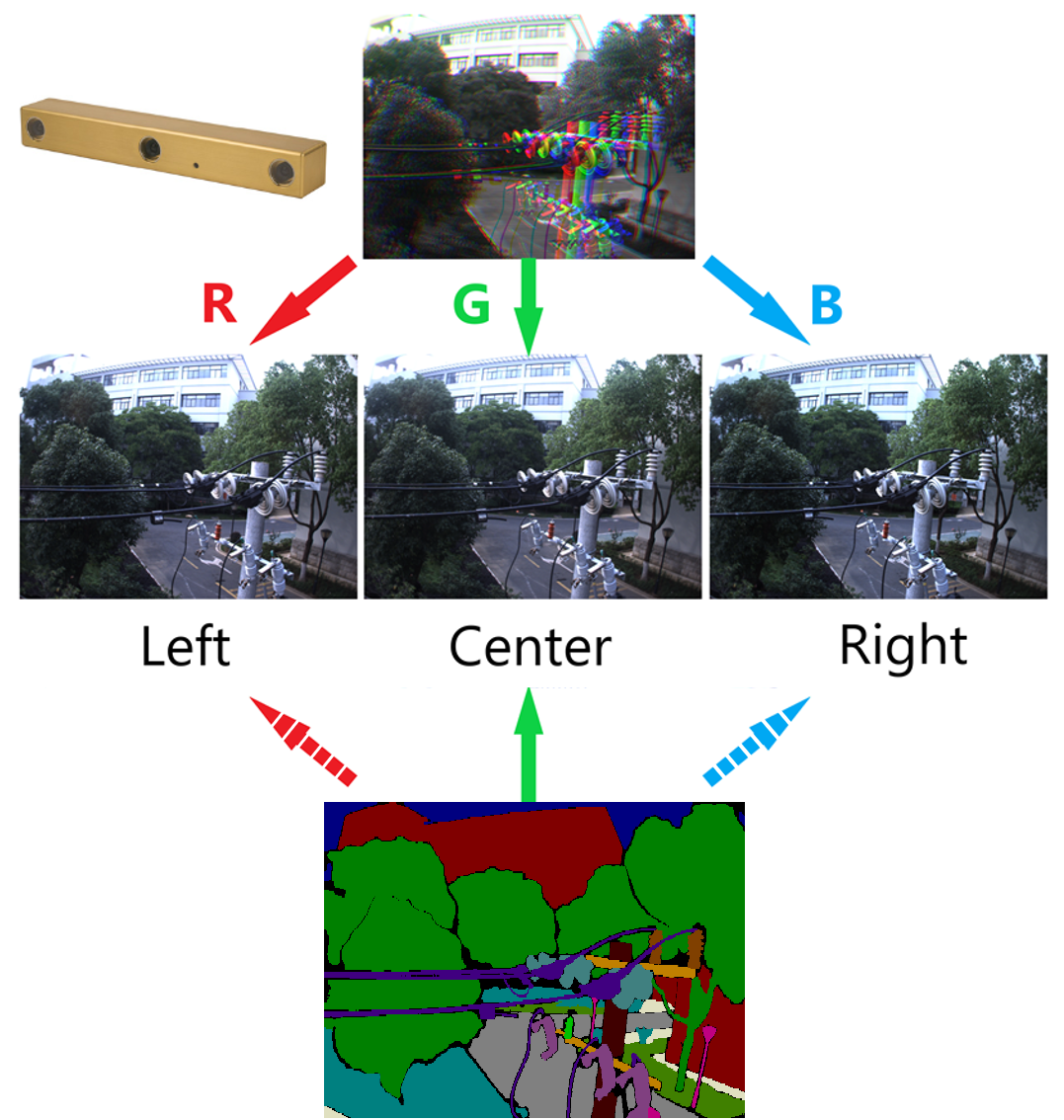}
\caption{\revise{Data association and augmentation based on stereo camera calibration results in section~\ref{visual-calib}}}
\label{index_stereo}
\end{figure}

\begin{figure}[t]
\centering
\includegraphics[height=6.30759cm,width=8.5cm]{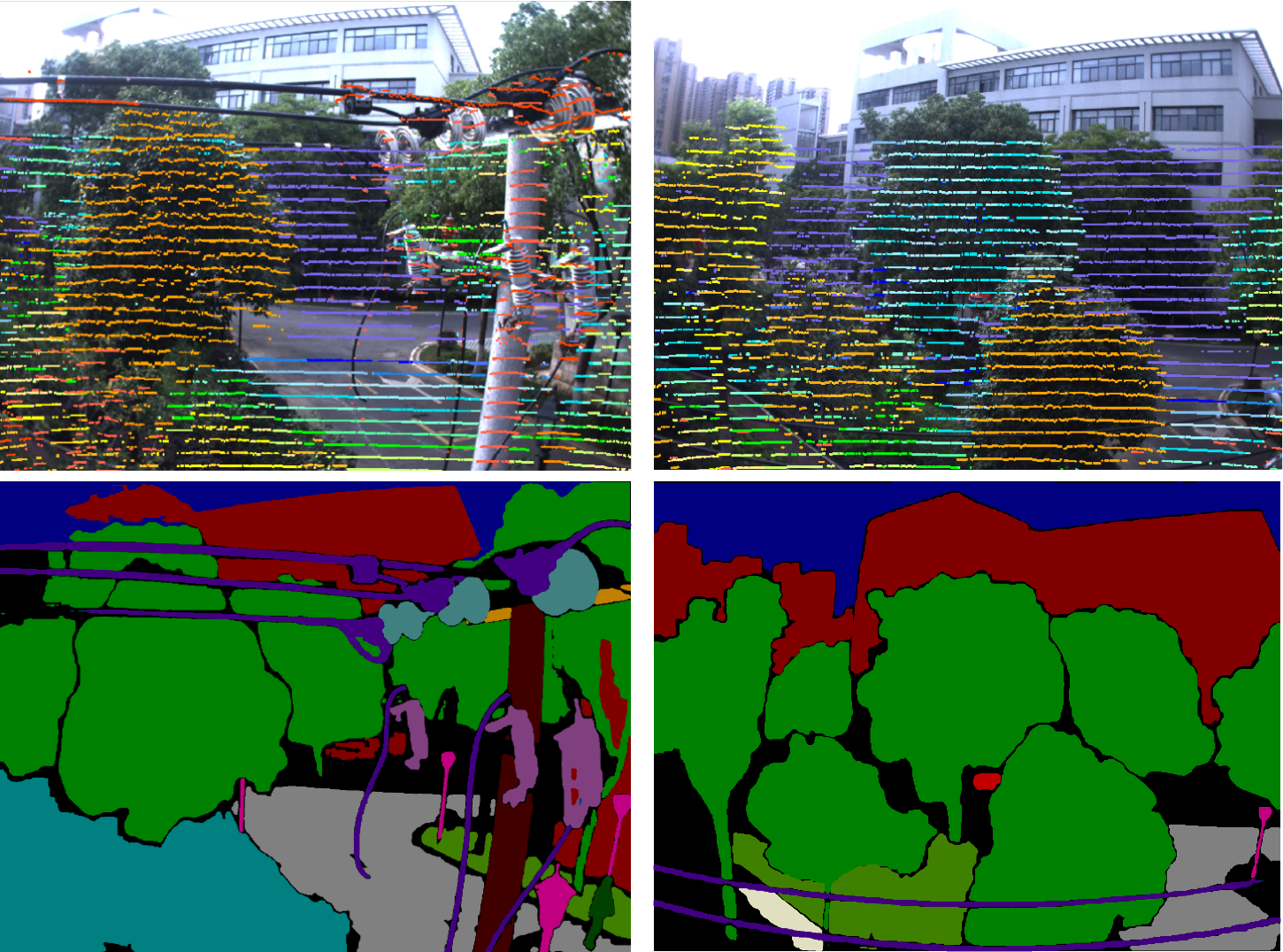}
\caption{\revise{Data association and augmentation based on LiDAR camera calibration results in section~\ref{lidar-camera-calib}}}
\label{index_velo2cam}
\end{figure}

\begin{figure}[t]
\centering
\includegraphics[height=6.6cm,width=8.5cm]{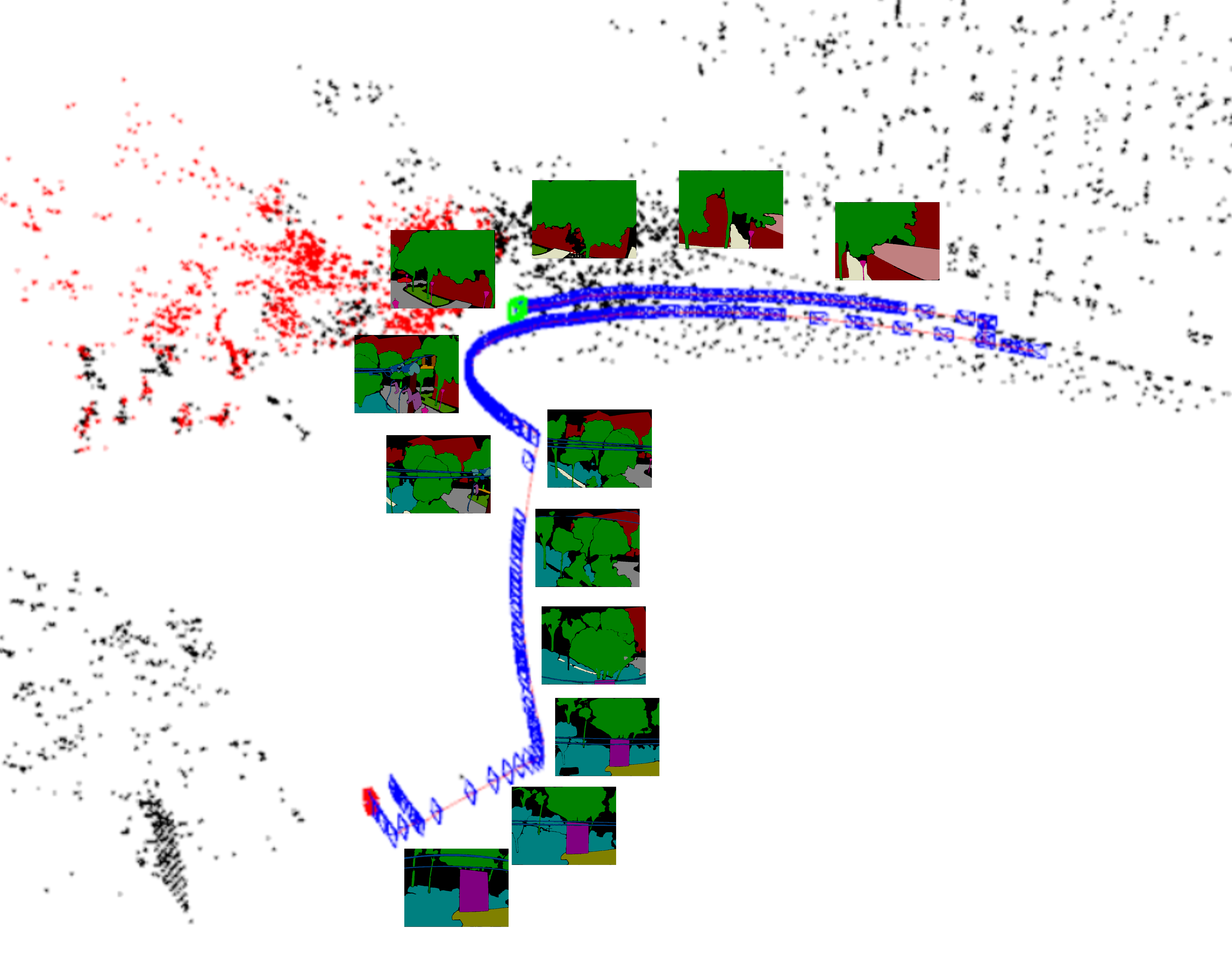}
\caption{\revise{Data association and augmentation based on SLAM results in section~\ref{section5}}}
\label{index_orb}
\end{figure}
\revise{
\subsection{Data Association and Augmentation}
Unlike datasets that focus only on semantic segmentation tasks, the Semantic FLICAR dataset is built upon the USTC FLICAR dataset. Therefore, the previously mentioned sensor calibration and SLAM-related content will play a crucial role in data association and augmentation for the Semantic FLICAR dataset.
}

\revise{
As depicted in Figure~\ref{index_stereo}, the segmentation annotations of the Bumblebee XB3 center image can be mapped to the left image and right image using the calibration results of the stereo camera. This mapping results in two additional stereo perspectives, effectively tripling the number of labels. In Figure~\ref{index_velo2cam}, the segmentation annotations of the Bumblebee XB3 center image can be mapped to the Velodyne HDL32 point cloud using the joint calibration results of the laser radar and the camera. This mapping process assists in incorporating semantic information into the low-resolution 32-beam LiDAR system.
}

\revise{
From the perspective of data association, SLAM is capable of linking together various elements such as scenes, objects, locations, and timestamps during the process of simultaneous localization and mapping. As shown in Figure~\ref{index_orb}, when employing the Bumblebee XB3 center camera for ORB-SLAM3, image frames are arranged according to their corresponding time and location. Therefore, for all image data collected by the Bumblebee camera at a rate of 10Hz, it can be associated based on the results of SLAM, allowing for annotations to be inserted between keyframes marked at a rate of 1Hz. This slightly compromises fidelity but greatly doubles the number of annotated frames.
}

\revise{
The Semantic FLICAR dataset encompasses a significant number of annotated images, with comprehensive fine-grained semantic segmentation and 128-beam Ouster LiDAR ring signal image segmentation. This dataset offers researchers a rich resource for training, evaluating, and advancing algorithms and models in various computer vision tasks, including object recognition, scene segmentation and understanding, and autonomous aerial work applications. The impact of the Semantic FLICAR dataset extends to practical applications as well. The meticulous annotations within this dataset can contribute to advancements in autonomous aerial work systems, robotics, and other domains that rely on precise scene analysis and understanding. By fostering research in these areas, the Semantic FLICAR dataset plays a crucial role in the development of safer and more intelligent AI systems for aerial work tasks.
}

\revise{
\section{Known Issues} \label{section7}
}
\revise{
Despite careful design and execution of the data collection experiments, we are aware of various issues that present additional challenges in processing and limit the achievable accuracy when compared to ground truth.
}

\revise{These are the issues that users will need to be aware of:}

\revise{
\textbullet\hspace{0.5em}Camera overexposure in direct sunlight and insufficient lighting at night, as shown in Figure~\ref{camera_light}:
    In certain conditions, such as direct sunlight, the camera may experience overexposure, resulting in washed-out or overly bright images. Conversely, during nighttime or low-light situations, the lighting may be inadequate, leading to underexposed images with limited visibility. Additionally, there might be the presence of lens flares or halos, which can occur when strong light sources are present in the frame.
}

\begin{figure}[t]
	\centering	\includegraphics[height=2.107534cm,width=8.5cm]{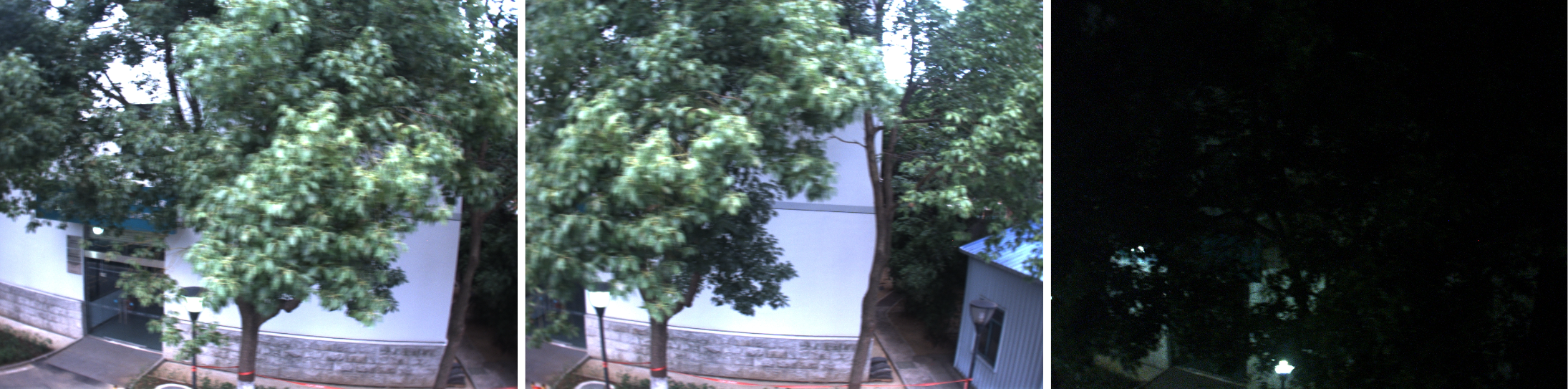}
	\caption{\revise{Motion blur in poor lighting conditions, data form HF007 and HF008}}
	\label{blur}
\end{figure}

\revise{
\textbullet\hspace{0.5em}Motion blur in poor lighting conditions, as shown in Figure~\ref{blur}:
    In challenging lighting conditions where the illumination is insufficient, the camera extends the exposure time to capture more light. As a result, when there is fast motion, such as large-scale rotations, motion blur may occur in the captured images. This motion blur can lead to reduced image clarity and potentially impact the accuracy of certain computer vision tasks.
}

\revise{
\textbullet\hspace{0.5em}Attention to timestamps in Ouster and LiVOX LiDAR rostopics:
Please pay attention to the timestamps recorded within the Ouster and LiVOX LiDAR rostopics. During the entire experimental process, the availability of GPS signals cannot be guaranteed at all times. In situations where the GPS signal is interrupted, the driver programs of these two LiDAR sensors may record the time elapsed since startup as timestamps instead of UTC time. To ensure accurate UTC timestamp synchronization, it is recommended to use the timestamps provided by the rostopic as a reference, as they record the computer system time. Both the computers that record Ouster and LiVOX LiDAR sensor data synchronize their time using the same server through NTP with others.
}

\revise{
\textbullet\hspace{0.5em}Attention to the accuracy of semantic segmentation label edges:
Please pay attention to the accuracy of the semantic segmentation labels, particularly at the edges. The semantic labels in our Semantic FLICAR dataset were generated by prompting SAM. As a result, the semantic segmentation labels may not achieve pixel-level accuracy, especially at the edges of objects. Additionally, the pixel accuracy of the labels in the Ouster OS0-128 signal images may be lower compared to the visible light images due to the lower resolution of the Ouster LIDAR sensor.
}

\section{Summary and future work} \label{section8}

The USTC FLICAR is a unique dataset that \revise{focuses} on the task of heavy-duty aerial work, featuring a special aerial platform, the bucket truck, which allows for greater payload capacity and stationarity compared to traditional drones. It is also the most sensor-rich aerial dataset to date, with a wide range of sensors including seven cameras, four 3D LiDARs, and three IMUs, which \revise{covers} 360 degrees of horizontal and vertical views. This dataset is designed to enable aerial work robots to effectively interact with complex aerial work environments, with millimeter-level outdoor ground truth obtained using a laser tracker. It is our hope that this dataset will serve as a valuable benchmark for evaluating the performance of various algorithms in this field, and inspire researchers to design sensor suites specifically tailored for autonomous aerial work systems. Moreover, the experimental results on our dataset also demonstrate that the novel combination of \revise{an} autonomous driving sensing kit and bucket truck is a general autonomous aerial platform with high potential for various aerial work tasks.

\revise{
 Looking ahead to the future, our goal is to expand into multiple domains and further advance the practical application of aerial robotics. One key aspect is to enrich and refine the semantic annotations based on existing labeled data, which will enable aerial robots to have a more accurate and detailed understanding of the environment and its objects. Additionally, we plan to incorporate more sensors and explore new sensor configurations. For instance, our aim is to include infrared night vision cameras and downward-facing cameras to enhance the dataset's capabilities. These additional features will significantly improve the overall abilities of the aerial robots. Furthermore, we intend to continue collecting new data to cover a broader range of scenarios and environments. This approach allows us to enhance the robustness and generality of our algorithms and models. By continuously expanding our dataset, we can improve the effectiveness and reliability of aerial robots in various real-world situations. In conclusion, we firmly believe that the USTC FLICAR dataset represents an important milestone in making aerial work safer, more efficient, and more accessible. Through our continuous efforts, our goal is to push the boundaries of innovation and contribute to the ongoing progress of aerial robotics technology.
}

\begin{acks}
The authors would like to acknowledge Yonggang Gu at the Experimental Center of Engineering and Material Sciences of USTC for his assistance in training and providing the laser tracking equipment. The authors would like to express our gratitude to STATE GRID ANHUI ELECTRIC POWER CO. LTD. for their support in providing us with a bucket truck. \revise{The author would like to express gratitude to Meta AI Research, FAIR for generously open-sourcing the Segment Anything project. This invaluable contribution played a pivotal role in enabling us to successfully construct the Semantic FLICAR dataset.} The authors gratefully acknowledge support from the National Key R\&D Program of China (No. 2018YFB1307403), the Fundamental Research Funds for the Central Universities, and the Cyrus Tang Foundation.
\end{acks}

\bibliographystyle{agsm}
\bibliography{main}

@article{huang2010high,
	title={{A high-rate, heterogeneous data set from the darpa urban challenge}},
	author={Huang, Albert S and Antone, Matthew and Olson, Edwin and Fletcher, Luke and Moore, David and Teller, Seth and Leonard, John},
	journal={The International Journal of Robotics Research},
	volume={29},
	number={13},
	pages={1595--1601},
	year={2010},
	publisher={Sage Publications Sage UK: London, England}
}

@article{pandey2011ford,
	title={{Ford campus vision and lidar data set}},
	author={Pandey, Gaurav and McBride, James R and Eustice, Ryan M},
	journal={The International Journal of Robotics Research},
	volume={30},
	number={13},
	pages={1543--1552},
	year={2011},
	publisher={SAGE Publications Sage UK: London, England}
}

@article{Geiger2013IJRR,
	author = {Andreas Geiger and Philip Lenz and Christoph Stiller and Raquel Urtasun},
	title = {{Vision meets Robotics: The KITTI Dataset}},
	journal = {International Journal of Robotics Research (IJRR)},
	year = {2013}
}

@article{carlevaris2016university,
	title={{University of Michigan North Campus long-term vision and lidar dataset}},
	author={Carlevaris, Nicholas and Ushani, Arash K and Eustice, Ryan M},
	journal={The International Journal of Robotics Research},
	volume={35},
	number={9},
	pages={1023--1035},
	year={2016},
	publisher={Sage Publications Sage UK: London, England}
}

@article{maddern20171,
	title={{1 year, 1000 km: The Oxford RobotCar dataset}},
	author={Maddern, Will and Pascoe, Geoffrey and Linegar, Chris and Newman, Paul},
	journal={The International Journal of Robotics Research},
	volume={36},
	number={1},
	pages={3--15},
	year={2017},
	publisher={SAGE Publications Sage UK: London, England}
}

@article{RadarRobotCarDatasetArXiv,
	author = {Dan Barnes and Matthew Gadd and Paul Murcutt and Paul Newman and Ingmar Posner},
	title = {{The Oxford Radar RobotCar Dataset: A Radar Extension to the Oxford RobotCar Dataset}},
	journal = {arXiv preprint arXiv: 1909.01300},
	url = {https://arxiv.org/pdf/1909.01300},
	year = {2019}
}

@article{nguyen2022ntu,
  title={{Ntu viral: A visual-inertial-ranging-lidar dataset, from an aerial vehicle viewpoint}},
  author={Nguyen, Thien-Minh and Yuan, Shenghai and Cao, Muqing and Lyu, Yang and Nguyen, Thien H and Xie, Lihua},
  journal={The International Journal of Robotics Research},
  volume={41},
  number={3},
  pages={270--280},
  year={2022},
  publisher={SAGE Publications Sage UK: London, England}
}

@misc{grupp2017evo,
	title={{evo: Python package for the evaluation of odometry and SLAM.}},
	author={Grupp, Michael},
	howpublished={\url{https://github.com/MichaelGrupp/evo}},
	year={2017}
}

@article{pire2019rosario,
	title={{The Rosario dataset: Multisensor data for localization and mapping in agricultural environments}},
	author={Pire, Taih{\'u} and Mujica, Mart{\'\i}n and Civera, Javier and Kofman, Ernesto},
	journal={The International Journal of Robotics Research},
	volume={38},
	number={6},
	pages={633--641},
	year={2019},
	publisher={SAGE Publications Sage UK: London, England}
}

@article{jeong2019complex,
	title={{Complex urban dataset with multi-level sensors from highly diverse urban environments}},
	author={Jeong, Jinyong and Cho, Younggun and Shin, Young-Sik and Roh, Hyunchul and Kim, Ayoung},
	journal={The International Journal of Robotics Research},
	volume={38},
	number={6},
	pages={642--657},
	year={2019},
	publisher={SAGE Publications Sage UK: London, England}
}

@inproceedings{yan2020eu,
	title={{EU long-term dataset with multiple sensors for autonomous driving}},
	author={Yan, Zhi and Sun, Li and Krajn{\'\i}k, Tom{\'a}{\v{s}} and Ruichek, Yassine},
	booktitle={2020 IEEE/RSJ International Conference on Intelligent Robots and Systems (IROS)},
	pages={10697--10704},
	year={2020},
	organization={IEEE}
}

@inproceedings{caesar2020nuscenes,
	title={{nuScenes: A multimodal dataset for autonomous driving}},
	author={Caesar, Holger and Bankiti, Varun and Lang, Alex H and Vora, Sourabh and Liong, Venice Erin and Xu, Qiang and Krishnan, Anush and Pan, Yu and Baldan, Giancarlo and Beijbom, Oscar},
	booktitle={Proceedings of the IEEE/CVF conference on computer vision and pattern recognition},
	pages={11621--11631},
	year={2020}
}

@article{burri2016euroc,
	title={{The EuRoC micro aerial vehicle datasets}},
	author={Burri, Michael and Nikolic, Janosch and Gohl, Pascal and Schneider, Thomas and Rehder, Joern and Omari, Sammy and Achtelik, Markus W and Siegwart, Roland},
	journal={The International Journal of Robotics Research},
	volume={35},
	number={10},
	pages={1157--1163},
	year={2016},
	publisher={SAGE Publications Sage UK: London, England}
}

@article{majdik2017zurich,
	title={{The Zurich urban micro aerial vehicle dataset}},
	author={Majdik, Andr{\'a}s L and Till, Charles and Scaramuzza, Davide},
	journal={The International Journal of Robotics Research},
	volume={36},
	number={3},
	pages={269--273},
	year={2017},
	publisher={SAGE Publications Sage UK: London, England}
}

@inproceedings{delmerico2019we,
	title={{Are we ready for autonomous drone racing? the UZH-FPV drone racing dataset}},
	author={Delmerico, Jeffrey and Cieslewski, Titus and Rebecq, Henri and Faessler, Matthias and Scaramuzza, Davide},
	booktitle={2019 International Conference on Robotics and Automation (ICRA)},
	pages={6713--6719},
	year={2019},
	organization={IEEE}
}

@article{zhang2000flexible,
	title={{A flexible new technique for camera calibration}},
	author={Zhang, Zhengyou},
	journal={IEEE Transactions on pattern analysis and machine intelligence},
	volume={22},
	number={11},
	pages={1330--1334},
	year={2000},
	publisher={IEEE}
}

@article{beltran2022,  
	author={Beltrán, Jorge and Guindel, Carlos and de la Escalera, Arturo and García, Fernando},  
	journal={IEEE Transactions on Intelligent Transportation Systems},   
	title={{Automatic Extrinsic Calibration Method for LiDAR and Camera Sensor Setups}},   
	year={2022}, 
	doi={10.1109/TITS.2022.3155228}
}

@inproceedings{rehder2016extending,
	title={{Extending kalibr: Calibrating the extrinsics of multiple IMUs and of individual axes}},
	author={Rehder, Joern and Nikolic, Janosch and Schneider, Thomas and Hinzmann, Timo and Siegwart, Roland},
	
	booktitle={2016 IEEE International Conference on Robotics and Automation (ICRA)},
	pages={4304--4311},
	year={2016},
	organization={IEEE}
}

@inproceedings{olson2011apriltag,
	title={{AprilTag: A robust and flexible visual fiducial system}},
	author={Olson, Edwin},
	booktitle={2011 IEEE international conference on robotics and automation},
	pages={3400--3407},
	year={2011},
	organization={IEEE}
}

@TechReport{Geneva2020TRVICON2GT,
	Title = {{vicon2gt: Derivations and Analysis}},
	Author = {Patrick Geneva and Guoquan Huang},
	Number = {RPNG-2020-VICON2GT},
	Institution = {University of Delaware},
	Note = {Available: \url{http://udel.edu/~ghuang/papers/tr_vicon2gt.pdf}},
	Year = {2020},
}

@inproceedings{zhang2014loam,
  title={{LOAM: Lidar Odometry and Mapping in Real-time}},
  author={Zhang, Ji and Singh, Sanjiv},
  booktitle={Robotics: Science and Systems},
  volume={2},
  number={9},
  pages={1--9},
  year={2014},
  organization={Berkeley, CA}
}

@article{qiu2020real,
	title={{Real-time temporal and rotational calibration of heterogeneous sensors using motion correlation analysis}},
	author={Qiu, Kejie and Qin, Tong and Pan, Jie and Liu, Siqi and Shen, Shaojie},
	journal={IEEE Transactions on Robotics},
	volume={37},
	number={2},
	pages={587--602},
	year={2020},
	publisher={IEEE}
}

@inproceedings{sturm2012benchmark,
	title={{A benchmark for the evaluation of RGB-D SLAM systems}},
	author={Sturm, J{\"u}rgen and Engelhard, Nikolas and Endres, Felix and Burgard, Wolfram and Cremers, Daniel},
	booktitle={2012 IEEE/RSJ international conference on intelligent robots and systems},
	pages={573--580},
	year={2012},
	organization={IEEE}
}

@article{campos2021orb,
	title={{ORB-SLAM3: An Accurate Open-Source Library for Visual, Visual–Inertial, and Multimap SLAM}},
	author={Campos, Carlos and Elvira, Richard and Rodr{\'\i}guez, Juan J G{\'o}mez and Montiel, Jos{\'e} MM and Tard{\'o}s, Juan D},
	journal={IEEE Transactions on Robotics},
	volume={37},
	number={6},
	pages={1874--1890},
	year={2021},
	publisher={IEEE}
}

@article{qin2018vins,
	title={{VINS-Mono: A Robust and Versatile Monocular Visual-Inertial State Estimator}},
	author={Qin, Tong and Li, Peiliang and Shen, Shaojie},
	journal={IEEE Transactions on Robotics},
	volume={34},
	number={4},
	pages={1004--1020},
	year={2018},
	publisher={IEEE}
}

@inproceedings{shan2018lego,
	title={{LeGO-LOAM: Lightweight and Ground-Optimized Lidar Odometry and Mapping on Variable Terrain}},
	author={Shan, Tixiao and Englot, Brendan},
	booktitle={2018 IEEE/RSJ International Conference on Intelligent Robots and Systems (IROS)},
	pages={4758--4765},
	year={2018},
	organization={IEEE}
}

@inproceedings{shan2020lio,
	title={{LIO-SAM: Tightly-coupled Lidar Inertial Odometry via Smoothing and Mapping}},
	author={Shan, Tixiao and Englot, Brendan and Meyers, Drew and Wang, Wei and Ratti, Carlo and Rus, Daniela},
	booktitle={2020 IEEE/RSJ international conference on intelligent robots and systems (IROS)},
	pages={5135--5142},
	year={2020},
	organization={IEEE}
}

@article{xu2021fast,
	title={{FAST-LIO: A Fast, Robust LiDAR-Inertial Odometry Package by Tightly-Coupled Iterated Kalman Filter}},
	author={Xu, Wei and Zhang, Fu},
	journal={IEEE Robotics and Automation Letters},
	volume={6},
	number={2},
	pages={3317--3324},
	year={2021},
	publisher={IEEE}
}

@inproceedings{shan2021lvi,
	title={{LVI-SAM: Tightly-coupled Lidar-Visual-Inertial Odometry via Smoothing and Mapping}},
	author={Shan, Tixiao and Englot, Brendan and Ratti, Carlo and Rus, Daniela},
	booktitle={2021 IEEE international conference on robotics and automation (ICRA)},
	pages={5692--5698},
	year={2021},
	organization={IEEE}
}

@article{sola2012quaternion,
	title={{Quaternion kinematics for the error-state KF}},
	author={Sola, Joan},
	journal={Laboratoire dAnalyse et dArchitecture des Systemes-Centre national de la recherche scientifique (LAAS-CNRS), Toulouse, France, Tech. Rep},
	year={2012}
}

@article{li2022automatic,
  title={{Automatic targetless LiDAR–camera calibration: a survey}},
  author={Li, Xingchen and Xiao, Yuxuan and Wang, Beibei and Ren, Haojie and Zhang, Yanyong and Ji, Jianmin},
  journal={Artificial Intelligence Review},
  pages={1--39},
  year={2022},
  publisher={Springer}
}

@article{kirillov2023segment,
  title={{Segment anything}},
  author={Kirillov, Alexander and Mintun, Eric and Ravi, Nikhila and Mao, Hanzi and Rolland, Chloe and Gustafson, Laura and Xiao, Tete and Whitehead, Spencer and Berg, Alexander C and Lo, Wan-Yen and others},
  journal={arXiv preprint arXiv:2304.02643},
  year={2023}
}

@inproceedings{ndt,
  title={{The normal distributions transform: A new approach to laser scan matching}},
  author={Biber, Peter and Stra{\ss}er, Wolfgang},
  booktitle={Proceedings 2003 IEEE/RSJ International Conference on Intelligent Robots and Systems (IROS 2003)(Cat. No. 03CH37453)},
  volume={3},
  pages={2743--2748},
  year={2003},
  organization={IEEE}
}

@inproceedings{zhang2016degeneracy,
  title={{On degeneracy of optimization-based state estimation problems}},
  author={Zhang, Ji and Kaess, Michael and Singh, Sanjiv},
  booktitle={2016 IEEE International Conference on Robotics and Automation (ICRA)},
  pages={809--816},
  year={2016},
  organization={IEEE}
}

@inproceedings{segal2009generalized,
  title={{Generalized-ICP}},
  author={Segal, Aleksandr and Haehnel, Dirk and Thrun, Sebastian},
  booktitle={Robotics: science and systems},
  volume={2},
  number={4},
  pages={435},
  year={2009},
  organization={Seattle, WA}
}

@inproceedings{zhu2022robust,
  title={{Robust Real-time LiDAR-inertial Initialization}},
  author={Zhu, Fangcheng and Ren, Yunfan and Zhang, Fu},
  booktitle={2022 IEEE/RSJ International Conference on Intelligent Robots and Systems (IROS)},
  pages={3948--3955},
  year={2022},
  organization={IEEE}
}

@article{everingham2010pascal,
  title={{The PASCAL Visual Object Classes (VOC) Challenge}},
  author={Everingham, Mark and Van Gool, Luc and Williams, Christopher KI and Winn, John and Zisserman, Andrew},
  journal={International Journal of Computer Vision},
  volume={88},
  pages={303--338},
  year={2010},
  publisher={Springer}
}

\end{document}